\documentclass{article}

\usepackage[nonatbib, final]{neurips_2021}

\usepackage[utf8]{inputenc} %
\usepackage[T1]{fontenc}    %
\usepackage{hyperref}       %
\usepackage{url}            %
\usepackage{booktabs}       %
\usepackage{amsfonts}       %
\usepackage{nicefrac}       %
\usepackage{microtype}      %
\usepackage{xcolor}         %

\usepackage[numbers]{natbib}

\usepackage{multirow}
\usepackage{graphicx}
\usepackage{float}
\usepackage{placeins}
\usepackage[tableposition=above]{caption}
\usepackage{subcaption}
\definecolor{prussian_for_text}{RGB}{59, 100, 170}
\definecolor{coral}{RGB}{255,173,116}
\definecolor{gray}{RGB}{172,167,166}

\usepackage{wrapfig}
\usepackage{hanging}

\usepackage[export]{adjustbox}
\usepackage{lipsum}

\usepackage{enumitem} %
\usepackage{layouts}
\usepackage{multicol}

\title{How Well do Feature Visualizations Support Causal Understanding of CNN Activations?}

\author{%
  Roland S. Zimmermann\textsuperscript{* 1}\\
  
  \And
  Judy Borowski\textsuperscript{* 1} \\
  \AND
  Robert Geirhos\textsuperscript{1} \\
  \And
  Matthias Bethge\textsuperscript{$\dagger$ 1}\\
  \And
  Thomas S. A. Wallis\textsuperscript{$\dagger$ 2}\\
  \And
  Wieland Brendel\textsuperscript{$\dagger$ 1} \\
  \AND
  \normalfont \small{\textsuperscript{1} T\"ubingen AI Center, University of T\"ubingen, Germany.}\\
  \small{\textsuperscript{2} Institute of Psychology and Centre for Cognitive Science, Technical University of Darmstadt, Germany.}\\
  \small{\textsuperscript{*} Shared first authorship, determined by coin flip. \texttt{firstname.lastname@uni-tuebingen.de}}\\
  \small{\textsuperscript{$\dagger$} Joint supervision.}
}

\begin{document}

\maketitle

\begin{abstract}
A precise understanding of why units in an artificial network respond to certain stimuli would constitute a big step towards explainable artificial intelligence. One widely used approach towards this goal is to visualize unit responses via activation maximization. These synthetic feature visualizations are purported to provide humans with precise information about the image features that \emph{cause} a unit to be activated --- an advantage over other alternatives like strongly activating natural dataset samples. If humans indeed gain causal insight from visualizations, this should enable them to predict the effect of an intervention, such as how occluding a certain patch of the image (say, a dog's head) changes a unit's activation. Here, we test this hypothesis by asking humans to decide which of two square occlusions causes a larger change to a unit's activation.
Both a large-scale crowdsourced experiment and measurements with experts show that on average the extremely activating feature visualizations by \citet{olah2017feature} indeed help humans on this task ($68 \pm 4$\,\% accuracy; baseline performance without any visualizations is $60 \pm 3$\,\%). However, they do not provide any substantial advantage over other visualizations (such as e.g.\ dataset samples), which yield similar performance ($66\pm3$\,\% to $67\pm3$\,\% accuracy). 
Taken together, we propose an objective psychophysical task to quantify the benefit of unit-level interpretability methods for humans, and find no evidence that a widely-used feature visualization method provides humans with better ``causal understanding'' of unit activations than simple alternative visualizations.

\end{abstract}

\section{Introduction}\label{introduction}

\begin{figure}[h!]
\begin{center}
\includegraphics[width=\textwidth]{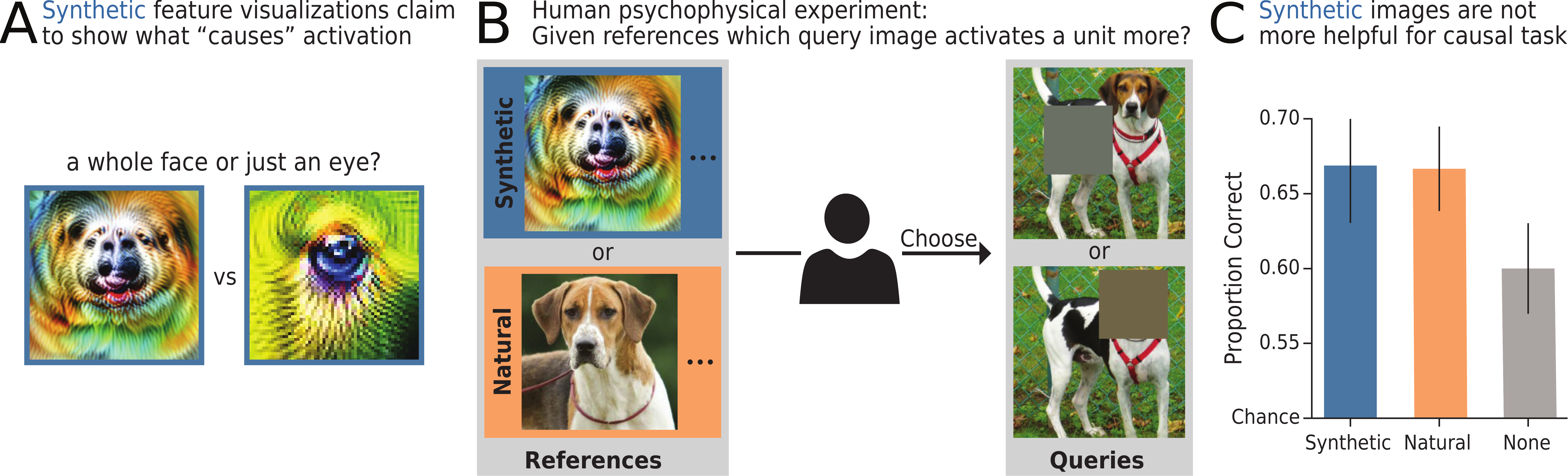}
\caption{How useful are feature visualizations to interpret the effects of interventions? \textbf{A:~``Causal'' synthetic feature visualizations.}
\textbf{B:~Human experiment.} Given strongly activating reference images (e.g. \textcolor{prussian_for_text}{synthetic} or \textcolor{coral}{natural}), a human participant chooses which out of two manipulated images activates a unit more. Note that this trial is made up --- real trials are often more difficult. \textbf{C:~Core result.} While participants are above chance for all visualization types, synthetic images only provide a substantial advantage over \emph{no} references and not over other alternatives such as natural references.
}
\label{fig:fig_1}
\end{center}
\vspace{-0.3cm}
\end{figure}

It is hard to trust a black-box algorithm, and it is hard to deploy an algorithm if one does not trust its output. Many of today's best-performing machine learning models, deep convolutional neural networks (CNNs), are also among the most mysterious ones with regards to their internal information processing. CNNs typically consist of dozens of layers with hundreds or thousands of units that distributively process and aggregate information until they reach their final decision at the topmost layer. Shedding light onto the inner workings of deep convolutional neural networks has been a long-standing quest that has so far produced more questions than answers.

One of the most popular tools for explaining the behavior of individual network units is to visualize unit responses via activation maximization \citep{erhan2009visualizing, mahendran2015understanding, nguyen2015deep, mordvintsev2015inceptionism, nguyen2016synthesizing, nguyen2017plug, tsipras2018robustness, engstrom2019adversarial}.
The idea is to start with an image (typically random noise) and iteratively change pixel values to maximize the activation of a particular network unit via gradient ascent.
The resulting synthetic images, called \emph{feature visualizations}, often show interpretable structures, and are believed to isolate and highlight exactly those features that ``cause'' a unit's response \citep{olah2017feature, schubert2021high}.
Some of the synthetic feature visualizations appear quite intuitive and precise.
As shown in Fig.~\ref{fig:fig_1}A, they might facilitate distinguishing whether, for example, a unit responds to just an eye or a whole dog's face.

However, other aspects cast a more critical light on feature visualization's ``causality'': Generating these synthetic images typically involves regularization mechanisms \citep{nguyen2017plug, mahendran2015understanding, nguyen2015deep, mordvintsev2015inceptionism}, which may influence how faithfully they visualize what ``causes'' a network unit's activation. Furthermore, to obtain a complete description of a mathematical function, one generally needs more information than just knowing its extrema. In view of this, it is an open question how well a unit can be characterized by simply visualizing the arguments of its maxima. Finally, a crucial unknown factor is whether \emph{humans} are able to obtain a causal understanding of CNN activations from these synthetic visualizations.

Given these points, we develop a psychophysical experiment to test whether feature visualizations by \citet{olah2017feature} indeed allow humans to gain a causal understanding of a unit's behavior.
Our task is based on the reasoning that being able to predict the effect of an intervention is at the heart of causal understanding.
Understanding the causal relation between variables implies an understanding of how changes in one variable affect another one \citep{pearl2009causality}.
In our proposed experiment, this means that participants can predict the effect of an intervention --- in form of an image manipulation --- if they know the causal relation between image features and a unit's activations.
Our experiment tests whether synthetic feature visualizations indeed provide information about such causal relations.
Specifically, we ask humans which of two manipulated images activates a CNN unit more strongly.
The interventions we test are obtained by placing an occlusion patch at two different locations in an image.
Taken together, this experiment probes the purported explanation method's advantage of causality in a counterfactual-inspired prediction set-up \citep{doshi2017towards}.

Besides feature visualizations, other visualization methods have been used to gain an understanding of the inner workings of CNNs.
In this experiment, we additionally test alternatives based on natural dataset examples and compare them with feature visualizations.
This is particularly interesting because dataset examples are often assumed to provide less ``causal'' information about a unit's response as they might contain misleading correlations \citep{olah2017feature}.
To continue the example above, dog eyes usually co-occur with dog faces; thus, separating the influence of one image feature from the other one using natural exemplars might be challenging.

Our data shows that:
\begin{itemize}[leftmargin=*]
    \item Synthetic feature visualizations provide humans with some helpful information about the most important patch in an image --- but not much more information than no visualizations at all.
    \item Dataset samples as well as other combinations and types of visualizations are similarly helpful.
    \item How easily the most important patch is identifiable depends on the unit, the images as well as the relative activation strength attributed to the patch.
\end{itemize}

\section{Related Work}\label{related_work}

\textbf{Feature visualizations} are a widely used method to understand the learned representations and decision-making mechanisms of CNNs \citep{mahendran2015understanding, nguyen2015deep, mordvintsev2015inceptionism, nguyen2016synthesizing, nguyen2017plug, tsipras2018robustness, engstrom2019adversarial, olah2017feature, nguyen2019understanding}. As such, several works leverage this method to study InceptionV1 \citep{olah2020zoom, olah2020an, cammarata2020curve, olah2020naturally, schubert2021high, cammarata2021curve, voss2021visualizing, voss2021branch, petrov2021weight} and other networks \citep{cadena2018diverse, gonthier2020analysis, goh2021multimodal}; others create interactive tools \citep{wong2021leveraging, OpenAIMi10:online, sietzen2021interactive} or introduce analysis frameworks \citep{zaeem2021cause}. In contrast, some researchers question whether this synthetic visualization technique, first introduced by \citet{erhan2009visualizing}, is too intuition-driven \citep{leavitt2020towards}, and how representative the appealing visualizations in publications are \citep{kriegeskorte2015deep}.
Further, as already mentioned above, the engineering of the loss function may influence their faithfulness \citep{nguyen2017plug, mahendran2015understanding, nguyen2015deep, mordvintsev2015inceptionism}. Another challenge is generating \emph{diverse} feature visualizations to represent the different aspects that one single unit may respond to \citep{olah2020zoom, nguyen2017plug}.
Finally, our recent human evaluation study \citep{borowskiandzimmermann2020exemplary} found that while these synthetic images do provide humans with helpful information in a forward simulation-inspired task, simple natural dataset examples are even more helpful.

\smallskip
\textbf{Human evaluation studies} are extensively used to quantify various aspects of interpretability. As an alternative to pure mathematical approximations \citep{amparore2021trust, zhou2021feature, van2019interpretable, ye2021evaluating}, researchers not only evaluate the understandability of explanation methods in psychophysical studies \citep{cai2019effects, mohseni2021quantitative, borowskiandzimmermann2020exemplary}, but also trust in these methods \citep{lim2009why, yin2019understanding}) as well as the human cognitive load necessary for parsing explanations \citep{abdul2021COGAM} or whether humans would follow an explained model decision \citep{poursabzi2018manipulating, diprose2020physician, prasad2020extent}. A recent study even demonstrates that metrics of the explanation quality computed \emph{with} human judgment are more insightful than those without \citep{biessmann2019psychophysics}.

\begin{wrapfigure}{r}{0.70\textwidth}
    \vspace{-1.4\intextsep}
    \begin{center}
    \includegraphics[width=0.69\textwidth]{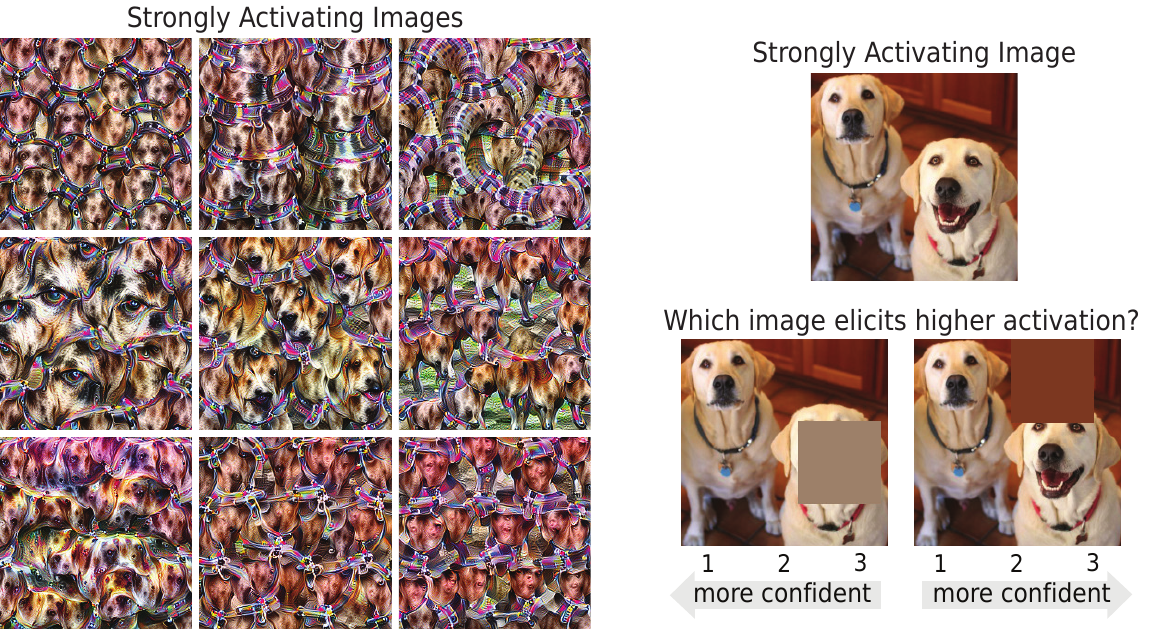}
    \end{center}
    
   \caption{
    \textbf{Schematic visualization of an example trial} in our psychophysical experiment. For a certain network unit, participants are shown several maximally activating images. While the ones on the left serve as reference images, the ones on the right serve as query images: The top one is a natural maximally activating image and the bottom ones are copies of said image with square occlusions at different locations. The task is to select the image that activates the given network unit more strongly.
    Participants answer by clicking on the number below the corresponding image according to the their confidence level ($1$:~not confident, $2$:~somewhat confident, $3$:~very confident). Correct answer: right image.}
    \label{fig:fig_2}
    \vspace{-0.5\intextsep}
\end{wrapfigure}

\smallskip
\textbf{Counterfactuals} are a popular paradigm for both \emph{creating} as well as \emph{evaluating} explanation methods. 
Intuitively, they provide answers to the question ``what should I change to achieve a different outcome?'' --- in the context of machine learning explanation methods, usually the smallest, realistic change to a data point is of interest. As examples, counterfactual explanation methods have been developed for vision- \citep{goyal2019counterfactual} and language-based \citep{wu2021polyjuice} models as well as for model-agnostic scenarios \citep{sharma2019certifai}. Further, they are set into context of the EU General Data Protection Regulation \citep{wachter2018counterfactual}. \citet{ustun2019actionable} investigate feasible and least-cost counterfactuals, while \citet{mahajan2020preserving} and \citet{karimi2020modelagnostic} take feature interactions into account. 
To \emph{evaluate} --- rather than create --- explanation methods, researchers often follow the ``counterfactual simulation'' task introduced by \citet{doshi2017towards}: Humans are given an input, an output, and an explanation and are then asked ``what must be changed to change the method's [model’s] prediction to a desired output?'' \citet{doshi2017towards}. Based on this task, \citet{lucic2020does} test their new explanation method and \citet{hase2020evaluating} compare different explanation methods to each other.

In this project, we design a counterfactual-inspired task to evaluate how well feature visualizations support causal understanding of CNN activations. This is the first study to apply such a paradigm to understanding the causes of individual units' activations. In order to scale the experiments, we simplify our task by having participants choose between two intervention \emph{options}, rather than having them freely determine interventions themselves.

\section{Methods}\label{methods}

We run an extensive psychophysical experiment with more than $12,000$ %
trials distributed over $323$ crowdsourced participants on Amazon Mechanical Turk (MTurk) and two experts (the two first authors).\footnote{Code and data are available at \href{https://github.com/brendel-group/causal-understanding-via-visualizations}{github.com/brendel-group/causal-understanding-via-visualizations}.
}
For more details than provided below, please see Appx.~Sec.~\ref{app:details_on_methods}.

\textbf{Design Principles}\hspace{0.5em}
Overall, our experimental design choices aim at (1) the \textit{best performance possible}, meaning that we select images that make the signal as clear as possible; (2) \textit{generality} over the network, meaning that we randomly sample units of different layers and branches (testing all units would be too costly); and (3) \textit{easy extendability}, meaning that we choose a between-participant design (each participant sees only one reference image condition) so that other visualizations methods can be added to the comparisons in the future.

\subsection{Psychophysical Task}\label{methods:task}
If feature visualizations indeed support causal understanding of CNN activations, this should enable humans to predict the effect of an intervention, such as how occluding an image region changes a unit's activation.
Based on this idea, we employ a two-alternative forced choice task (chance performance: $50\%$) where human observers are presented with two different occlusions in an image, and asked to estimate which of them causes a smaller change to the given unit's activation (see Fig.~\ref{fig:fig_2} for an example trial). 
More specifically, participants choose the \emph{query} image that they believe to also elicit a strong activation given a set of 9 \emph{reference} images. Such references could for instance consist of synthetic feature visualizations of a certain unit (purportedly ``causal''), or alternative visualizations.
To summarize, the task requires humans to first identify the shared aspect in the reference images and to then choose the query image in which that aspect is more visible.
Since we do not make any assumptions about whether participants are familiar with machine learning, we avoid asking participants about activations of a unit in the CNN. 
Instead, we explain that an image would be ``favored'' by a machine, and the task is to select the image which is ``more favored''. The complete set of instructions shown to participants can be found in Appx.~Fig.~\ref{fig:instructions_one}~and~\ref{fig:instructions_two}. In addition to each participant's image choice, the subjective confidence level and reaction time are also recorded.

\subsection{Stimulus Generation}\label{methods:stimulus_gen}
To generate stimuli, we follow \citet{olah2017feature} and use an InceptionV1 network \citep{szegedy2015going} trained on ImageNet \citep{imagenet_cvpr09, ILSVRC15}. Throughout this paper, we refer to a CNN's channel as a ``unit'' and imply taking the spatial average of all neurons in one channel.\footnote{Other papers might refer to a channel as a ``feature map'', e.g. \citep{borowskiandzimmermann2020exemplary}.} We test units sampled from $9$ layers and $2$ Inception module branches (namely $3 \times 3$ and \textsc{Pool}). For more details on the generation procedures of the respective stimuli, see Appx.~\ref{app:stimuli_selection}.

We use five different types of \textbf{reference images}:
\begin{itemize}[leftmargin=*]
    \item \textbf{Synthetic references}: The synthetic images are the optimization results of the feature visualization method by \citet{olah2017feature} with the channel objective for $9$ diverse images.

    \item \textbf{Natural references}: The reference images are the most strongly activating\footnote{To reduce compute requirements, we use a random subset of the training set ($\approx 50\%$).} dataset samples from ImageNet \citep{imagenet_cvpr09, ILSVRC15}.

    \item \textbf{Mixed references}: This is a combination of the previous two conditions: the $5$ most strongly activating natural and $4$ synthetic reference images are used. The motivation is that this condition combines the advantages of both worlds --- namely precise information from feature visualizations and easily understandable natural images --- and, thus, has the potential to give rise to higher performance in the task. Jointly looking at these two visualization types is common in practice \citep{olah2017feature}.

    \item  \textbf{Blurred references}: To increase the informativeness of natural images for this task, we modify them by blurring everything but a single patch. This patch is chosen in the same way as in the maximally activating query image (see below). Consequently, this method cues participants to the most important image feature. In a way, these images can be seen as an approximate inverse of the maximally activating query image and might improve performance on our task.

    \item \textbf{No references}: This is a control condition in which participants do not see any reference images and have to solve the task purely based on query images.
\end{itemize}
   
To generate \textbf{query images}, we place a square patch of $90 \times 90$ pixels of the average RGB color of the occluded pixels into a most strongly activating image chosen from ImageNet.
The location of the occlusion patch is chosen such that the activation of the manipulated image is either minimal or maximal among all possible occlusion locations. These images then yield the distractor and target query images respectively.

\subsection{Structure of the Psychophysical Experiment}\label{methods:exp_structure}

We test the five different reference image types as separate experimental conditions. In each condition, we collect data from a total of $50$ different MTurk participants, each assigned to a single Human Intelligence Task (HIT) consisting of an instruction block, a variable number of practice blocks and a main block. The instructions extensively explain a hand-crafted example trial (see Appx. Fig.~\ref{fig:instructions_one}~and~\ref{fig:instructions_two}). The blocks of $4$ practice trials each - which are randomly sampled from a pool of $10$ trials - have to be repeated until reaching $100\%$ performance; except in the none condition, as there is no obvious ground truth due to the absence of reference images. Finally, $18$ main trials follow that are randomly interleaved with a total of $3$ obvious catch trials. While feedback is provided during practice trials, no feedback is provided in the other trials. At the end, participants can share comments via an optional free-text field. Across all conditions, all participants see the same query images for the instruction, practice and catch trials. In contrast, the query images differ across participants in the main trials: In each reference image condition, we test $10$ different sets of query images, each responded to by $5$ different MTurk participants, hence $50$ HITs per condition. The order of the main and catch trials per participant is randomly arranged, and identical across conditions. Each MTurk participant takes part in only one reference image condition (i.e. reference images are a between-participants factor). For more details, see Appx. Sec.~\ref{app:trials}.

\subsection{Ensuring High-Quality Data in an Online Experiment}\label{methods:ensure_high_quality}

To ensure that the data we collect in our online experiment is of high quality, we take two measures: 
(1) We integrate hidden checks which were set before data collection. Only if a participant passes all five of them do we include his/her data in our analysis.
First, these \emph{exclusion criteria} comprise a performance threshold on the practice trials as well as a maximum number of blocks a participant may attempt. Further, they include a performance threshold for catch trials, a minimum image choice variability as well as a minimum time spent on both the instructions and the whole experiment.
For more details, see Appx. Sec.~\ref{app:data_collection}.
(2) Our previous human evaluation study in a well-controlled lab environment found that natural reference images are more informative than synthetic feature visualizations when choosing which of two different images is more highly activating for a given unit \citep{borowskiandzimmermann2020exemplary}. 
We replicate this main finding on MTurk based on a subset of the originally tested units (see Appx.~\ref{app:replication}) which indicates that the experiment's environment does not influence this task's outcome. Our decision to leverage a crowdsourcing platform is further corroborated by our result in \citet{borowskiandzimmermann2020exemplary}, that there is no significant difference between expert and lay performance.

\subsection{Baselines}\label{methods:baselines}

In order to both set MTurk participants' performance into context as well as evaluate different strategies participants could use to perform our task, we further evaluate a few baselines. 

\begin{itemize}[leftmargin=*]

    \item \textbf{Expert Baseline}: The two first authors answer all $18$ trials in all $5$ reference conditions on all $10$ image sets. As they are familiar with the task design and are certainly engaged, this data serves as an upper human bound.

    \item \textbf{Center Baseline}: In natural images from ImageNet, important objects are likely to be closer to the center of the image.
    If participants were biased to assume that units respond to \emph{objects}, a potential strategy to decide which occluding patch produces a smaller effect on the unit's activation would therefore be to choose the image with the most eccentric occlusion. The Center Baseline model performs this strategy for all images. 
    
    \item \textbf{Primary Object Baseline}: The Center Baseline is not a perfect measurement of an object-biased strategy because primary objects can appear away from the center. To account for this, the two first authors and the last author manually label all trials, choosing the image for which the occlusion hides as little information as possible from the most prominent object in the scene. In approximately one third of the trials ($58/180$), the authors' confidence ratings are very low (reflecting e.g. the absence of a primary object); in these cases we repeatedly replace the decisions by random binomial choices. Thus, in the results, we report the estimated expected values, but cannot perform a by-trial analysis.
    For more details, see Appx. Sec.~\ref{app:baselines}.
    
    \item \textbf{Variance Baseline}: Another assumption participants might make is that a patch in a low-contrast region, e.g. a blue sky, is unlikely to have a large effect on the unit's activation.
    This baseline selects the query image whose content is less affected by the introduction of the occlusion patch. To simulate this, we calculate the standard deviation over the occluded pixels and choose the one of the lower standard deviation.

    \item \textbf{Saliency Baseline}: As a complement to the baselines above, this baseline selects the query image whose original pixels hidden by the occlusion patch have a lower probability of being looked at by the participants. This simulates that participants select the image with a patch that occludes less prominent information and is estimated with the saliency prediction model DeepGaze IIE \citep{linardos2021calibrated}.
    For more details, see Appx. Sec.~\ref{app:baselines}.
\end{itemize}

\section{Results}\label{results}

The results shown in this section are based on $7350$ \footnote{($18$ main + $3$ catch trials)$\times 50$ MTurk participants $\times 5$ conditions + ($18$ main + $3$ catch trials)$\times 20$ expert measurements $\times 5$ conditions.}
trials from MTurk participants, who passed all exclusion criteria, and experts distributed over five conditions. In all figures, \emph{Synthetic} refers to the purportedly ``causal'', activation-maximizing feature visualizations, \emph{Natural} to ImageNet samples, \emph{Mixed} to the combined presentation of synthetic and natural images, \emph{Blur} to the blurred images, and \emph{None} to the condition with no reference images at all. Further, error bars indicate two standard errors above and below the participant-mean over network units and image sets, unless stated otherwise.

\begin{wrapfigure}{r}{0.72\textwidth}
    \vspace{-1.4\intextsep}
    \begin{center}
    \includegraphics[width=0.72\textwidth]{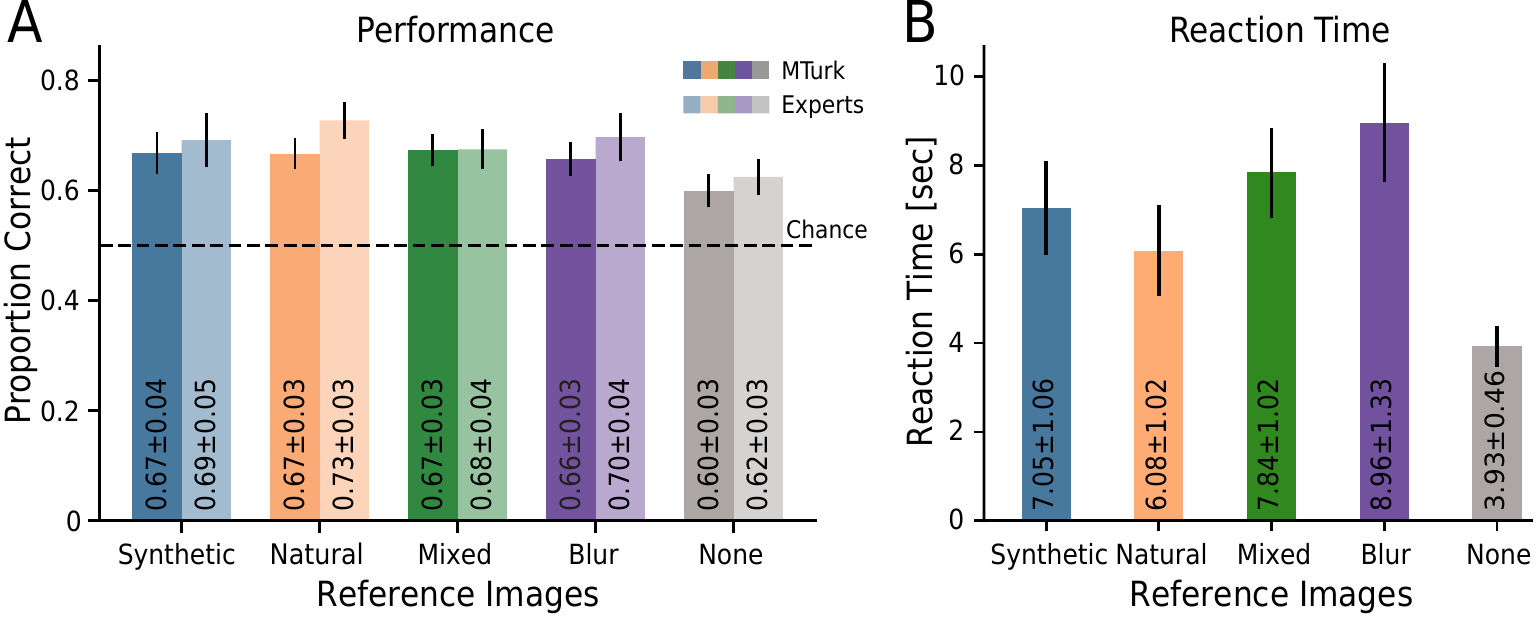}
    \end{center}
    
   \caption{
    \textbf{A: Task accuracy.} On average, humans reach the same performance regime with any visualization method.
    This holds for both lay participants on MTurk (darker colors) as well as experts (brighter colors).
    \textbf{B: Reaction times.} MTurk participants need several seconds to answer a trial, indicating that they carefully make their decision.
    For more details see Appx. Fig.~\ref{fig:apx_reaction_time}.}
    \label{fig:fig_3}
\end{wrapfigure}

\subsection{No Significant Advantage of Synthetic Feature Visualizations}
If feature visualizations provide humans with useful information about the image features causing high unit activations and other visualizations do not, participants' accuracy in our task should be higher given feature visualizations than for all other visualization types or no reference images.
This is only partly what we find: On average, accuracy for feature visualizations is slightly higher than when no reference images are given ($67 \pm 4\%$ vs. $60 \pm 3\%$). However, the accuracy for feature visualizations is not significantly higher than for other visualization methods (see Fig.~\ref{fig:fig_3}A, dark bars). For the latter, MTurk participants reach between $66 \pm 3$\,\% and $67 \pm 5$\,\% depending on the visualization type. Statistically, only the condition without reference images is different from all other conditions ($p<0.05$, Mann-Whitney U test). Taken together, these findings suggest that all visualization methods are similarly helpful for humans in our counterfactual-inspired task, and that they only seem to offer a small improvement over no visualizations at all.

\subsubsection{MTurk Participants Carefully Make Their Choices}
Similar performances for various conditions such as those found in Fig.~\ref{fig:fig_3}A might suggest that participants would not give their best when doing our experiment. However, several aspects speak against this: 
(1) Measurement of the two first authors, i.e. experts who designed and thus clearly understand the task, and certainly engage during the experiment, again show very similar performance (see Fig.~\ref{fig:fig_3}A, bright bars): This estimated upper bound is just $1-6\%$ better than MTurk participant performance. 
(2) With our strict exclusion criteria, we check for doubtful participant behavior and only include data from participants who pass all five criteria.
(3) Reaction times per trial (see Fig.~\ref{fig:fig_3}B) lie between $\approx4$\,s and $\approx9$\,s. This, as well as the fact that participants take longer for the conditions \emph{with} references than for the \emph{None} condition, suggest that they carefully make their decisions. 
(4)~Several MTurk participants' comments in an optional free-text field indicate that they engage in the task: ``[...] I did my best'', ``It was engaging'', ``interesting task''.
(5) Trial-by-trial responses between MTurk participants are more similar than expected by chance (see Fig.~\ref{fig:baseline_perf_cohens_kappa}B discussed below), which suggests that humans use the available information.

\subsubsection{Simple Baselines Can Reach the Same Above-Chance Performance Regime}

\begin{wrapfigure}{r}{0.70\textwidth}
    \vspace{-1.4\intextsep}
    \begin{center}
    \includegraphics[width=0.70\textwidth]{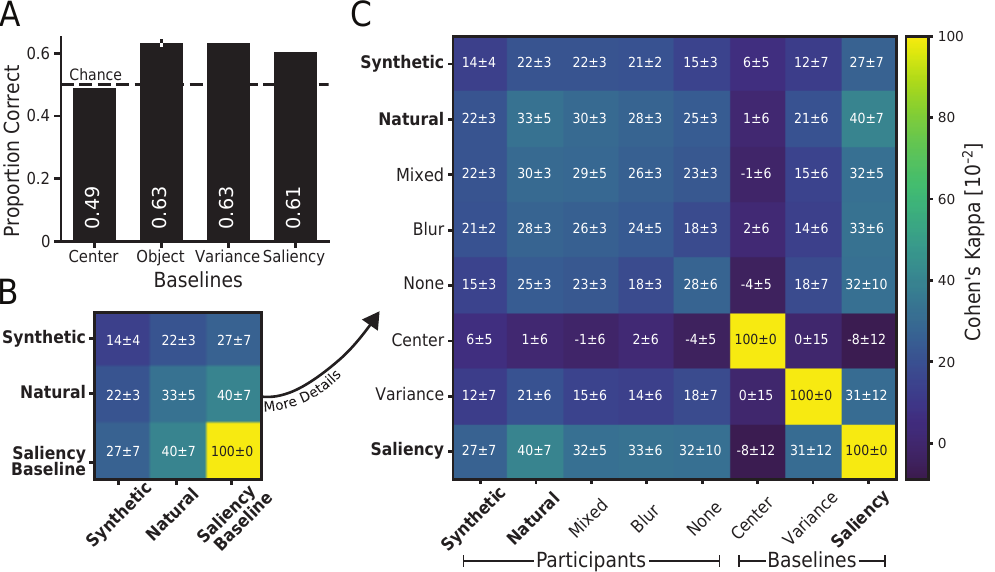}
    \end{center}
    
   \caption{
   \textbf{A: Baseline performances.} Simple baselines can reach above chance level.\protect\footnotemark\,\,
   \textbf{B, C: Decision consistency.} The mean and two standard errors of the mean of Cohen's kappa averaged over participants and image sets quantifies the pairwise consistency of decision patterns.\protect\footnotemark\,\,
   While they vary across participants, they are higher between conditions with natural references and highest between the Saliency Baseline and other conditions.
   For more details, see Appx.~Fig~\ref{app_fig:cohens_kappa_details}.}
    \label{fig:baseline_perf_cohens_kappa}
    \vspace{-0.75\intextsep}
\end{wrapfigure}

\addtocounter{footnote}{-1}
\footnotetext{Only the Object Baseline has an error bar because in trials with, e.g. no clear primary object, we replace decisions by random binomial choices. The reported values are the estimated expectation value and standard deviation.}
\stepcounter{footnote}
\footnotetext{There is no data for the Object Baseline because about one third of the trials do not have a clear answer from the three author responses. For more details, see Appx.~\ref{app:baselines}. }

Decision-making strategies can be diverse. To set human performance into context, we evaluate several simple strategies as baselines: How high is performance if one always chooses the query image with an unoccluded center (Center Baseline) or primary object (Object Baseline)? Or such that the more varying or salient image region is unoccluded (Variance and Saliency Baseline)? Fig.~\ref{fig:baseline_perf_cohens_kappa}A shows that these strategies have varying performances with the best ones --- namely the Object and Variance baselines --- reaching $63 \pm 1$\,\% and $63$\,\%, respectively. Since already these simple heuristics, which do not require reference visualizations, can reach the same performance regime as participants, the additional advantage of visualizations (reaching just up to $4$\,\% better performance) appears limited.

\subsection{By-trial Decisions Show Systematic but Fairly Low Agreement}
While accuracy is the most common metric to evaluate task performance, it does not suffice to compare two systems' decision-making processes \citep{ma2020neural, geirhos2020beyond, funke2021five}. Instead, a quantitative trial-by-trial error analysis is necessary to ascertain or distinguish strategies. Here, we use Cohen's kappa \citep{cohen1960coefficient} to calculate the degree of agreement in classification while taking the expected chance agreement into account. A value of $1$ corresponds to perfect agreement, while a value of $0$ corresponds to as much agreement as would be expected by chance. Negative values indicate systematic disagreement.

In Fig.~\ref{fig:baseline_perf_cohens_kappa}B and C, we plot consistency between MTurk participants of the same and different reference conditions as well as between MTurk participants and baselines. Since Cohen's kappa only allows for comparisons of two decision makers, we compute this statistic for all possible pairs across image sets, and report the mean over participants and image sets and two standard errors of the mean.
All values between participants as well as between participants and baselines are in an intermediate regime (up to $0.40$). This suggests that there is systematic agreement, but also quite some room for subjective decisions.
Among participant-baseline comparisons, highest agreement is found for the saliency baseline\footnote{From a different perspective, this result can be seen as a confirmation that the CNN learned to look at the ``important'' part of the image for downstream classification.}
, while lowest agreement is found for the Center Baseline.
Within participant to participant comparisons, decision strategies for conditions involving unmodified natural images (\emph{Natural}, \emph{Mixed}) are more similar to each other as well as slightly more similar to other strategies than the \emph{Synthetic}, \emph{Blur} or \emph{None} condition to other strategies.
Within the \emph{Synthetic} condition, participants are relatively inconsistent. We hypothesize that due to the fact that humans are more familiar with natural images, they use more consistent information from these types of reference images and, thus, their decisions are more similar.

\subsection{Performance Varies across Units, Image Sets and Activation Differences, but Less So for Reference Conditions}
Having found that feature visualizations do not offer an overall advantage over other techniques, we now ask: Is performance similar across units, query images and their activation differences?

\begin{figure}[h!]
    \begin{center}
        \includegraphics[width=\textwidth]{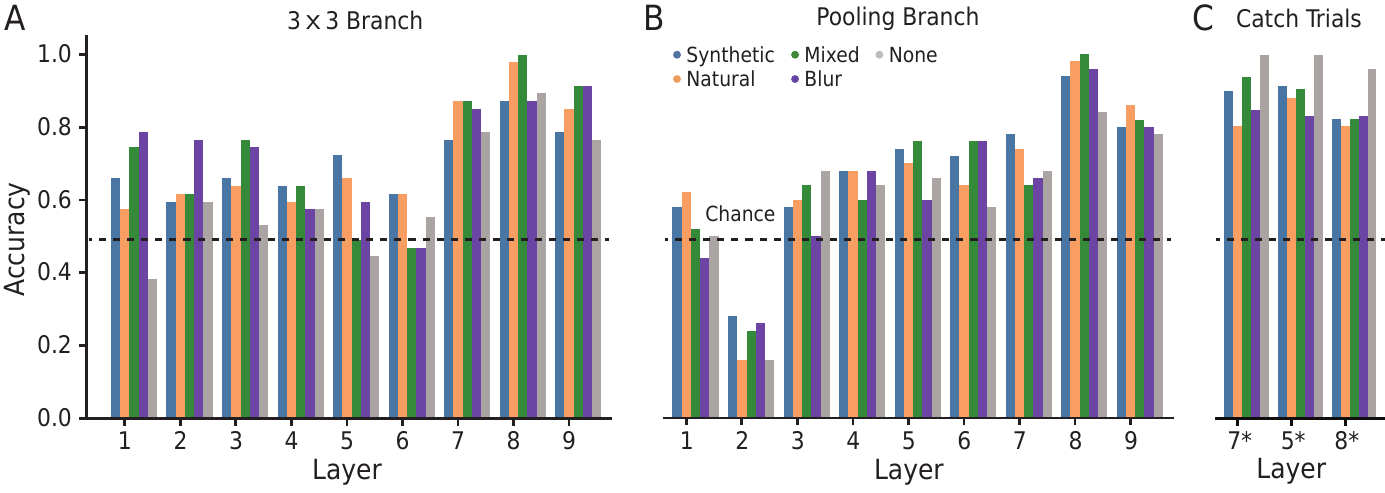}
        \caption{While for some units predicting the effect of an intervention is relatively easy, for most units performance is close to or just above chance. %
        \textbf{A} and \textbf{B} show the \textbf{performance per unit} in the main trials separated by branch ($3 \times 3$ and \textsc{Pool} respectively) and layer. 
        \textbf{C} shows the performance per unit in the hand-picked trials used as catch trials (hence the *), though selected from those MTurk participants who pass the exclusion criteria without the catch trial exclusion criterion.
        Note that each bar represents averages over participants and image sets.
        }
        \label{fig:accuracy_vs_layers}
    \end{center}
\end{figure}

\paragraph{Units and Image Sets}
As evident from Fig.~\ref{fig:accuracy_vs_layers}, performance varies by unit, but usually not much by reference condition: While only one unit (layer $2$, \textsc{Pool}) is clearly below chance level, many units reach around average performance and a few units stand out with high performances (e.g. layer $8$, \textsc{Pool}). Further, the five reference conditions are relatively close to each other for most units.
Finally, on the image set level, we observe fairly high variance - probably partly due to the limited number of participants per image set (see Appx.~Fig.~\ref{fig:performance_per_image_set}). 

\begin{figure}[h!]
    \begin{center}
        \includegraphics[width=\textwidth]{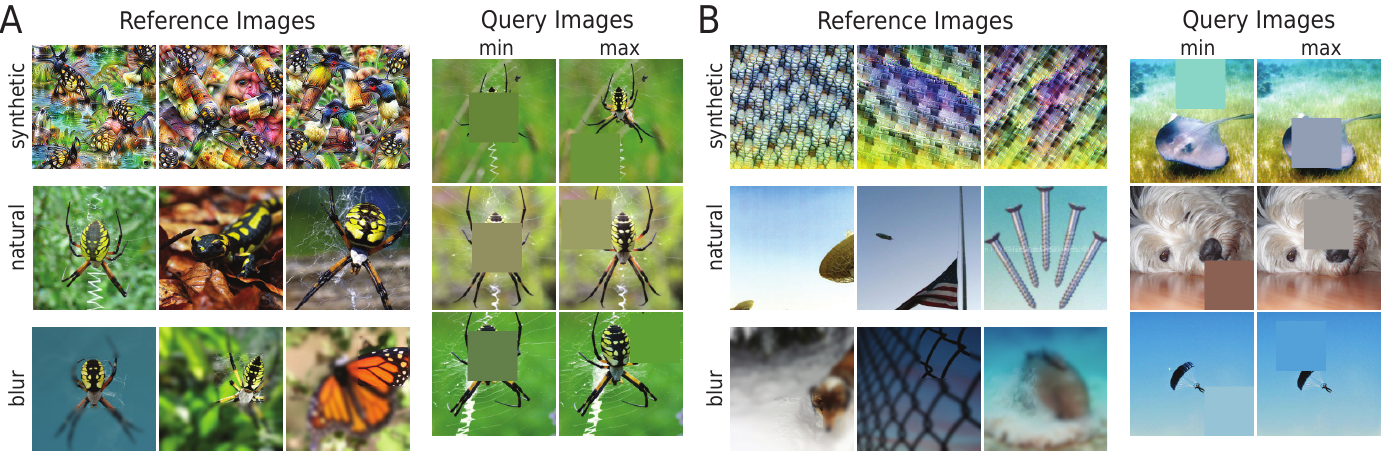}
        \caption{\textbf{Example reference and query images} for a unit with high (\textbf{A}) and low (\textbf{B}) performance from layer $8$ and $2$ of the \textsc{Pool} branch, respectively. }
        \label{fig:easy_hard_query_reference_images}
    \end{center}
\end{figure}

Fig.~\ref{fig:easy_hard_query_reference_images} further illustrates the different difficulty levels as well as the strong unit- and image-dependency: For the shown easy unit (Fig.~\ref{fig:easy_hard_query_reference_images}A), the (presumably yellow-black) feature is fairly clearly identifiable and visible in the diverse reference and query images. In contrast, for the shown difficult unit (Fig.~\ref{fig:easy_hard_query_reference_images}B), the unit's feature selectivity is unclear not only in the reference but also in the query images.%

\vspace{-0.3cm}

\begin{figure}[h!]
    \begin{center}
        \includegraphics[width=\textwidth]{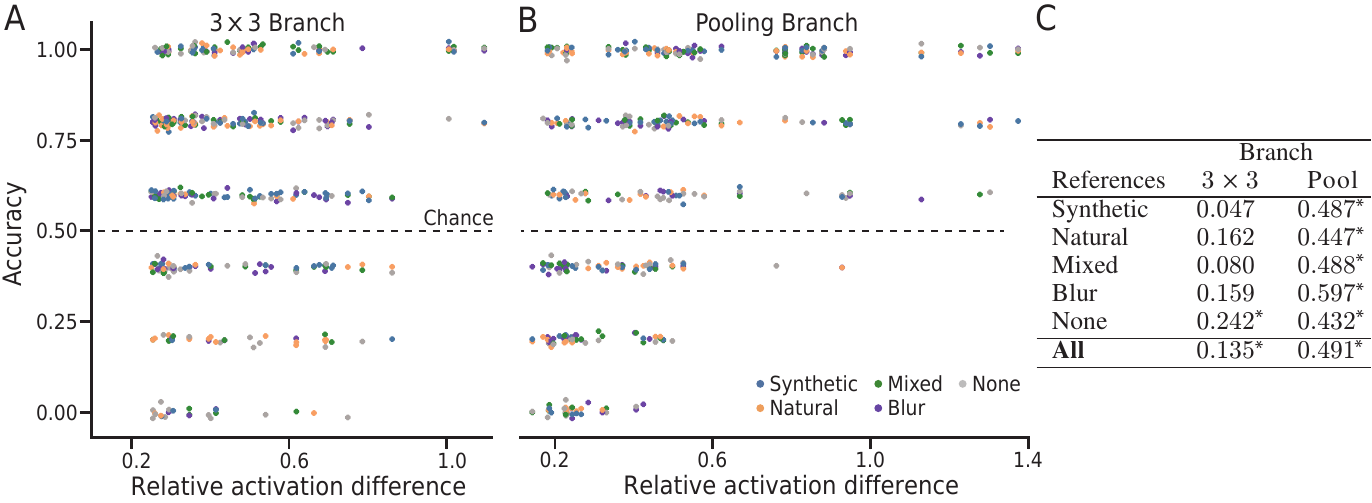}
        \caption{Performance tends to increase with the relative activation difference between query images. This effect is stronger for the \textsc{Pool} branch (\textbf{B}) than for the $3 \times 3$ branch (\textbf{A}) as quantified by Spearman's rank correlations (\textbf{C}). Stars signal significance ($p<.05$).
        Note that each dot in A and B represents the participant-averages, i.e. there is one dot per combination of layers, branch and image set. For an alternative visualization see Appx.~Fig.~\ref{fig:apx_accuracy_vs_activation_difference}.
        }
        \label{fig:accuracy_vs_relative_activations}
    \end{center}
\end{figure}

\paragraph{Activation Differences}
We hypothesize that our task might be easier if the difference in activations between the two interventions of the query images is larger.
In Fig.~\ref{fig:accuracy_vs_relative_activations}A and B, we plot by-image-set performance against the relative activation differences, i.e. the difference between activations elicited by the two manipulated images normalized by the unperturbed query image's activation.
The figure shows that even though we select query images as the most strongly activating images for a unit, the relative activation differences vary widely.
Furthermore, human performance indeed tends to increase with higher relative activation difference, confirming our hypothesis. This trend is stronger in the \textsc{Pool} than in the $3 \times 3$ branch as quantified by the Spearman's rank correlations in  Fig.~\ref{fig:accuracy_vs_relative_activations}C.

\section{Discussion \& Conclusions}\label{discussion_and_conclusions}

Explanation methods such as feature visualizations have been criticized as intuition-driven \citep{leavitt2020towards}, and it is unclear whether they allow humans to gain a precise understanding of which image features ``cause'' high activation in a unit.
Here, we propose an objective psychophysical task to quantify how well these synthetic images support causal understanding of CNN units.
Through a time- and cost-intensive evaluation (based on $24,439$ trials taking more than $81$ participant hours including all pilot and reported experiments), we put this widespread intuition to a quantitative test. Our data provides no evidence that humans can predict the effect of an image intervention (occlusion) particularly well when supported with feature visualizations.
Instead, human performance is only moderately above a baseline condition where humans are not shown any visualization at all, and similar to that of other visualization methods such as simple dataset samples.
Further, by-trial decisions show systematic but fairly low agreement between participants.
Finally, task performance depends on the unit choice, image selections and activation differences between query images.
These results add quantitative evidence against the generally-assumed usefulness of feature visualizations for understanding the causes of CNN unit activations.

Our counterfactual-inspired task is the \emph{first} quantitative evaluation of whether feature visualizations support causal understanding of unit activations, but it is certainly not the \emph{only} possible way to evaluate causal understanding.
For example, our interventions are constrained to occlusions of a fixed size and shape, imposing an upper limit on the precision with which the occlusions can cover the part of the image that is most responsible for driving a unit's activation.
Future work could explore more complex intervention techniques, extend our study to more units of InceptionV1 as well as to different networks, and investigate additional visualization methods. 
Thanks to the between-participant design, new conditions can be added to the data without the requirement to re-run already collected trials. 

Taken together, the empirical results of our quantitative evaluation method indicate that the widely used visualization method by \citet{olah2017feature} does not provide causal understanding of CNN activations beyond what can be obtained from much simpler baselines.
This finding is contrary to wide-spread community intuition and reinforces the importance of testing falsifiable hypotheses in the field of interpretable artificial intelligence \citep{leavitt2020towards}.
With increasing societal applications of machine learning, the importance of feature visualizations and interpretable machine learning methods is likely to continue to increase. 
Therefore, it is important to develop an understanding of what we can --- and cannot --- expect from explainability methods. We think that human benchmarks, like the one presented in this study, help to expose a precise notion of interpretability that is quantitatively measurable and comparable to competing methods or baselines. 
The paradigm we developed in this work can be easily adapted to account for other notions of causality and, more generally, interpretability as well.
For the future, we hope that our task will serve as a challenging test case to steer further development of feature visualizations.

\subsection*{Author Contributions}
{\footnotesize The idea to test how well feature visualizations support causal understanding of CNN activations was born out of several reviewer and audience comments on our previous paper \citep{borowskiandzimmermann2020exemplary}. The first idea of how to test this in a psychophysical experiment came from TSAW.
JB led the project.
JB, RSZ, WB and TSAW jointly improved the experimental set-up with input from MB and RG.
RSZ led and JB helped with the implementation and execution of the experiment; JB led and RSZ contributed to the generation of stimuli.
RSZ and JB both coded the baselines, and TSAW guided the replication experiment with statistical power simulations.
The data analysis was performed by RSZ and JB with advice and feedback from RG, TSAW, WB and MB. TSAW and WB provided day-to-day supervision.
While JB and RSZ created the first draft of the manuscript, RG and TSAW heavily edited the manuscript and all authors contributed to the final version.}

\subsection*{Acknowledgments}
{\footnotesize We thank Felix A. Wichmann and Isabel Valera for a helpful discussion. We further thank Ludwig Schubert for information on technical details via \url{slack.distill.pub}. In addition, we thank our colleagues for helpful discussions, and especially Matthias K\"ummerer, Dylan Paiton, Wolfram Barfuss, and Matthias Tangemann for valuable feedback on our task, and/or technical support. 
Moreover, we thank our various reviewers and other researchers for comments on our previous paper inspiring us to investigate causal understanding of visualization methods.
And finally, we thank all our participants for taking part in our experiments.}

\subsection*{Funding}
{\footnotesize The authors thank the International Max Planck Research School for Intelligent Systems (IMPRS-IS) for supporting JB, RSZ and RG.
This work was supported by the German Federal Ministry of Education and Research (BMBF) through the Competence Center for Machine Learning (TUE.AI, FKZ 01IS18039A) and the Bernstein Computational Neuroscience Program T\"ubingen (FKZ 01GQ1002), the Cluster of Excellence Machine Learning: New Perspectives for Sciences (EXC2064/1), and the German Research Foundation (DFG, SFB 1233, Robust Vision: Inference Principles and Neural Mechanisms, TP3, project number 276693517). MB and WB acknowledge funding from the MICrONS program of the Intelligence Advanced Research Projects Activity (IARPA) via Department of Interior/Interior Business Center (DoI/IBC) contract number D16PC00003. WB acknowledges financial support via the Emmy Noether Research Group on The Role of Strong Response Consistency for Robust and Explainable Machine Vision funded by the German Research Foundation (DFG) under grant no. BR 6382/1-1.}

\newpage
\bibliography{references}
\bibliographystyle{plainnat}

\newpage
\appendix

\section{Appendix}

\subsection{Details on Methods of Counterfactual-Inspired Experiment}\label{app:details_on_methods}

We closely follow our previous work \citep{borowskiandzimmermann2020exemplary} and hence often refer to specific sections of it in this Appendix.

\subsubsection{Data Collection}\label{app:data_collection}

\paragraph{Exclusion Criteria} In order to acquire data of high quality from MTurk, we integrate five exclusion criteria. If one or more of these criteria is not met, we post the same HIT again:
\begin{itemize}[leftmargin=*]
    \item Maximal number of attempts to reach $100\%$ performance in practice trials: 5
    \item Performance threshold for catch trials: two out of three trials have to be correctly answered
    \item Answer variability: at least one trial must be chosen from the less frequently selected side  (to discard participants who only responded with ``left'' or ``right'')
    \item Time to read the instructions: at least $20$\,s ($15$\,s in the none condition)
    \item Time for the whole experiment: at least $90$\,s and at most $900$\,s (at least $40$\,s, and at most $900$\,s in the none condition)
\end{itemize}

\paragraph{Minimize Biases} To minimize a bias to either query image, the location of the truly maximally activating query image is randomized and participants have to center their mouse cursor by pressing a centered button ``Continue'' after each trial.

\paragraph{Expert Measurements} The two first authors complete all 10 image sets in multiple conditions: At first, they label the query images for the Primary Object Baseline. Then they answer the none, synthetic or natural (counterbalanced between the two authors), mixed, and blur condition. Clicking through the trials several times means that they see identical images repeatedly.

\subsubsection{Stimulus Generation}\label{app:stimuli_selection}

\paragraph{Model} In line with previous work (e.g. \citet{borowskiandzimmermann2020exemplary, olah2017feature}), we use an Inception V1 network \citep{szegedy2015going} trained on ImageNet \citep{imagenet_cvpr09, ILSVRC15}. For more details, see Sec. A.1.2 ``Stimuli Selection - Model'' in \citet{borowskiandzimmermann2020exemplary}.

\paragraph{Natural Images as Query and Reference Images} The natural reference and query images are selected from a random subset of $599,552$ training images of the ImageNet ILSVRC 2012 dataset \citep{ILSVRC15}. For each unit, we select those images that elicit a maximal activation. More specifically, we choose the very most activating images as the query images and the next most activating images as reference images and ensure no overlap between query and references images as well as between image sets. As we follow our work published in \citet{borowskiandzimmermann2020exemplary}, please see A.1.2 for more details on the sampling procedure. In total, we generate $20$ different image sets per unit. In the presented data, we only use half of these sets.%

\paragraph{Query Images} For the query images, we use the $20$ maximally activating images for a given unit. To produce the manipulated query images, a square patch of $90\times90$ pixels is placed on the unperturbed query image. The side length of a patch corresponds to $40\%$ of a preprocessed image's side length. The position of the occlusion patch is chosen such that the manipulated image's activation for a given unit is minimal (maximal) among all possible manipulated images' activations. This maximizes the signal in the query images and means that patches of the two query images can overlap.

In a control experiment, we test whether the partial occlusions of the natural ImageNet images cause the manipulated images to lie outside the natural image distribution. If this was the case, the query images would fail to be representative of the network's activity for natural images. Here, we test how similar the response to the unperturbed and partially occluded images is. Specifically, we count how often there is an overlap of the top-5 predictions. If network activations were drastically different for the occluded than for the unperturbed images, we should find low agreement. However, we do find an agreement for $97.8,\%$ of all tested images. Therefore, the square occlusions only have a marginal effect on the network’s overall activity/predictions.

\begin{figure}
    \centering
    \includegraphics[width=0.55\linewidth]{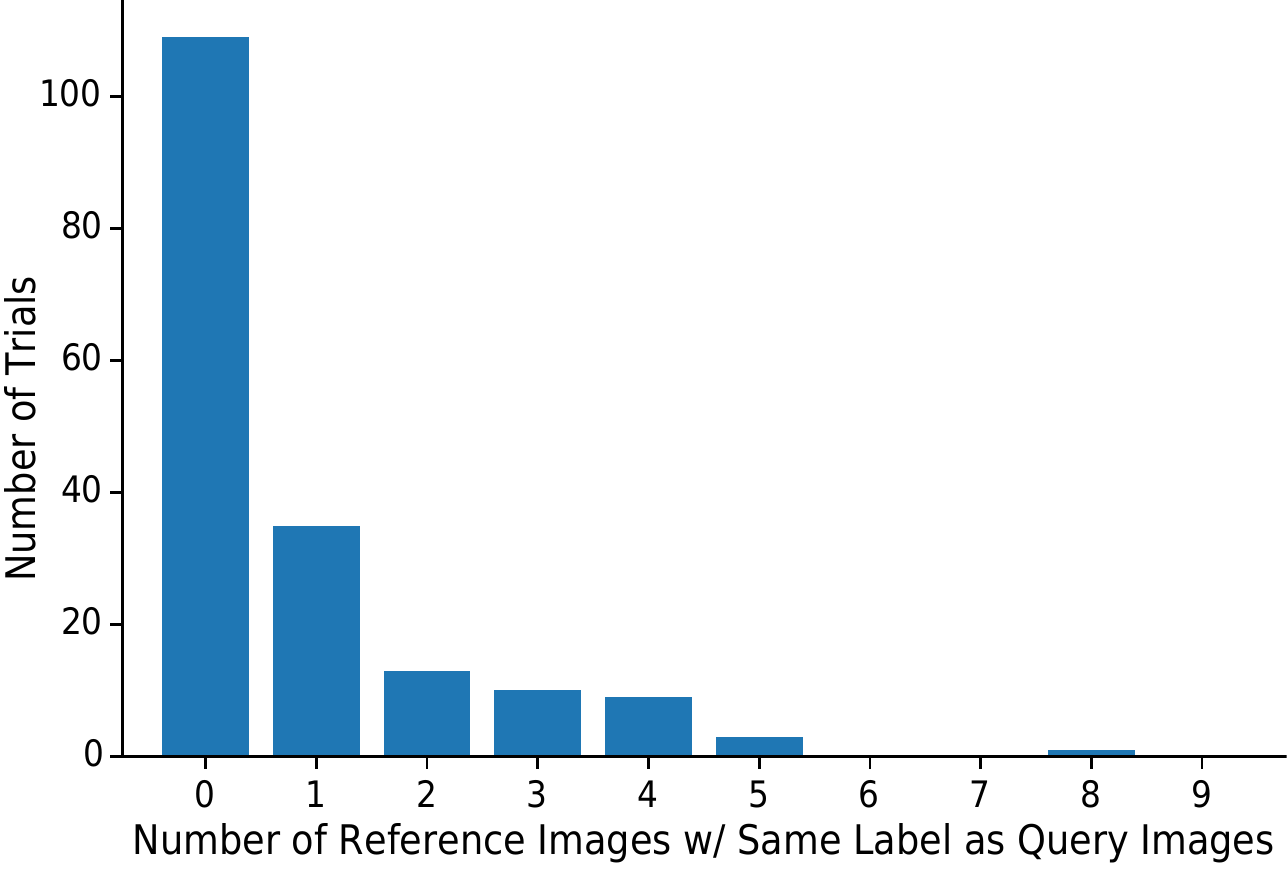}
    \caption{Distribution of the number of natural reference images that have the same label as the query image over the main trials used in the counterfactual-inspired experiment.}
    \label{fig:reference_query_label_histogram}
\end{figure}

\paragraph{Reference Images: Natural Images} In a control experiment, we test how often the label of the reference images coincide with the query image's label. If there was a high correspondence of these ImageNet labels, this could suggest that our experiment would rather reveal insights on how well humans would be able to \emph{classify} images according to \emph{labels} rather than to answer a counterfactual-inspired task based on the unit activations. Fig.~\ref{fig:reference_query_label_histogram} shows that the overlap of labels between query and reference images is low.

\paragraph{Reference Images: Blurred Images} The blurred reference images are created by blurring all but one patch with a Gaussian kernel of size $(21,21)$. This parameter choice allows participants to still get a general impression of an image, but not recognize details. Further, it is in line with work by \citet{fong2019understanding}. The image choices are identical to the natural condition. Further --- and just like for the query images --- the position of the unblurred patch is chosen such that the manipulated image's activation for a given unit is maximal among all possible manipulated images' activations. Finally, the size of the unblurred patch is identical to the occlusion patch size: $40\%$ of a preprocessed image's side length.

\paragraph{Reference Images: Synthetic Images from Feature Visualization} Depending on the condition, we adjust the number of feature visualizations we generate: For the purely synthetic condition, we generate $9$ visualizations, for the mixed condition, we generate $4$ visualizations. As we follow our work published in \citet{borowskiandzimmermann2020exemplary}, please see A.1.2 for further details.

\subsubsection{Baselines}\label{app:baselines}

\paragraph{Primary Object Baseline} The Primary Object Baseline simulates that the more strongly activating manipulated image would be the one where the occlusion hides as little as possible from the most prominent object of the query image. To this end, the first two authors and the last author label all images. When doing so, they use a slightly modified logic: They select the image whose most prominent object is \emph{most} occluded. If they cannot clearly identify a primary object in the image, they flag these trials, which are then treated differently in the analysis.
For the analysis, the image choice is inverted again to counteract the inverted task that the authors responded to.

The performance reported in Fig.~\ref{fig:baseline_perf_cohens_kappa} is calculated by averaging over the three individual performances. Each individual performance itself is in turn estimated as the expectation value over random sampling for query images with no clear primary object. This analysis is in line with how the performance of MTurk participants is analyzed. An alternative option would be to take the majority vote of the three answers. When randomly sampling the choice for query images with no clear primary object, taking the majority votes and evaluating the expected accuracy, the performance would evaluate to $0.70 \pm 0.02$. Notably, $58$ of all $180$ trials are affected by the sampling as two or more authors responded with a confidence of $1$ in $36$ trials, and one author responded with a confidence of $1$ while the other two gave opposing answers in $22$ trials. This represents a fairly large fraction and reflects that many images on ImageNet have more than one prominent object \citep{tsipras2020imagenet, beyer2020we}. Consequently, there may not be a ground-truth for each trial in the Primary Object Baseline.

\paragraph{Saliency Baseline} The Saliency Baseline simulates that participants select the image with a patch occluding the less prominent image region. To this end, we pass the unoccluded query image through the saliency prediction model DeepGaze IIE \cite{linardos2021calibrated} which yields a probability density over the entire image. Next, we integrate said density over each of the two square patches. We then select the image with a lower value indicating that less important information is hidden by the occlusion patch.

\subsubsection{Trials}\label{app:trials}

\paragraph{Main trials} For both the $3\times3$ and the \textsc{Pool} branch of each of the $9$ layers with an Inception module, one randomly chosen unit is tested (see Table~\ref{tab:main_feature_maps_experiment}). These are the same units as in Experiment~I of \citet{borowskiandzimmermann2020exemplary}.

\begin{table}[ht]
    \caption{Units used as main trials in the $3 \times 3$ as well as the \textsc{Pool} branch in the counterfactual-inspired experiment. The numbers in brackets after each layer's name correspond to the numbering used in all our plots.}
    \label{tab:main_feature_maps_experiment}
    \vspace{0.2cm}
    \centering
    \begin{tabular}[t]{ccc}
        \toprule
        & \multicolumn{2}{c}{Unit} \\
        \cmidrule(r){2-3}
        Layer & $3\times3$ & \textsc{Pool} \\
        \midrule
        mixed3a (1)  & 189 & 227 \\
        mixed3b (2) & 178 & 430 \\
        mixed4a (3) & 257 & 486 \\
        mixed4b (4) & 339 & 491 \\
        mixed4c (5) & 247 & 496 \\
        mixed4d (6) & 342 & 483 \\
        mixed4e (7) & 524 & 816 \\
        mixed5a (8) & 278 & 743 \\
        mixed5b (9) & 684 & 1007 \\
        \bottomrule
    \end{tabular}
\end{table}

\paragraph{Instruction, Practice and Catch Trials}
The instruction, practice and catch trials are hand-picked by the two first authors. As a pool of units, the appendix overview of \citet{olah2017feature} as well as the ``interpretable'' \textsc{Pool} units used in Experiment~I and all units used in Experiment~II of \citet{borowskiandzimmermann2020exemplary} are used. After generating all $20$ reference and query image sets for these units, the authors select those units and image sets that they consider easiest (see Table~\ref{tab:feature_maps_instruction_catch_practice}).

\begin{figure}[h!]
    \begin{center}
    \begin{subfigure}[b]{0.45\textwidth}
            \includegraphics[trim=70 35 70 20, clip, width=\textwidth]{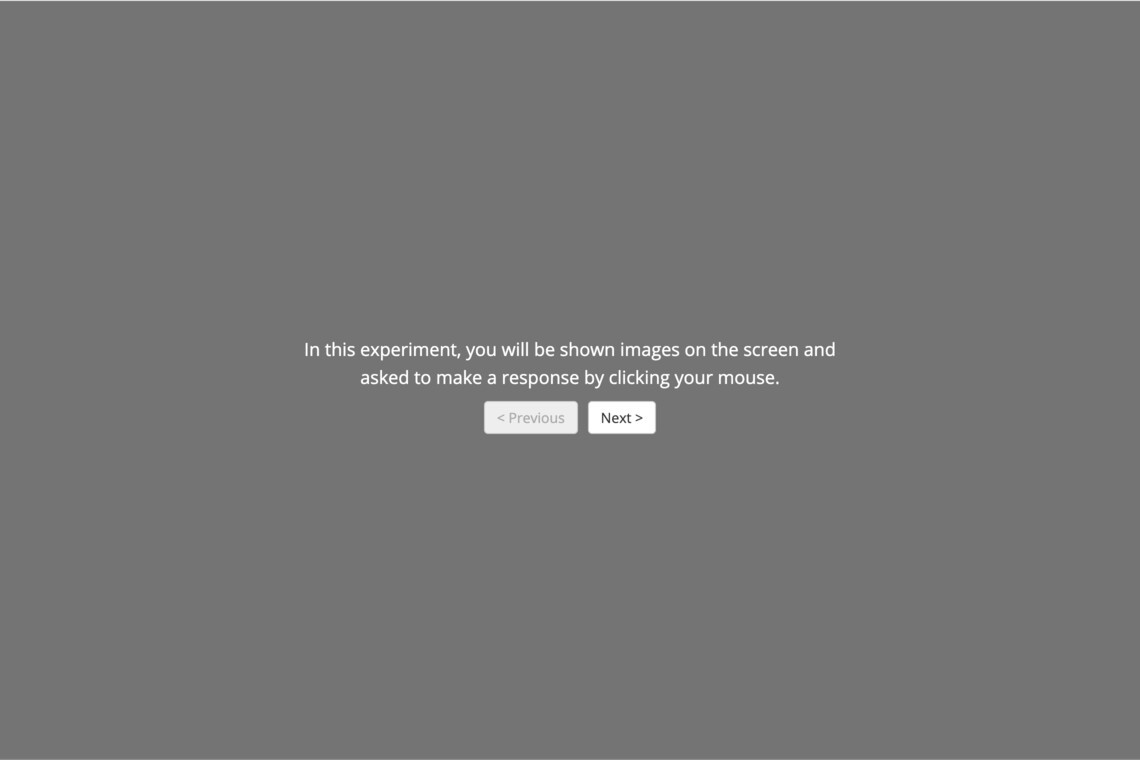}
    \end{subfigure}
    \begin{subfigure}[b]{0.45\textwidth}
            \includegraphics[trim=70 35 70 20, clip, width=\textwidth]{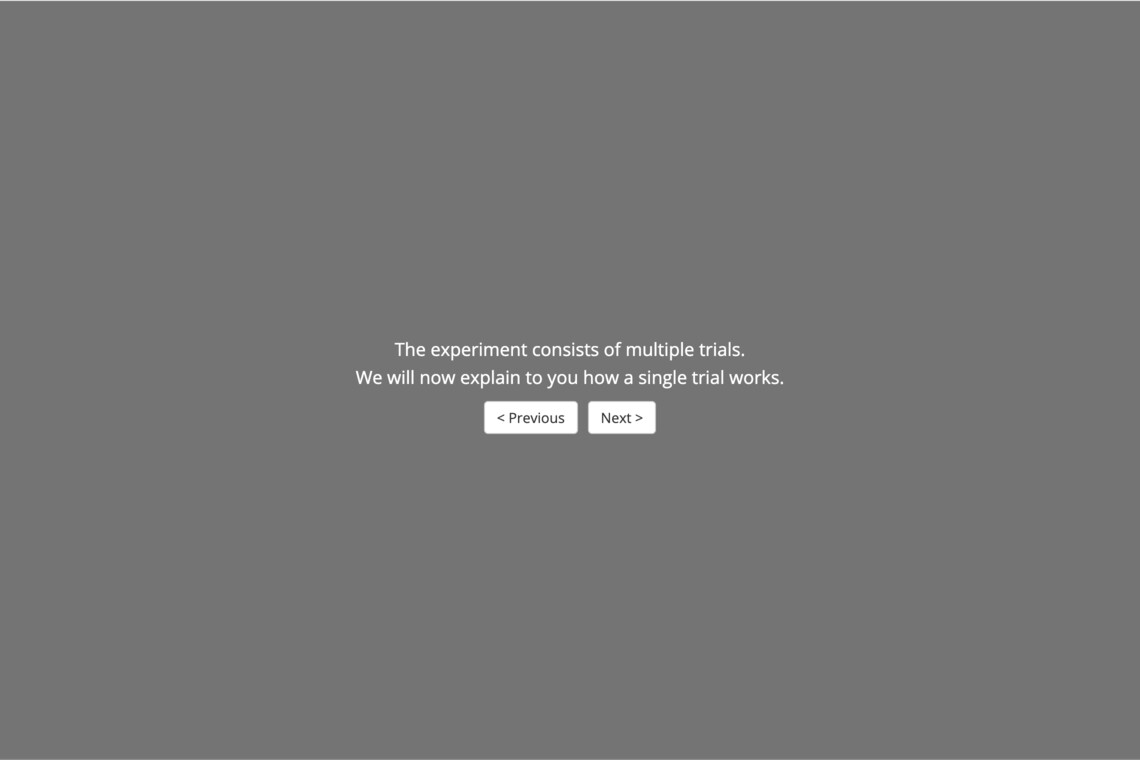}
    \end{subfigure}
    
    \begin{subfigure}[b]{0.45\textwidth}
            \includegraphics[trim=70 35 70 20, clip, width=\textwidth]{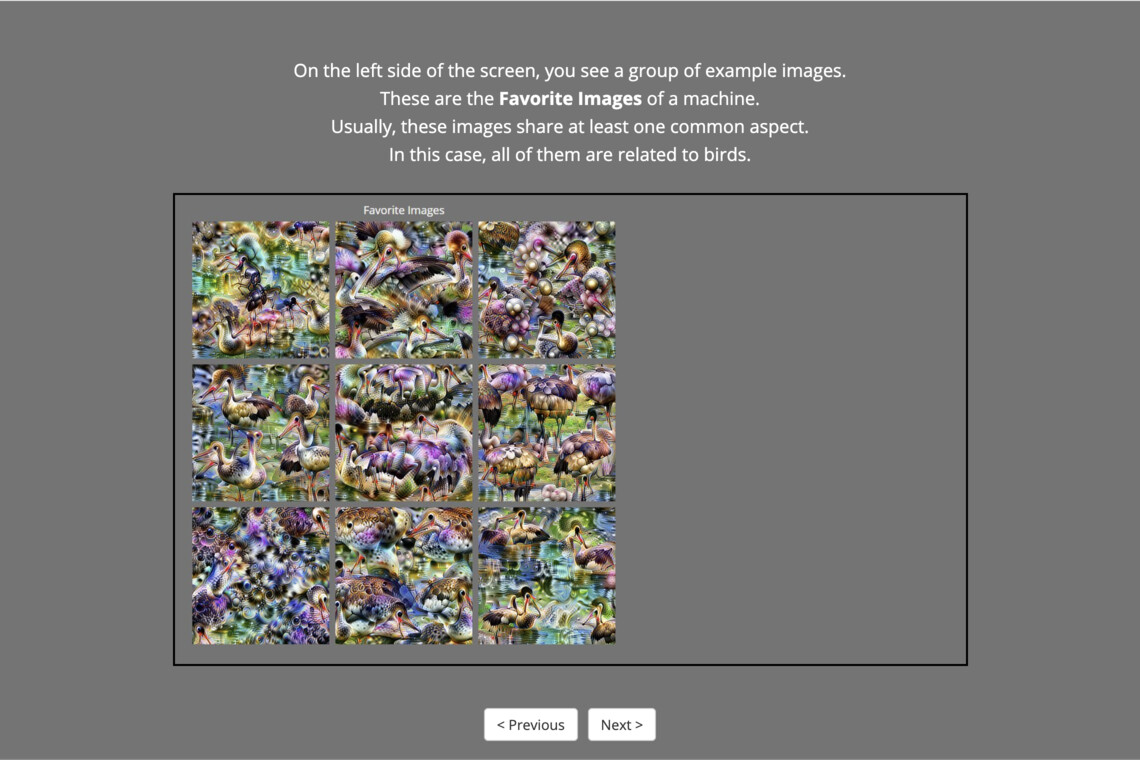}
    \end{subfigure}
    \begin{subfigure}[b]{0.45\textwidth}
            \includegraphics[trim=70 35 70 20, clip, width=\textwidth]{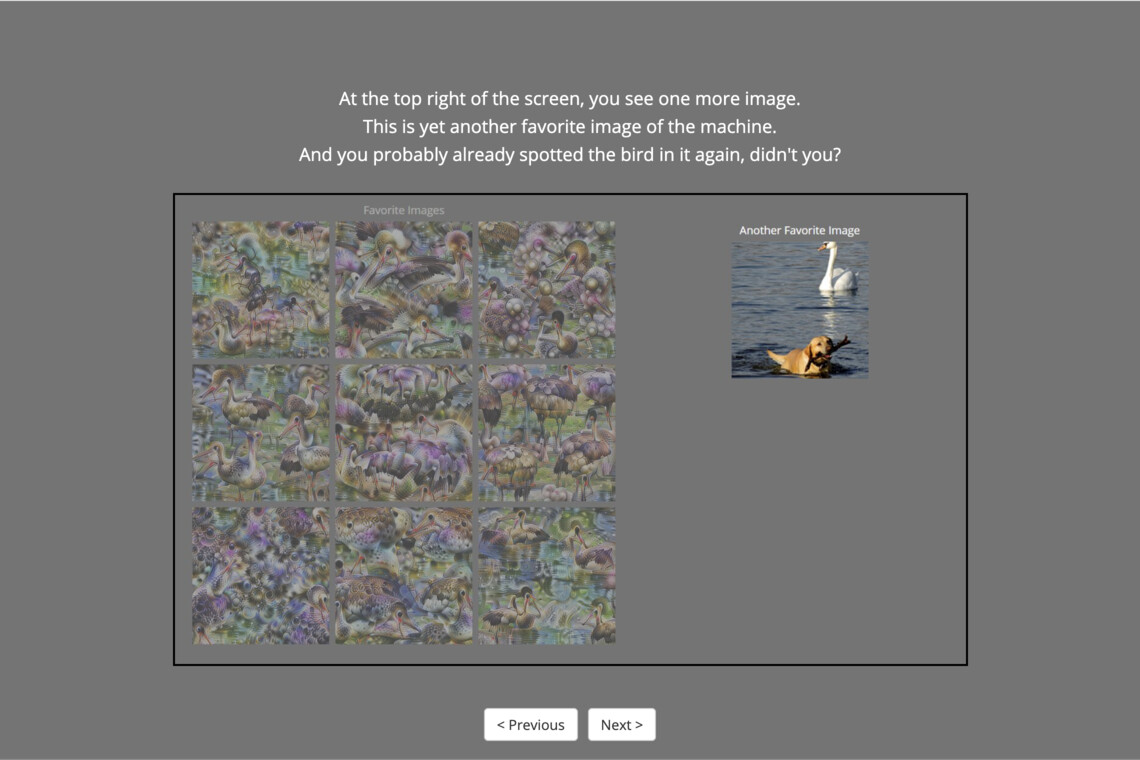}
    \end{subfigure}
    
    \begin{subfigure}[b]{0.45\textwidth}
            \includegraphics[trim=70 35 70 20, clip, width=\textwidth]{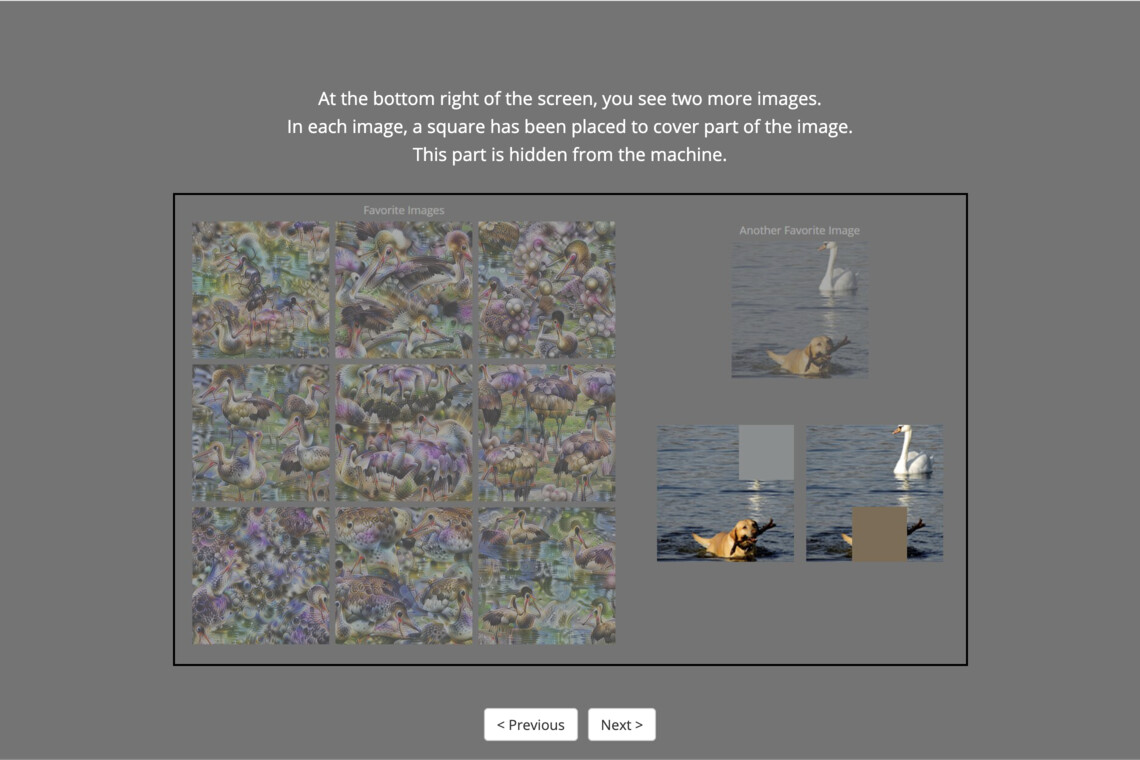}
    \end{subfigure}
    \begin{subfigure}[b]{0.45\textwidth}
            \includegraphics[trim=70 35 70 20, clip, width=\textwidth]{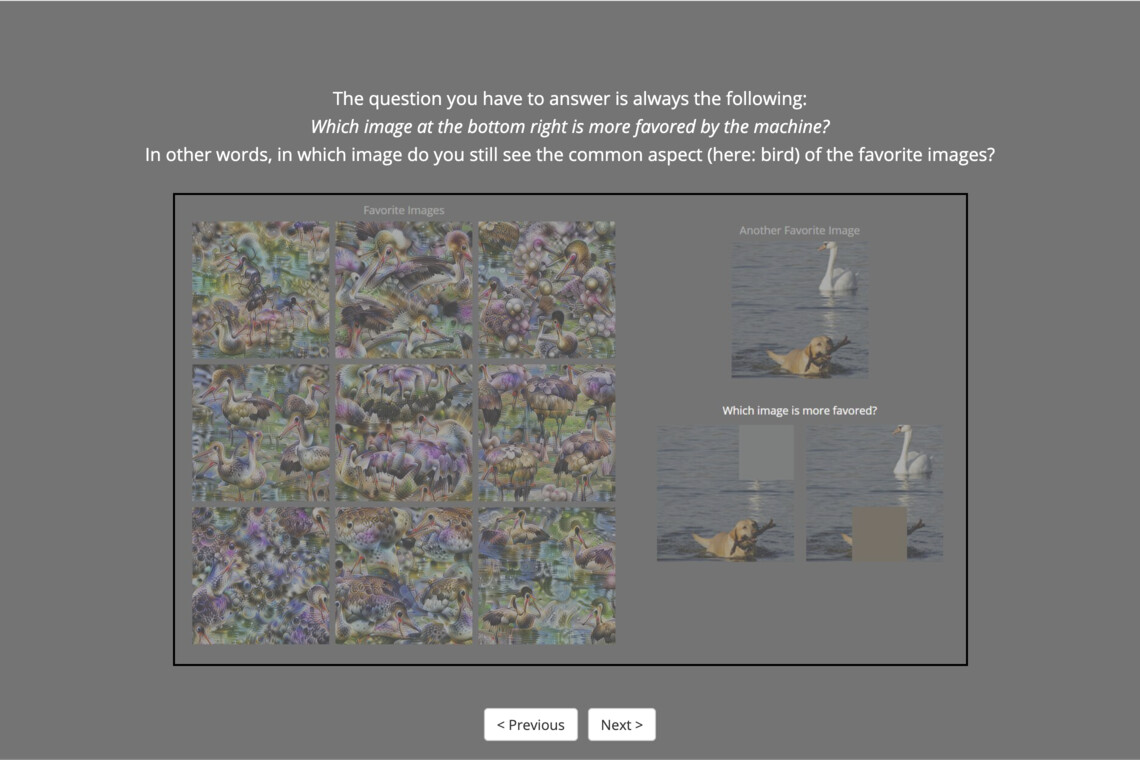}
    \end{subfigure}
    
    \begin{subfigure}[b]{0.45\textwidth}
            \includegraphics[trim=70 35 70 20, clip, width=\textwidth]{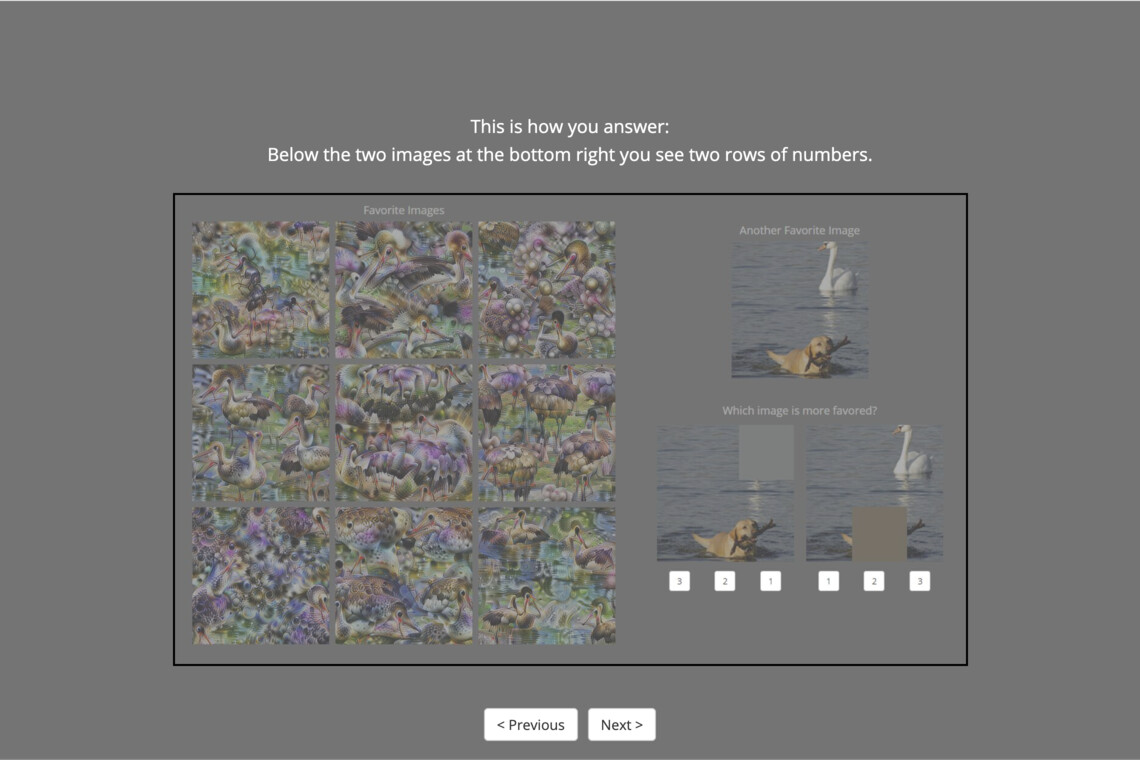}
    \end{subfigure}
    \begin{subfigure}[b]{0.45\textwidth}
            \includegraphics[trim=70 35 70 20, clip, width=\textwidth]{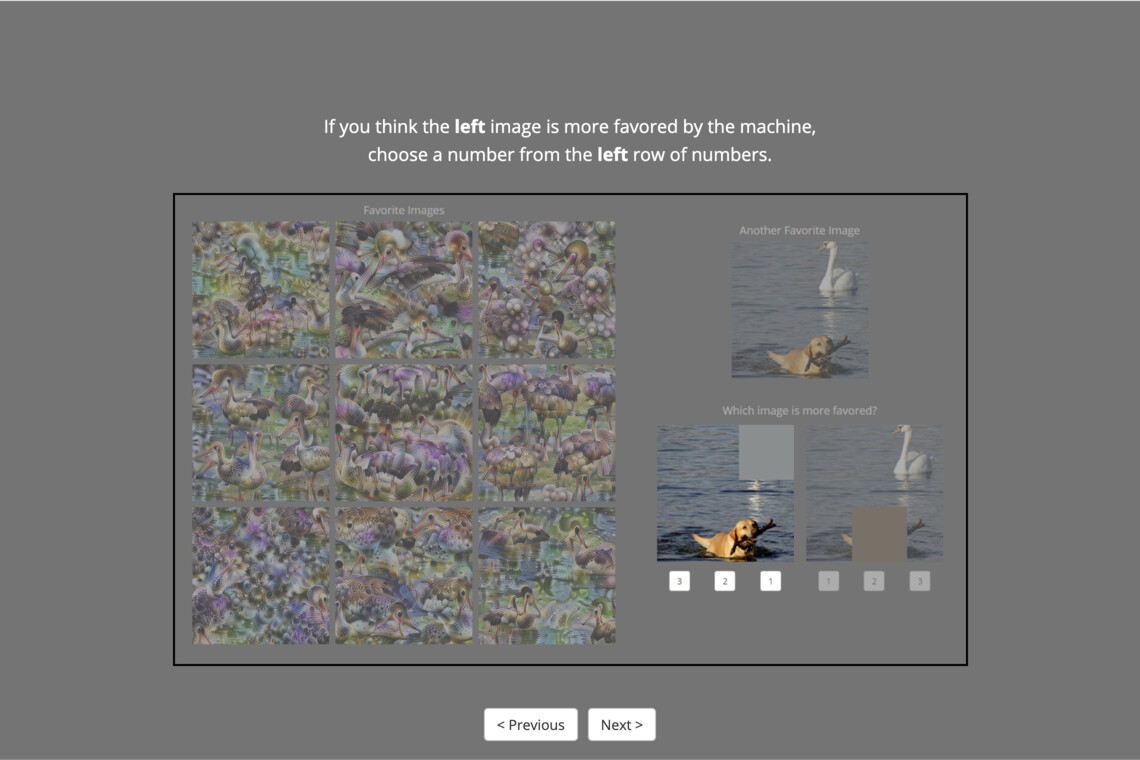}
    \end{subfigure}
    \end{center}
    \caption{First eight instructions at the beginning of the counterfactual-inspired experiment.
    }
    \label{fig:instructions_one}
\end{figure}

\begin{figure}[h!]
    \begin{center}
    \begin{subfigure}[b]{0.45\textwidth}
            \includegraphics[trim=70 35 70 20, clip, width=\textwidth]{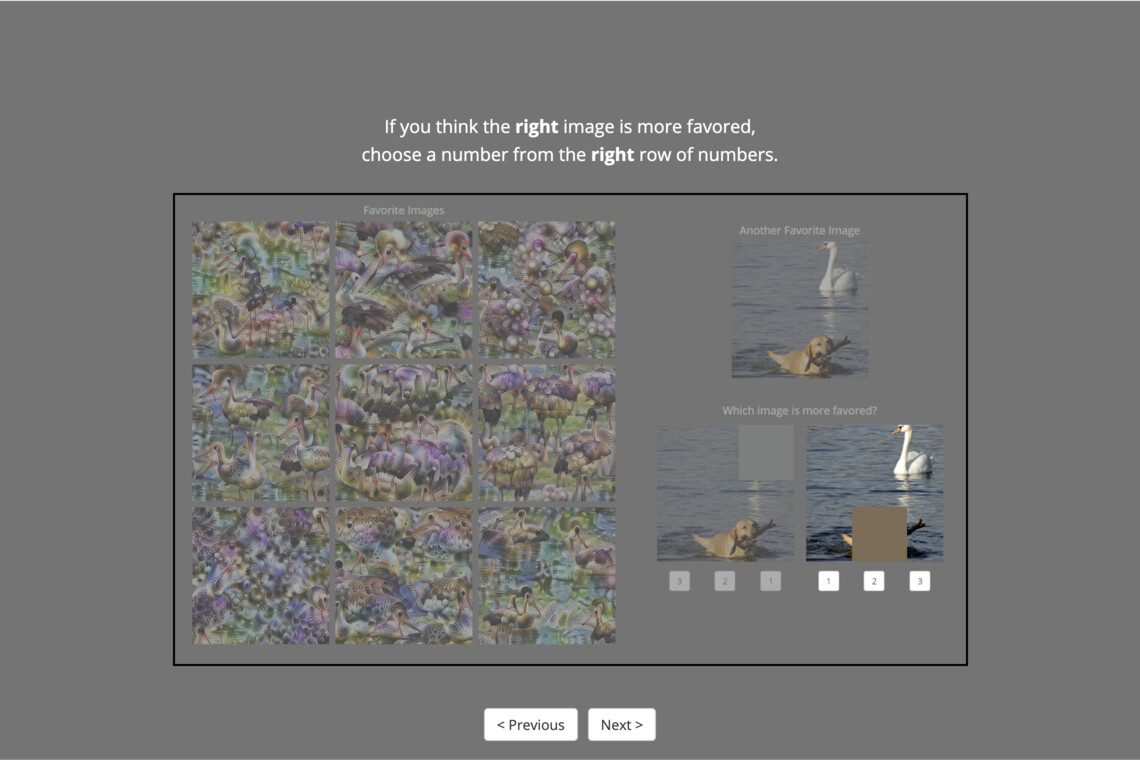}
    \end{subfigure}
    \begin{subfigure}[b]{0.45\textwidth}
            \includegraphics[trim=70 35 70 20, clip, width=\textwidth]{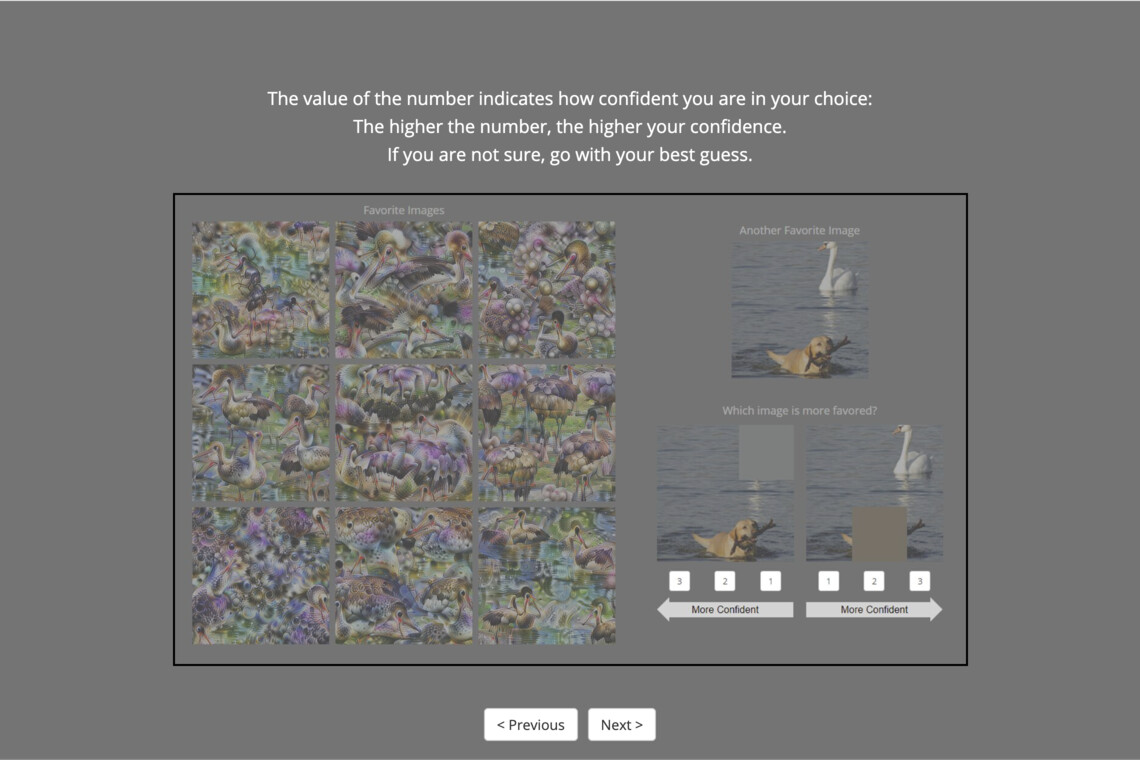}
    \end{subfigure}
    
    \begin{subfigure}[b]{0.45\textwidth}
            \includegraphics[trim=70 35 70 20, clip, width=\textwidth]{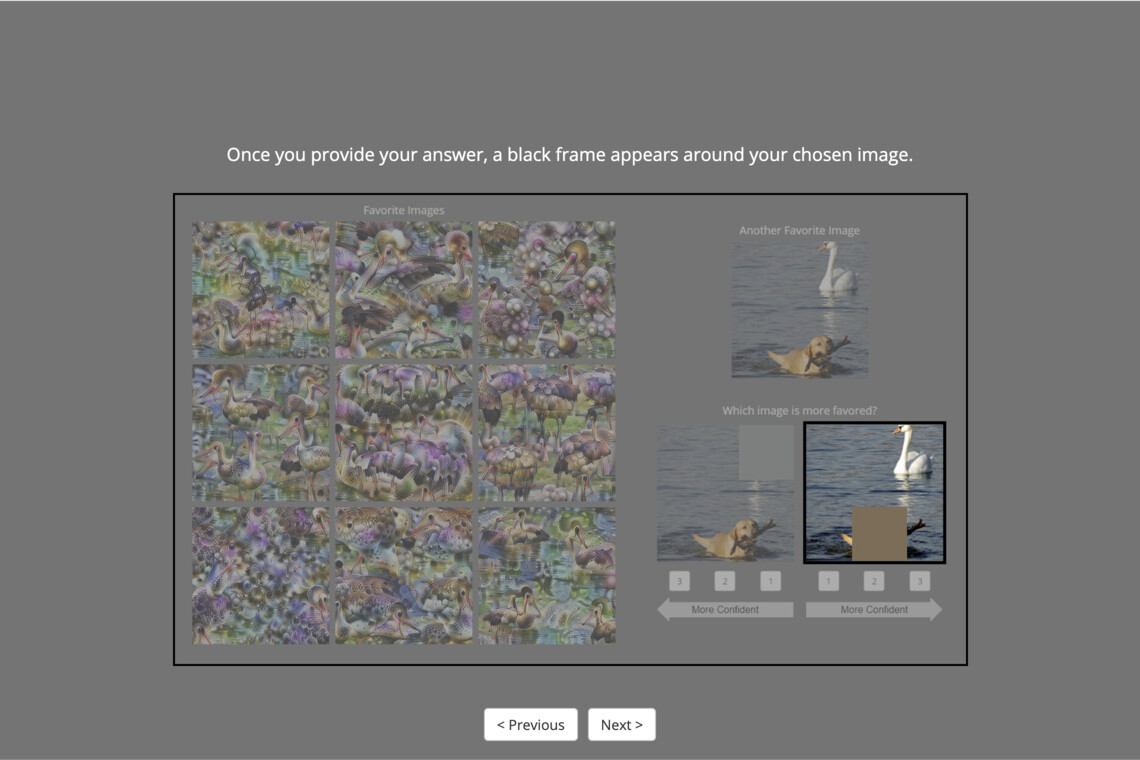}
    \end{subfigure}
    \begin{subfigure}[b]{0.45\textwidth}
            \includegraphics[trim=70 35 70 20, clip, width=\textwidth]{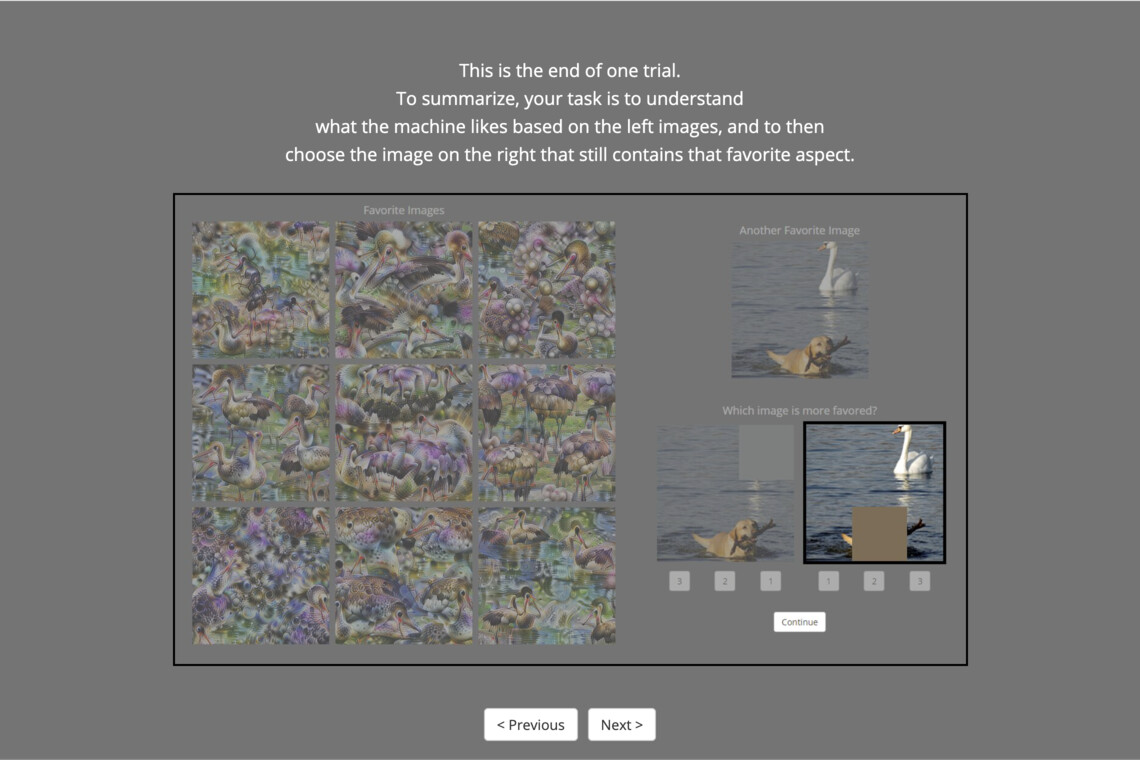}
    \end{subfigure}
    
    \begin{subfigure}[b]{0.45\textwidth}
            \includegraphics[trim=70 35 70 20, clip, width=\textwidth]{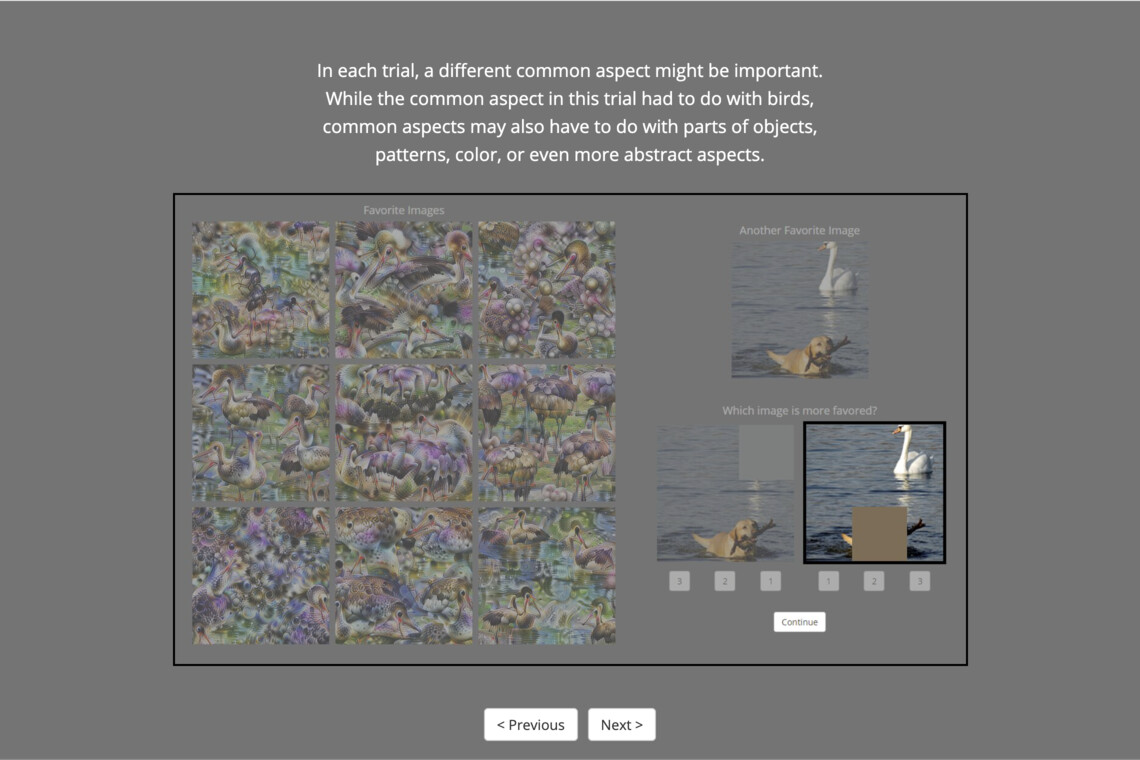}
    \end{subfigure}
    \begin{subfigure}[b]{0.45\textwidth}
            \includegraphics[trim=70 35 70 20, clip, width=\textwidth]{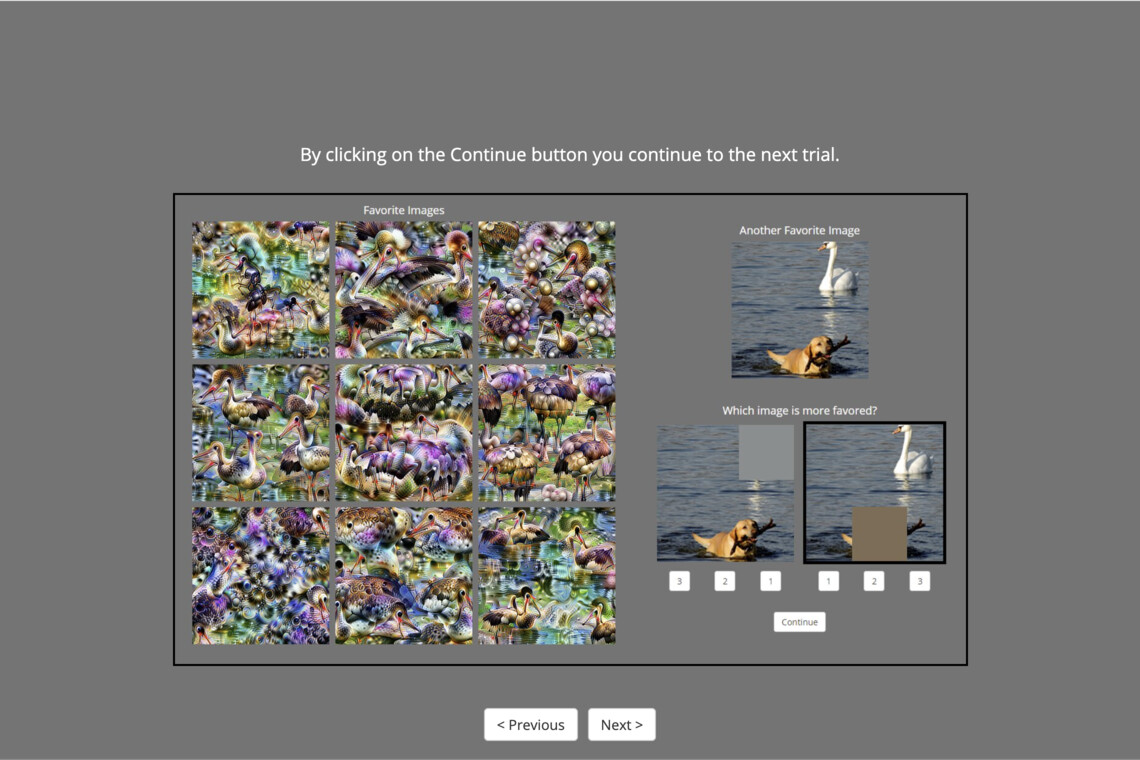}
    \end{subfigure}
    
    \begin{subfigure}[b]{0.45\textwidth}
            \includegraphics[trim=70 35 70 20, clip, width=\textwidth]{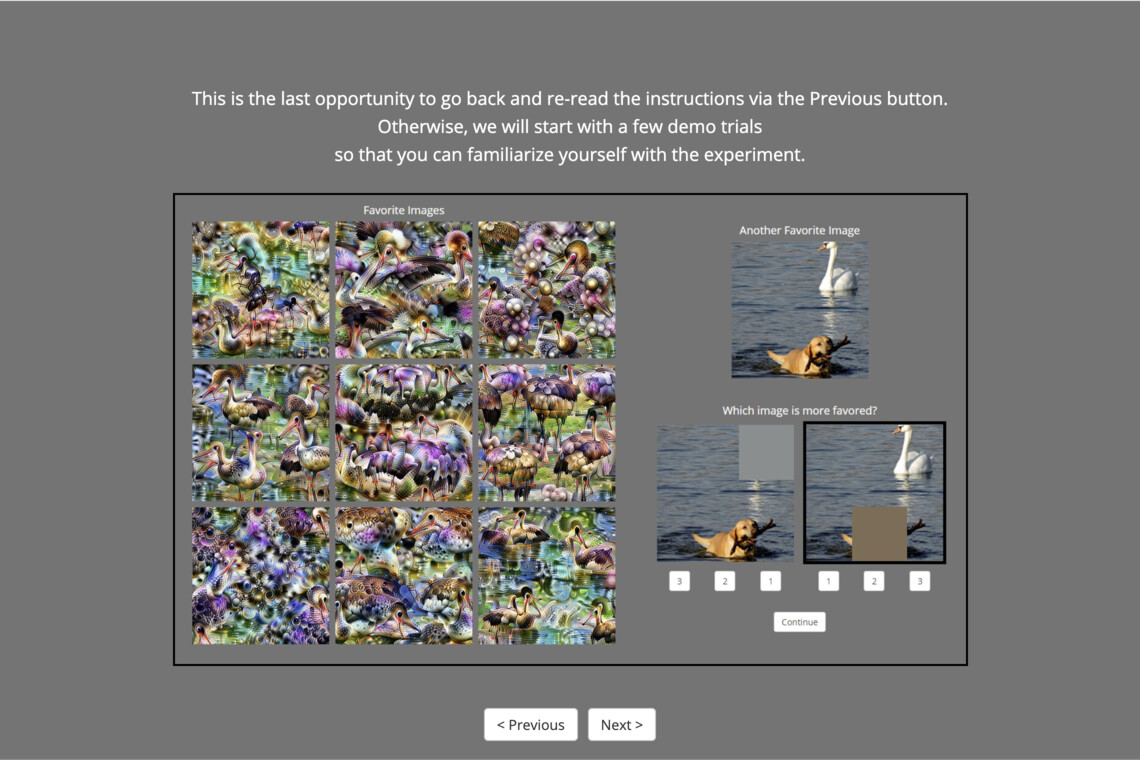}
    \end{subfigure}
    \begin{subfigure}[b]{0.45\textwidth}
            \includegraphics[trim=70 35 70 20, clip, width=\textwidth]{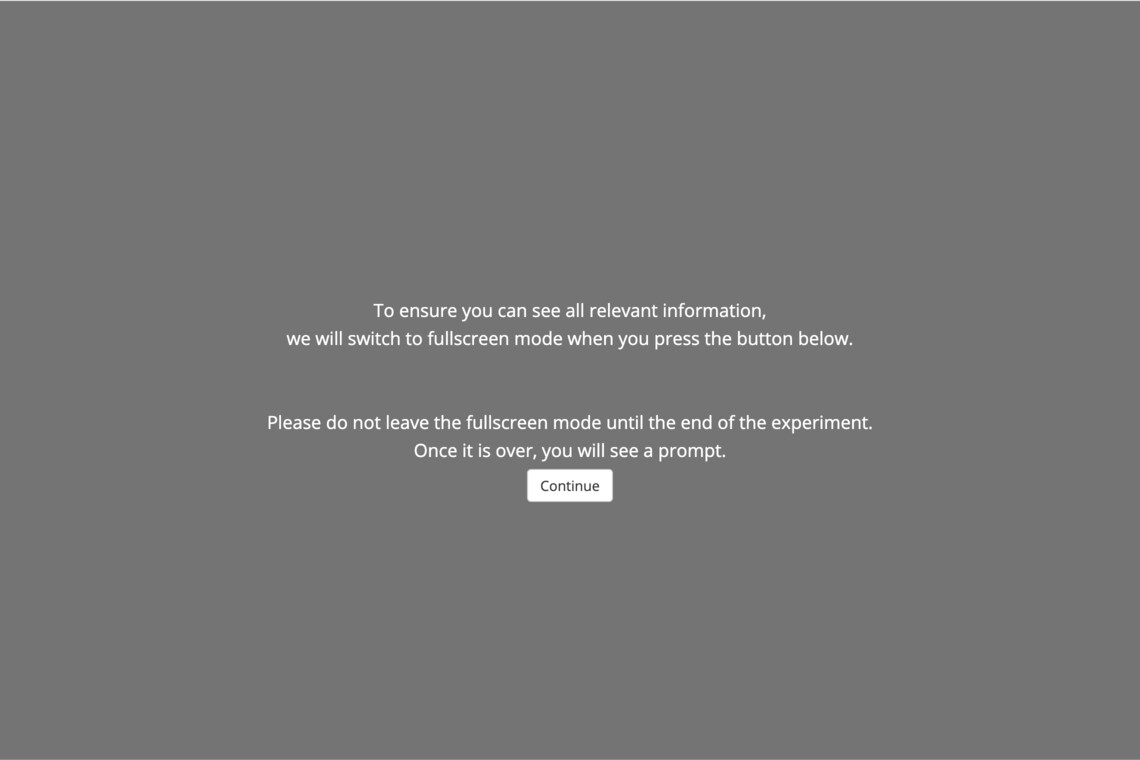}
    \end{subfigure}
    \end{center}
    \caption{Second eight instructions at the beginning of the counterfactual-inspired experiment.
    }
    \label{fig:instructions_two}
\end{figure}

\paragraph{Instruction Trial}
To explain the task as intuitively as possible, we construct an easy, artificial instruction trial (see Fig.~\ref{fig:instructions_one}~and~\ref{fig:instructions_two}): At first, we select a unit with easily understandable feature visualizations: The synthetic images of unit $720$ of the \textsc{Pool} branch in layer $8$ show relatively clear bird-like structures. From a popular image search engine, we then select an image\footnote{\url{https://pixnio.com/fauna-animals/dogs/dog-water-bird-swan-lake-waterfowl-animal-swimming} released into public domain under CC0 license by Bicanski.} which not only clearly shows a bird but also other objects, namely a dog and water. To construct the minimally and maximally activating query images, we place the occlusion patches manually on the bird and dog.

\begin{table}[ht]
    \caption{Hand-picked unit choices for instruction, catch and practice trials in the counterfactual-inspired experiment.}%
    \label{tab:feature_maps_instruction_catch_practice}
    \vspace{0.2cm}
    \centering
    \begin{tabular}[t]{ccccc}
        \toprule
        Trial Type & Layer & Branch & Unit & Difficulty Level\\
        \midrule
        instruction & mixed5a & pool & 720 & very easy\\
        \midrule
        \multirow{3}{*}{catch} & mixed4e & pool & 783 & very easy\\
                               & mixed4c & pool & 484 & very easy\\
                               & mixed5a & $3\times3$ & 557 & very easy\\
        \midrule
        \multirow{10}{*}{practice} & mixed4e & $1\times1$ & 6 & very easy\\
                                & mixed4a & pool & 505 & very easy\\
                                & mixed4e & pool & 809 & very easy\\
                                & mixed4c & pool & 449 & easy\\
                                & mixed4b & pool & 465 & easy\\
                                & mixed4c & $1\times1$ & 59 & easy\\
                                & mixed4e & $1\times1$ & 83 & easy\\
                                & mixed3a & $1\times1$ & 43 & easy\\
                                & mixed3b & pool & 472 & easy\\
                                & mixed4b & $1\times1$ & 5 & easy\\
        \bottomrule
    \end{tabular}
\end{table}


\paragraph{Practice Trials} 
In each attempt to pass the practice block, the trials are randomly sampled from a pool of $10$ trials (see Table~\ref{tab:feature_maps_instruction_catch_practice}). Please note that unlike in any other trial type, participants receive feedback in the practice block: After each trial, they learn whether their chosen image truly is the query image of higher activation.

\paragraph{Catch Trials}
While all conditions with reference images use hand-picked easy trials (see Table~\ref{tab:feature_maps_instruction_catch_practice}), the none condition cannot rely on straight-forward clues from references. Therefore, we exchange the minimal query image with a minimal query image of a different, otherwise unused unit. This ensures that the catch trials in the none condition are also obvious.

\subsubsection{Infrastructure}\label{app:instrastructure}

The online experiment is hosted on an Ubuntu 18.04 server running on an Intel(R) Xeon(R) Gold 5220 CPU. The experiment is implemented in JavaScript using jspsych 6.3.1 \citep{de2016psychophysics} and flask via Python 3.6.
The generation of the stimuli shown in the experiment was completed in approximately 35 hours on a single GeForce GTX 1080 GPU. The calculation of all baselines required 8 additional GPU hours.

\subsubsection{Amazon Mechanical Turk}\label{app:MTurk}

\paragraph{MTurk participants} To increase the chance that all MTurk participants understand the English instructions at the beginning of the experiment, we restrict access to workers from the following English-speaking countries: USA, Canada, Great Britain, Australia, New Zealand and Ireland.

\paragraph{Financial Compensation} Based on an estimated duration and pilot experiments as well as a targeted hourly rate of US\$\,$15$, we calculate the pay to be US\$\,$0.70$ for the none condition and US\$\,$1.95$ for all other conditions. MTurk participants whose data we include need a mean time of $209.64 \pm 79.53$\,s and $396.87 \pm 145.78$\,s for the whole experiment for the none condition and for all other conditions, respectively, which results in an hourly compensation of $\approx 12.02$\, US\$/hour and $17.69$\,US\$/hour, respectively. All MTurk participants who fully complete a HIT are paid, regardless of whether their responses meet the exclusion criteria. A total of US\$\,$1989.06$ is spent on all pilot and real replication and counterfactual-inspired experiments.

\paragraph{Rights to Data} We do not gather personal identifiable data from any MTurk participant. According to the MTurk Participation Agreement 3a \footnote{ \url{https://www.mturk.com/participation-agreement}, accessed on May 22nd, 2021}, workers agree to vest all ownership and intellectual property rights to the requester (i.e., the authors of this study). Besides informing MTurk participants in the HIT preview about the academic and image classification nature of the experiment, we restate that ``By completing this HIT, you consent to your anonymized data being shared with us for a scientific study.'' Further, we provide an email address, which some MTurk participants used to share feedback.

\clearpage
\subsection{Details on Results of Counterfactual-Inspired Experiment}

\subsubsection{Different Query images}

\begin{figure}[!htbp]
    \begin{subfigure}[b]{\textwidth}
        \begin{center}
            \includegraphics[width=\textwidth]{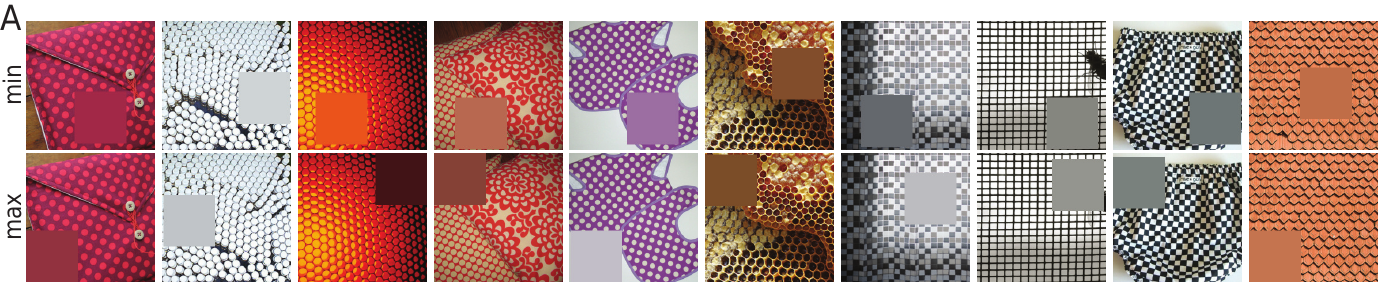}%
            \label{fig:query_image_examples_a}
        \end{center}
    \end{subfigure}
    \begin{subfigure}[b]{\textwidth}
        \begin{center}
            \includegraphics[width=\textwidth]{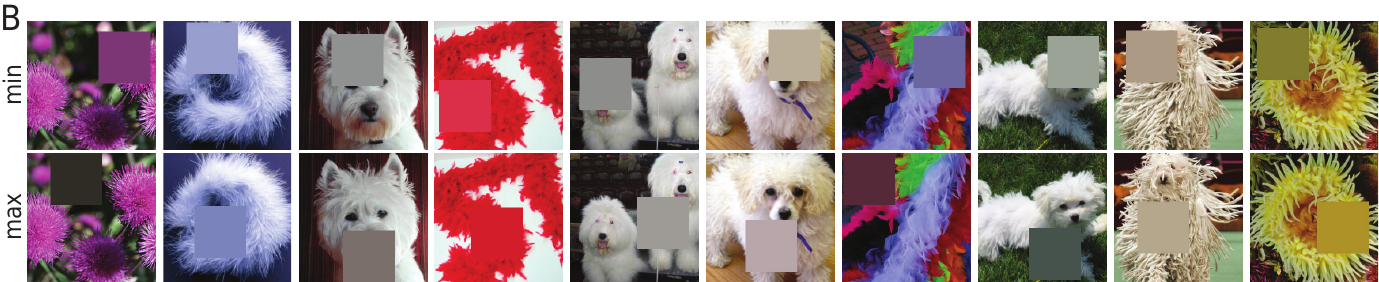}%
            \label{fig:query_image_examples_b}
        \end{center}
    \end{subfigure}
    \caption{For each unit, we test $10$ different image sets in the counterfactual-inspired experiment. The diversity of query images for layer $3$ of the $3 \times 3$ branch (\textbf{A}), and layer $7$ of the \textsc{Pool} branch (\textbf{B}) gives an intuitive explanation for varying performances.
    }
    \label{fig:query_image_examples}
\end{figure}

\subsubsection{Confidence Ratings and Reaction Times}

\begin{figure}[!htbp]
    \centering
    \begin{subfigure}{0.32\textwidth}
        \includegraphics[width=\textwidth]{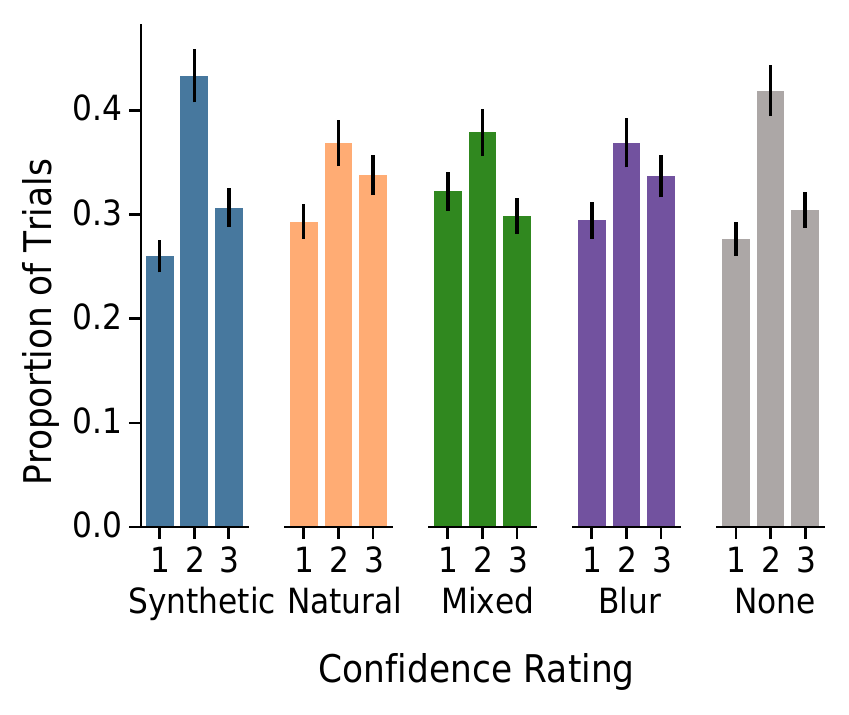}
        \caption{}
        \label{fig:apx_confidence_conditioned_on_correctness_all}
    \end{subfigure}
    \begin{subfigure}{0.32\textwidth}
        \includegraphics[width=\textwidth]{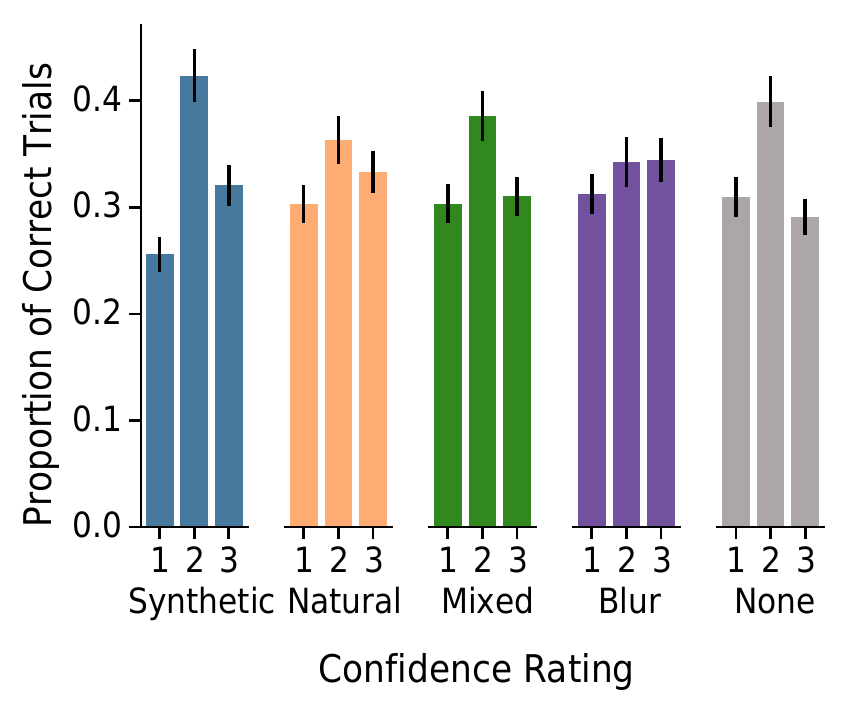}
        \caption{}
        \label{fig:apx_confidence_conditioned_on_correctness_correct}
    \end{subfigure}
    \begin{subfigure}{0.32\textwidth}
        \includegraphics[width=\textwidth]{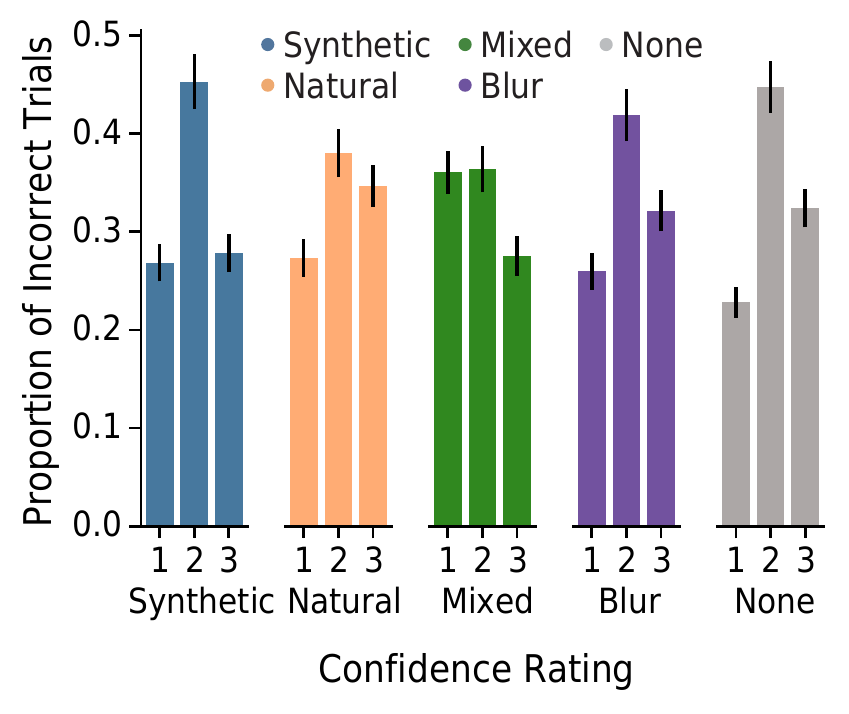}
        \caption{}
        \label{fig:apx_confidence_conditioned_on_correctness_incorrect}
    \end{subfigure}
    
    \caption{Confidence ratings of MTurk participants in the different reference conditions for (a) all, (b) only correct or (c) only incorrect trials of the counterfactual-inspired experiment.}
    \label{fig:apx_confidence}
\end{figure}

\begin{figure}[!htbp]
    \centering
    \begin{subfigure}{0.32\textwidth}
        \includegraphics[width=\textwidth]{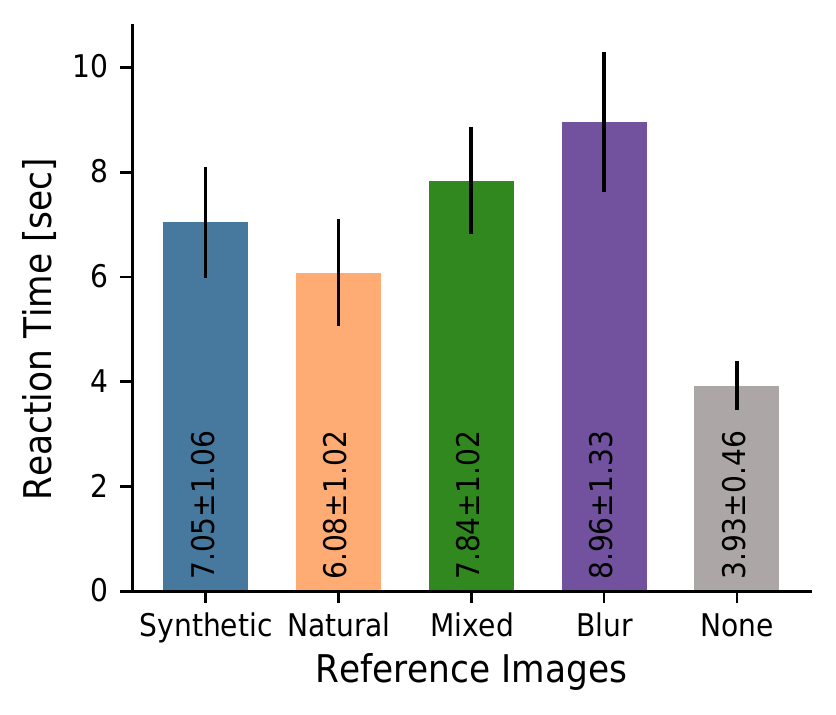}
        \caption{}
        \label{fig:apx_reaction_time_conditioned_on_correctness_all}
    \end{subfigure}
    \begin{subfigure}{0.32\textwidth}
        \includegraphics[width=\textwidth]{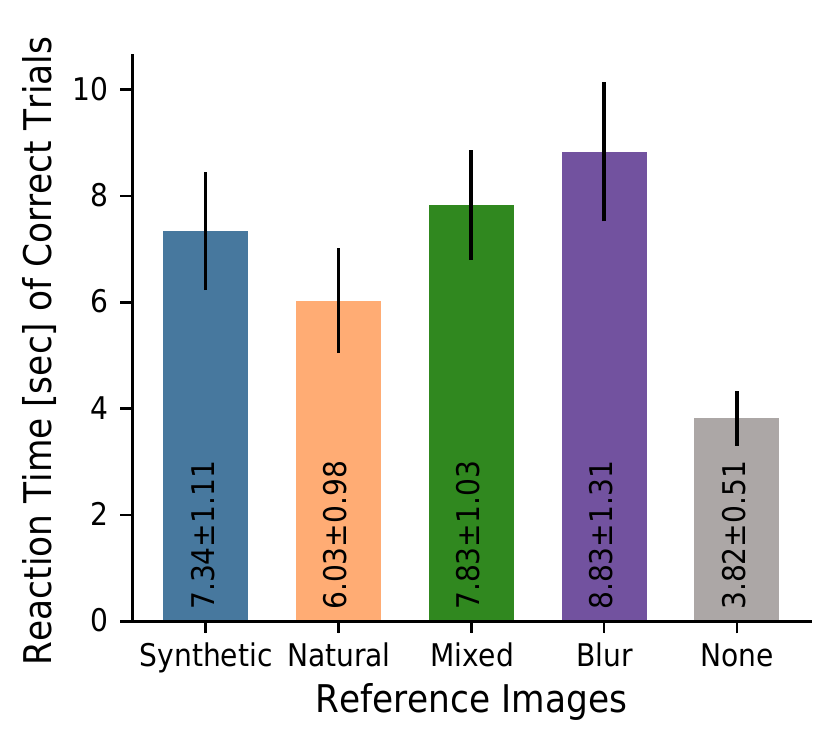}
        \caption{}
        \label{fig:apx_reaction_time_conditioned_on_correctness_correct}
    \end{subfigure}
    \begin{subfigure}{0.32\textwidth}
        \includegraphics[width=\textwidth]{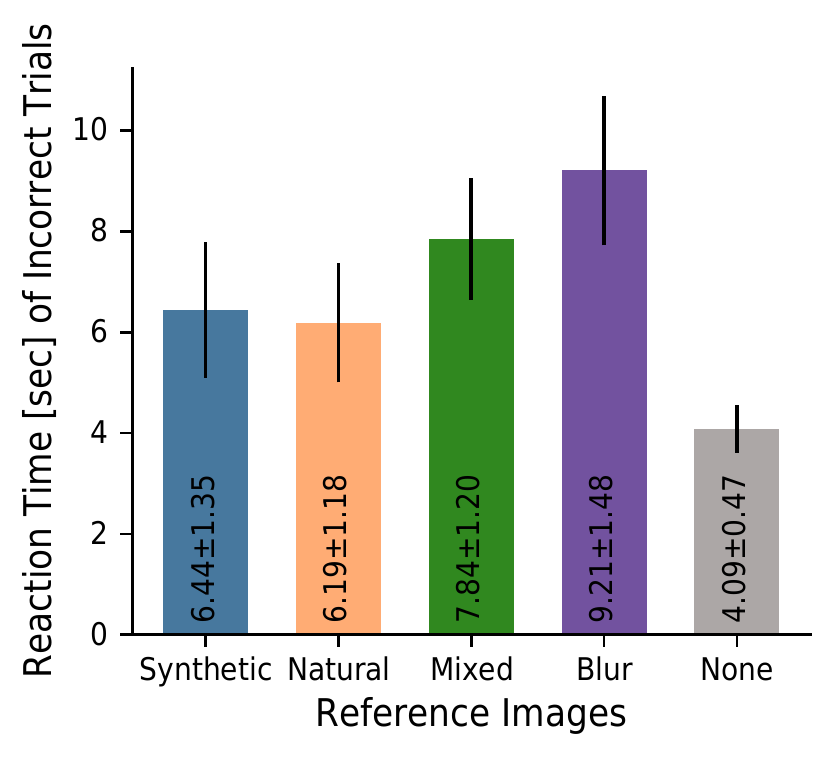}
        \caption{}
        \label{fig:apx_reaction_time_conditioned_on_correctness_incorrect}
    \end{subfigure}
    
    \caption{Reaction times of MTurk participants in the different reference conditions for (a) all, (b) only correct or (c) only incorrect trials of the counterfactual-inspired experiment.}
    \label{fig:apx_reaction_time}
\end{figure}

\FloatBarrier
\clearpage
\subsubsection{Performance per Image Set}
\begin{figure}[!htbp]
\begin{center}
\begin{subfigure}{0.98\textwidth}
\includegraphics[width=\textwidth]{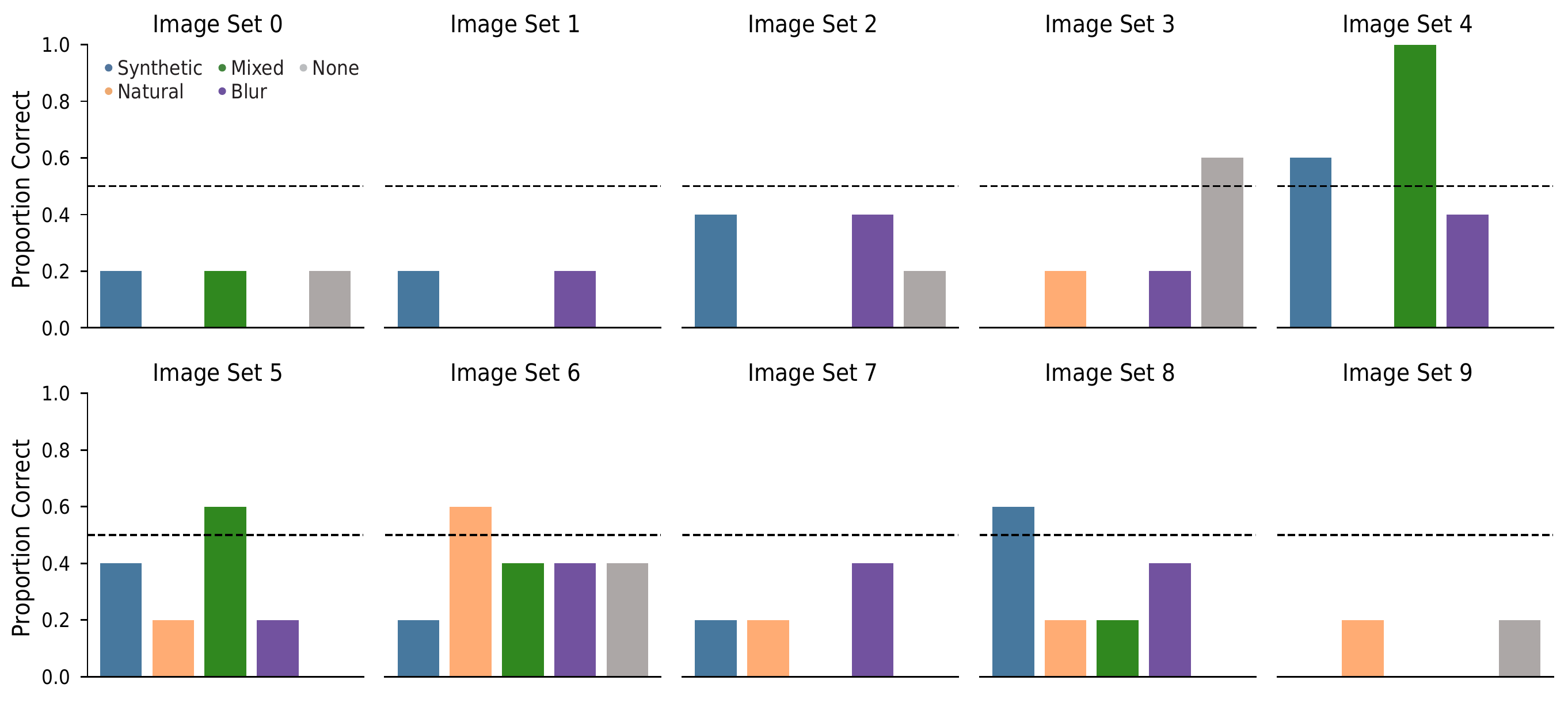}
\caption{Difficult unit.}
\end{subfigure}

\begin{subfigure}{0.98\textwidth}
\includegraphics[width=\textwidth]{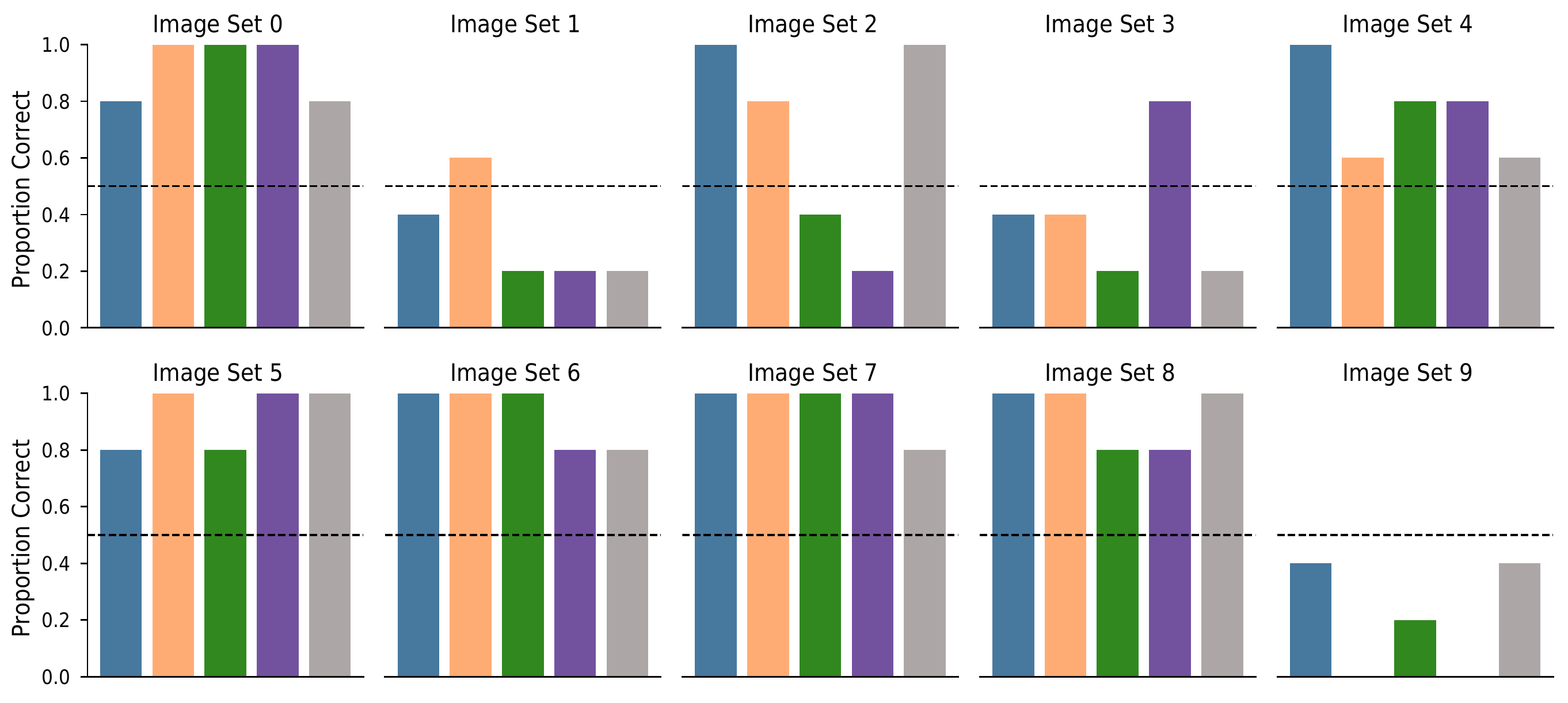}
\caption{Intermediate unit.}
\end{subfigure}

\begin{subfigure}{0.98\textwidth}
\includegraphics[width=\textwidth]{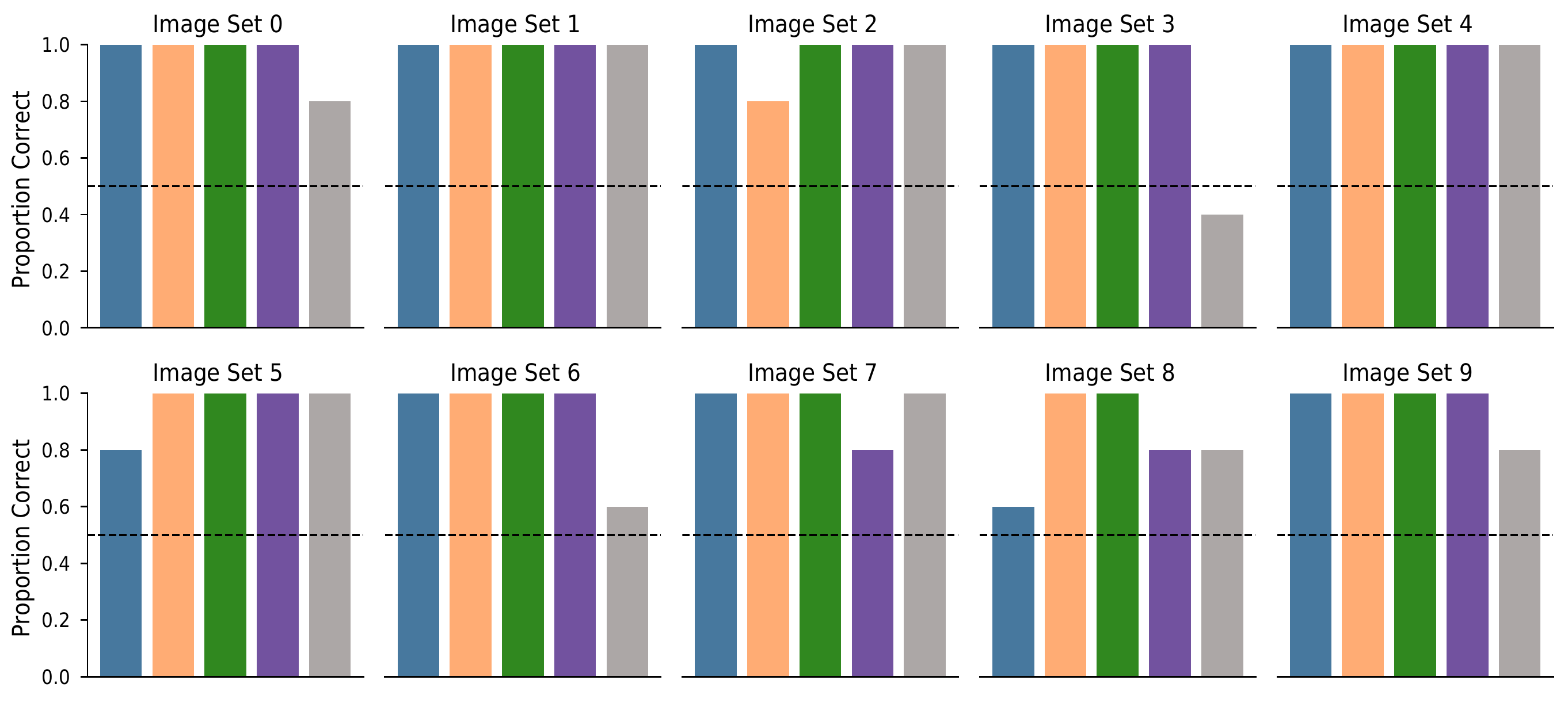}
\caption{Easy unit.}
\end{subfigure}

\end{center}
\caption{Performance in the counterfactual-inspired experiment split up by image sets and conditions for a difficult (layer $3$, \textsc{Pool}), intermediate (layer $7$, \textsc{Pool}) and easy unit (layer $8$, \textsc{Pool}). Each bar shows the average over 5 MTurk participants.}%
\label{fig:performance_per_image_set}
\end{figure}

\FloatBarrier
\clearpage
\subsubsection{Strategy Comparisons}
\begin{figure}[!htbp]
\begin{center}
\includegraphics[width=\textwidth]{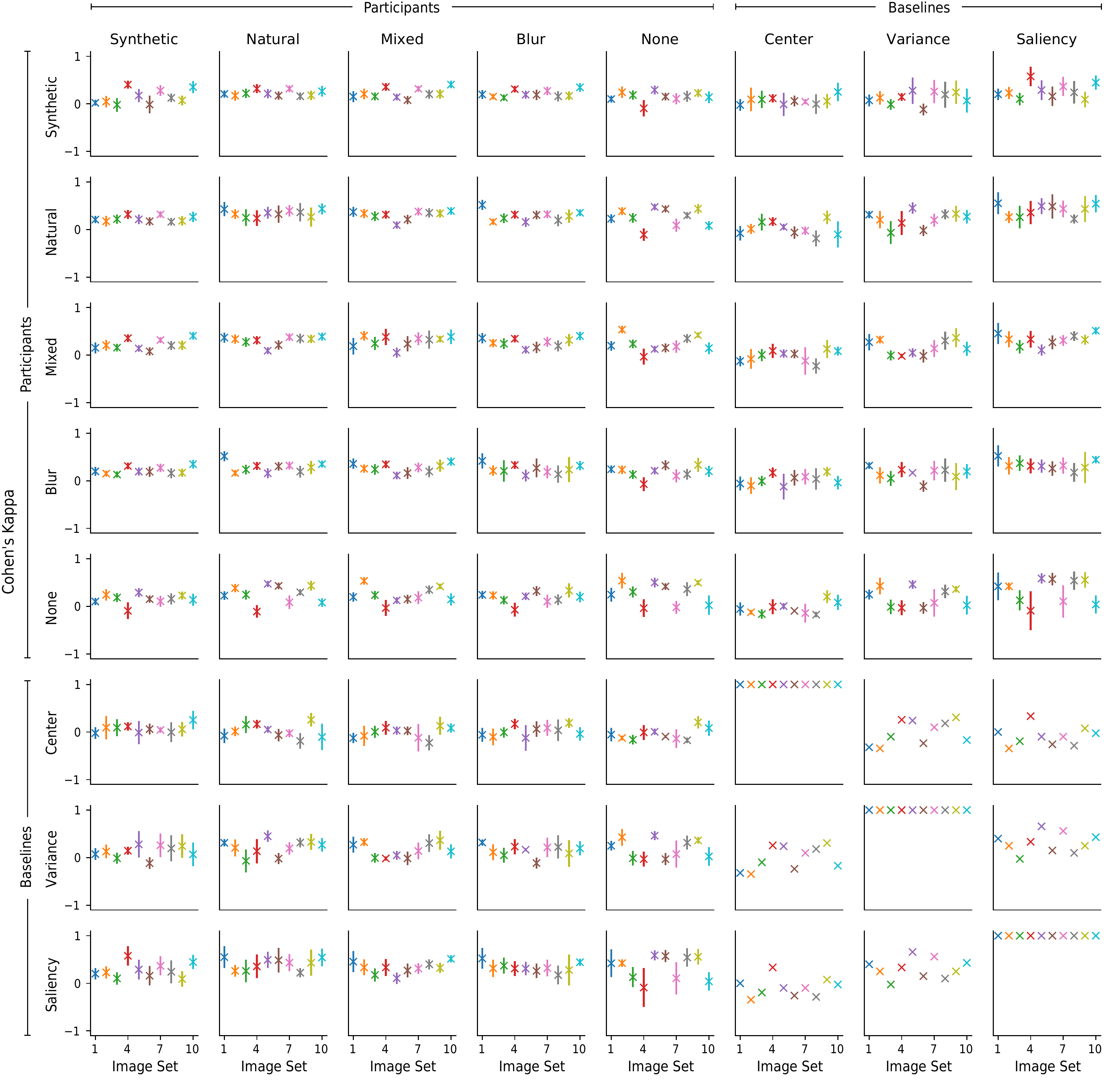}
\end{center}
\caption{Cohen's kappa per image set in the counterfactual-inspired experiment (averages over participant-participant-, participant-baseline- or baseline-baseline-pairs). Error bars denote two standard errors of the mean.}
\label{app_fig:cohens_kappa_details}
\end{figure}

\FloatBarrier
\clearpage
\subsubsection{Relative Activation Differences}

\begin{figure}[!ht]
\begin{center}
\begin{subfigure}{0.49\textwidth}
\centering
\includegraphics[width=.8\textwidth]{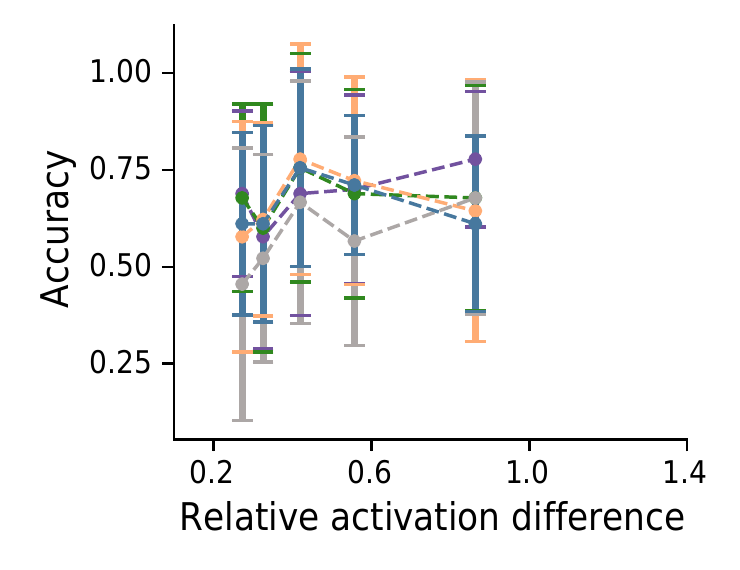}
\caption{$3 \times 3$ branch.}
\end{subfigure}
\begin{subfigure}{0.49\textwidth}
\centering
\includegraphics[width=.8\textwidth]{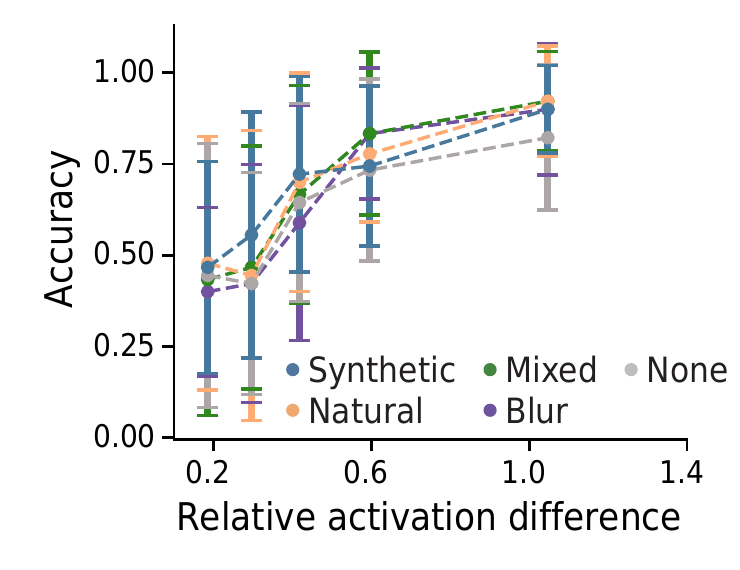}
\caption{\textsc{Pool} branch.}
\end{subfigure}
\end{center}
\caption{Accuracy in the counterfactual-inspired experiment as a function of the relative activation difference between the two query images for the (a) $3\times3$ branch and the (b) \textsc{Pool} branch. Here, the data points shown in Fig.~\ref{fig:accuracy_vs_relative_activations} are summarized in $5$ bins of equal counts; the plot shows the mean and standard deviation for each of the bins.}
\label{fig:apx_accuracy_vs_activation_difference}
\end{figure}

\FloatBarrier
\clearpage
\subsubsection{Exclusion Criteria}

\begin{figure}[H]
\begin{center}
\begin{subfigure}{0.38\textwidth}
\includegraphics[width=\textwidth]{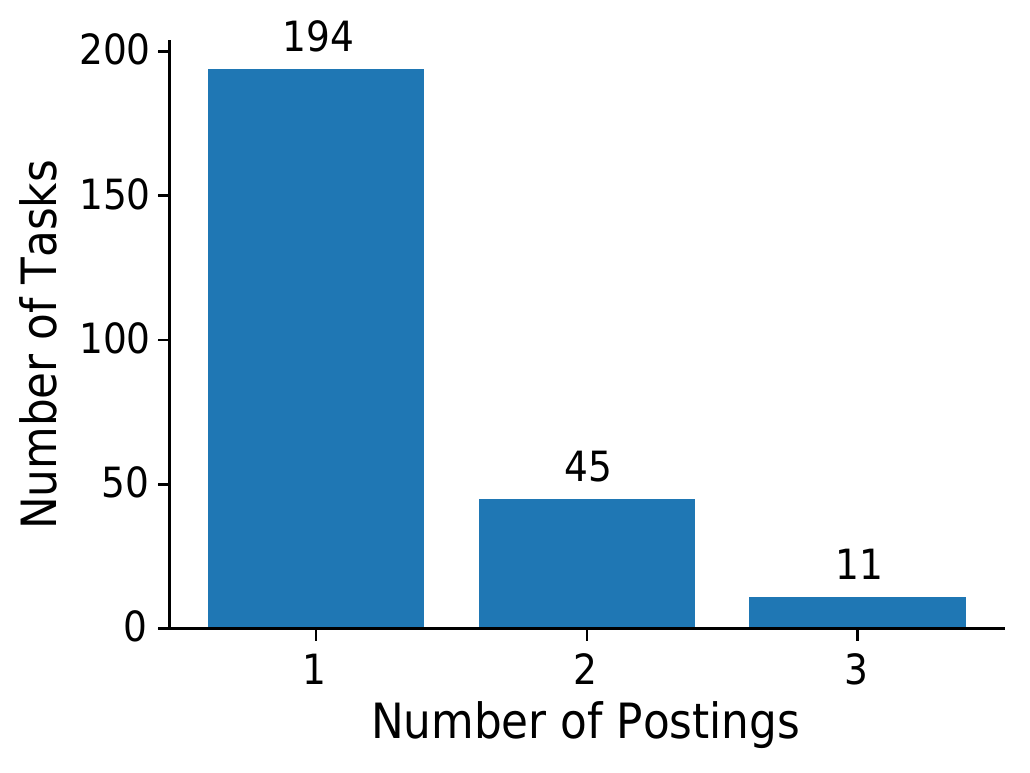}
\caption{Number of times a HIT is posted.}
\end{subfigure}
\begin{subfigure}{0.38\textwidth}
\includegraphics[width=\textwidth]{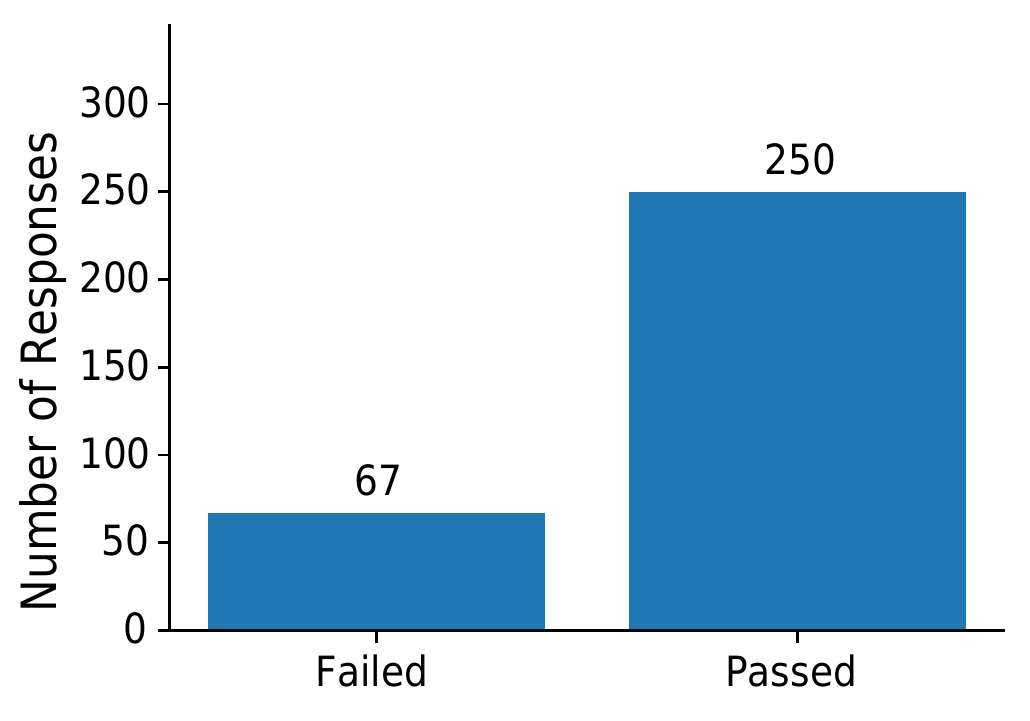}
\caption{All exclusion criteria.}
\end{subfigure}

\begin{subfigure}{0.38\textwidth}
\includegraphics[width=\textwidth]{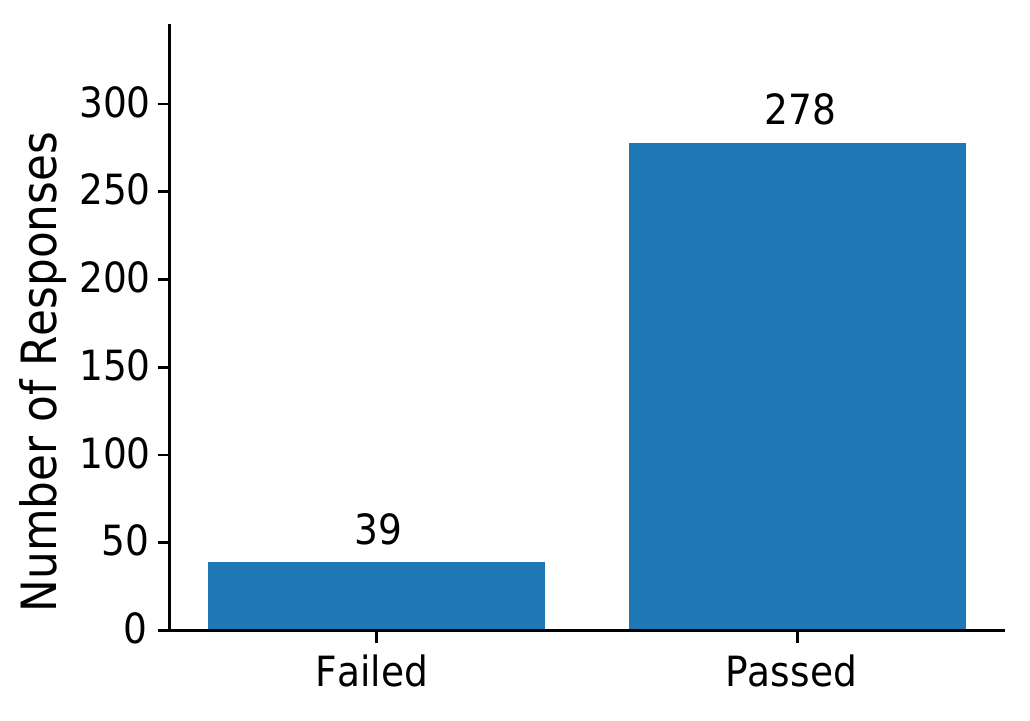}
\caption{Exclusion criterion: catch trials.}
\end{subfigure}
\begin{subfigure}{0.38\textwidth}
\includegraphics[width=\textwidth]{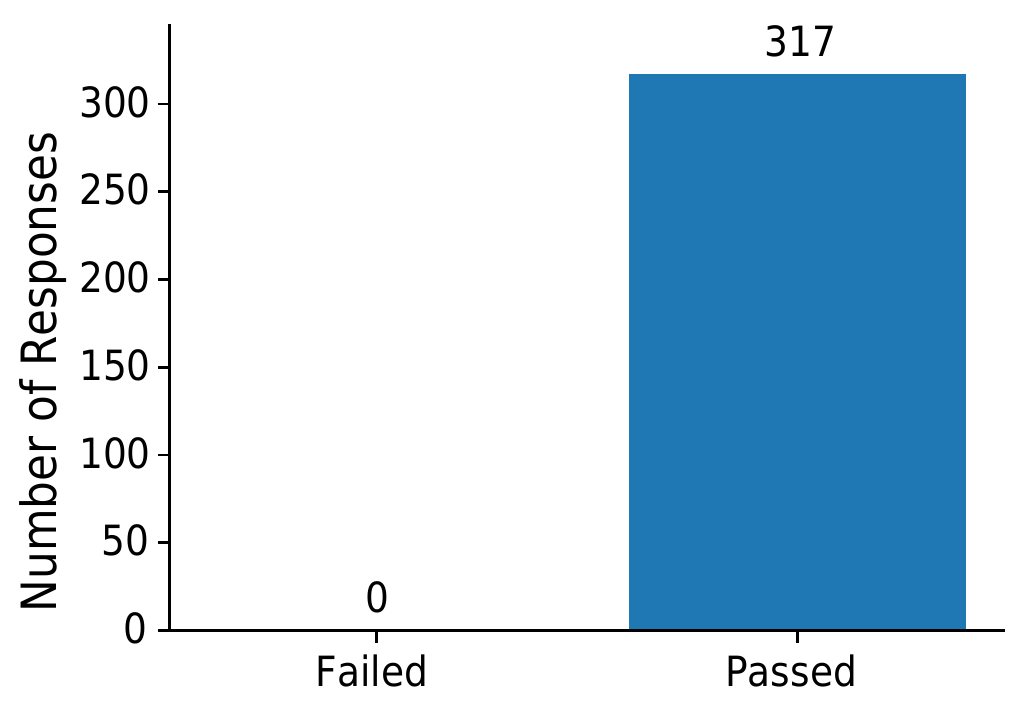}
\caption{Exclusion criterion: row variability.}
\end{subfigure}

\begin{subfigure}{0.38\textwidth}
\includegraphics[width=\textwidth]{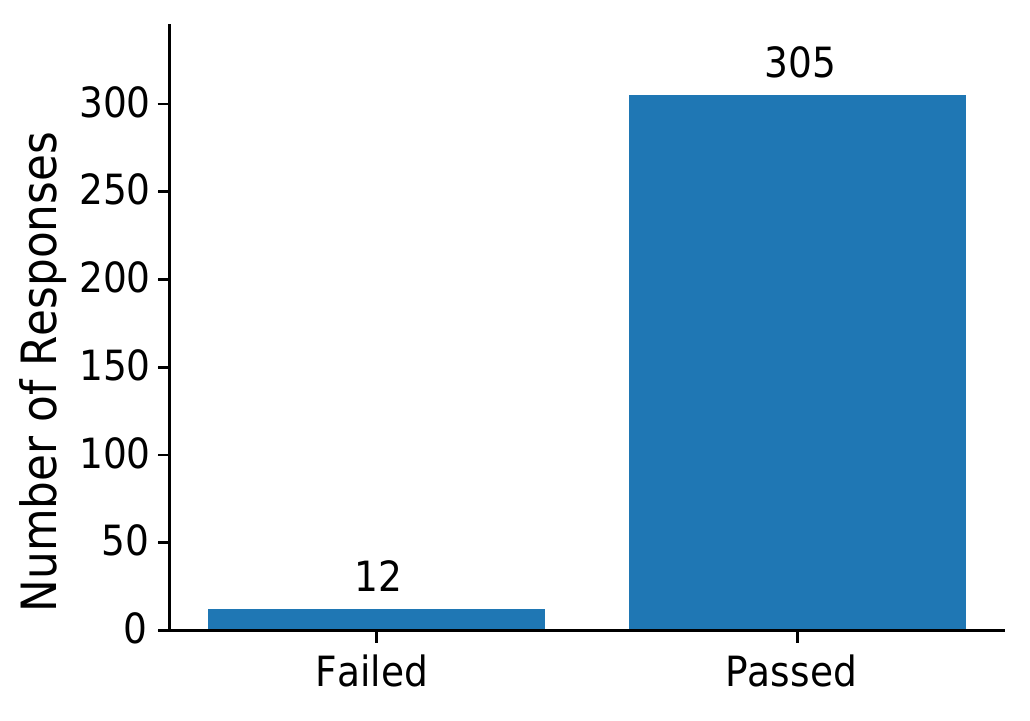}
\caption{Exclusion criterion: instruction time.}
\end{subfigure}
\begin{subfigure}{0.38\textwidth}
\includegraphics[width=\textwidth]{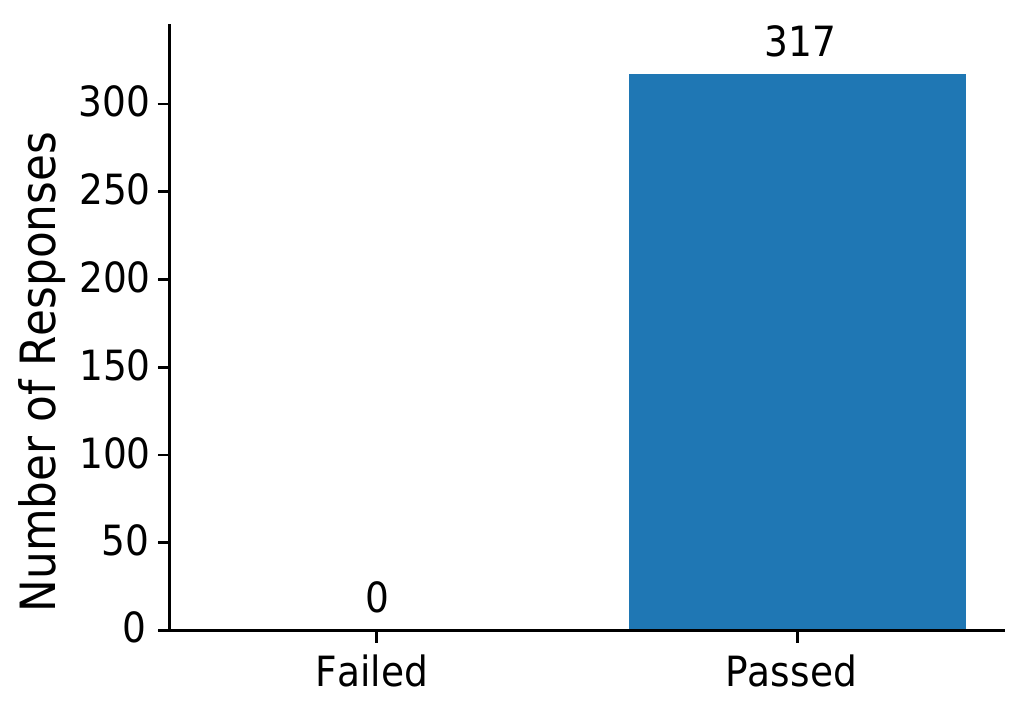}
\caption{Exclusion criterion: total response time.}
\end{subfigure}

\begin{subfigure}{0.38\textwidth}
\includegraphics[width=\textwidth]{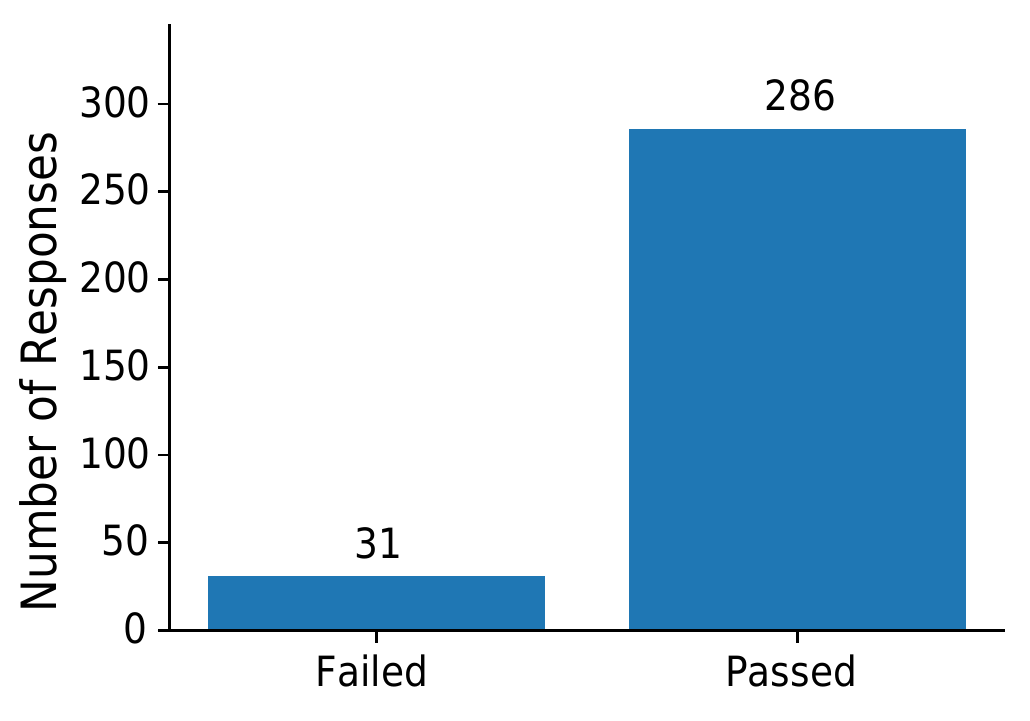}
\caption{Exclusion criterion: practice block.}
\end{subfigure}

\end{center}
\caption{(a) Number of times a HIT is posted. To limit the financial risk, we limit the maximal number of times that a HIT can be posted at $3$. (b-g) Distributions of MTurk participants that passed/failed the exclusion criteria in the counterfactual-inspired experiment on MTurk. Note that the sum of the counts of responses for the individual exclusion criteria in c-f is higher than the summary in b because a participant may have failed more than one exclusion criterion.}
\label{fig:intervention_experiment_mturk_analysis_postings_exclusion_criteria}
\end{figure}

\begin{figure}[H]
\begin{center}
\begin{subfigure}{0.32\textwidth}
    \includegraphics[width=\textwidth]{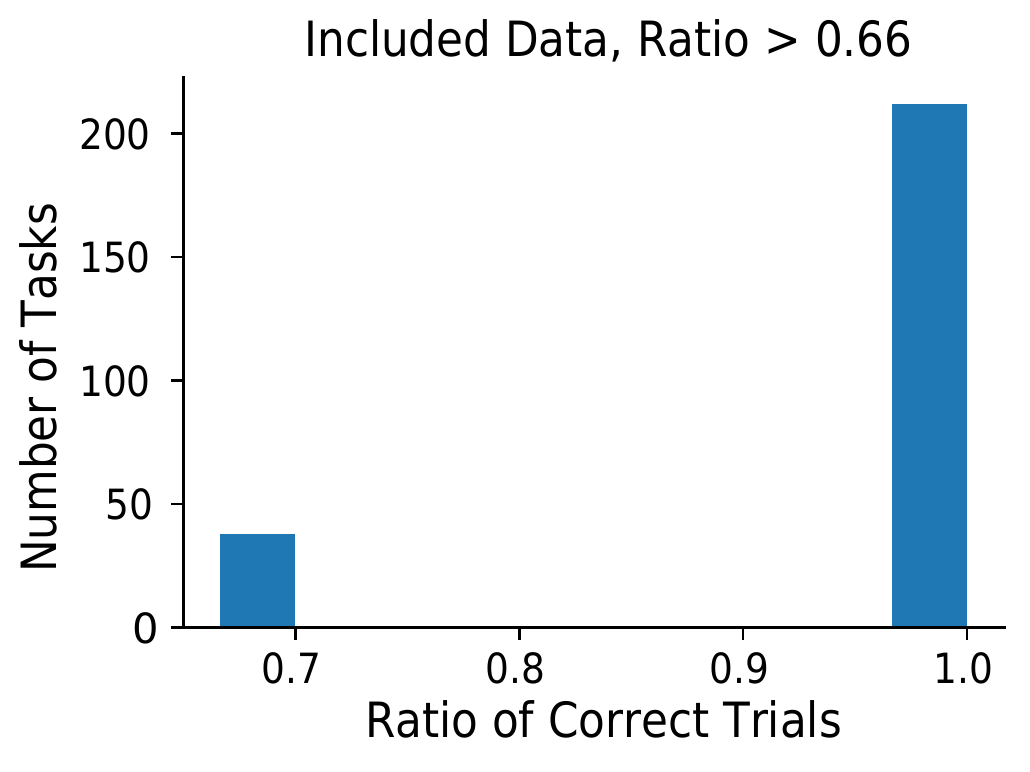}
    \caption{Catch trials from included data.\\\phantom{.}}
\end{subfigure}
\begin{subfigure}{0.32\textwidth}
    \includegraphics[width=\textwidth]{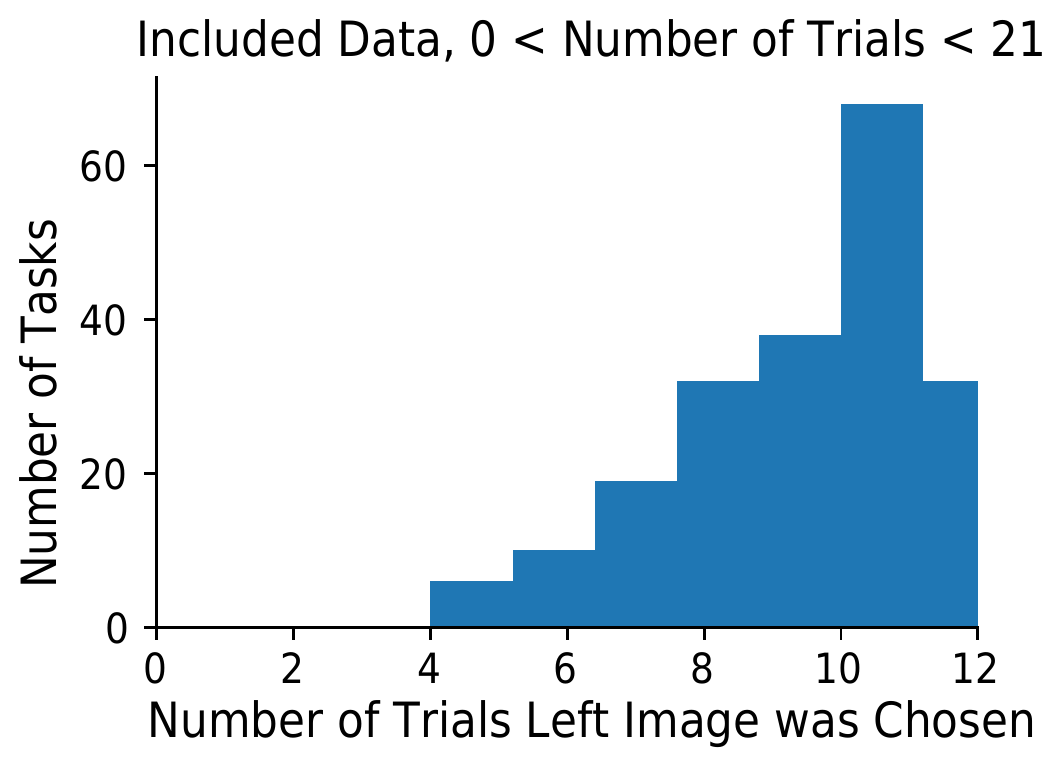}
    \caption{Row variability from included data.}
\end{subfigure}
\begin{subfigure}{0.32\textwidth}
    \includegraphics[width=\textwidth]{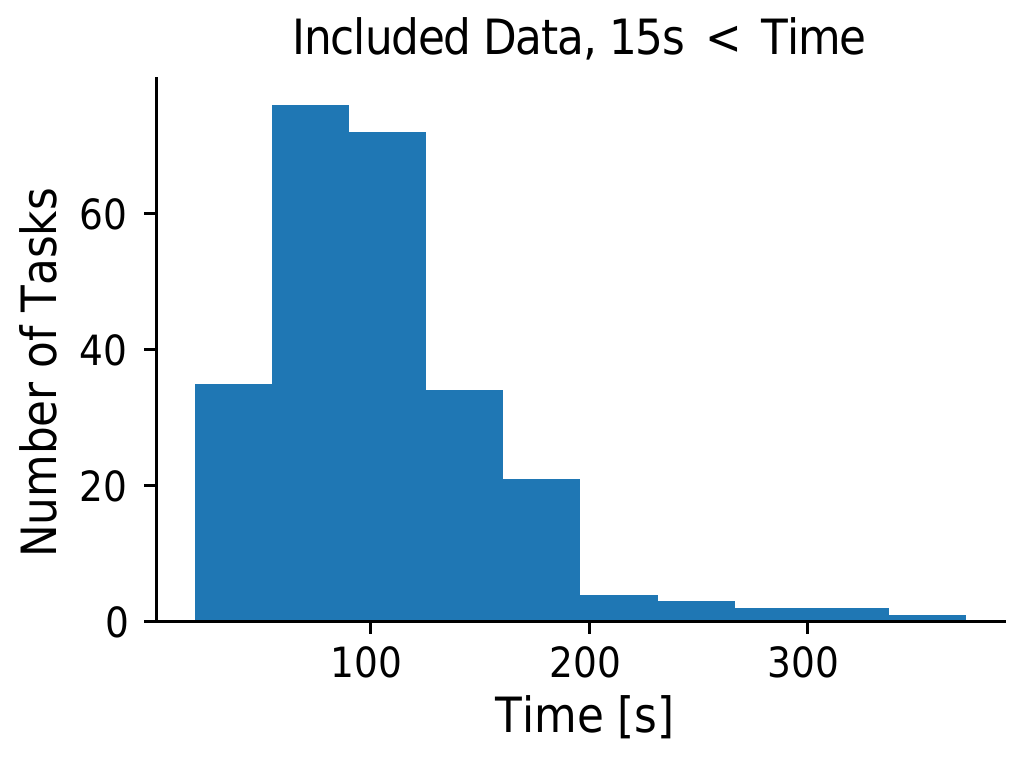}
    \caption{Instruction time from included data.}
\end{subfigure}

\begin{subfigure}{0.32\textwidth}
    \includegraphics[width=\textwidth]{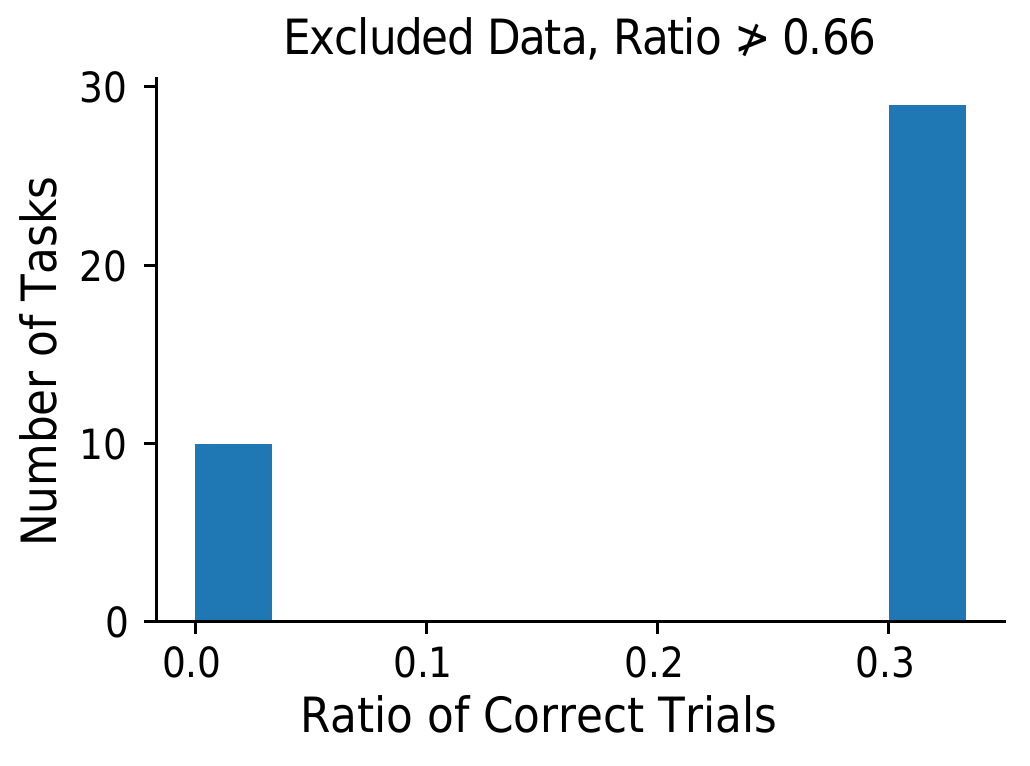}
    \caption{Catch trials from excluded data.\\\phantom{.}}
\end{subfigure}
\begin{subfigure}{0.32\textwidth}
    \includegraphics[width=\textwidth]{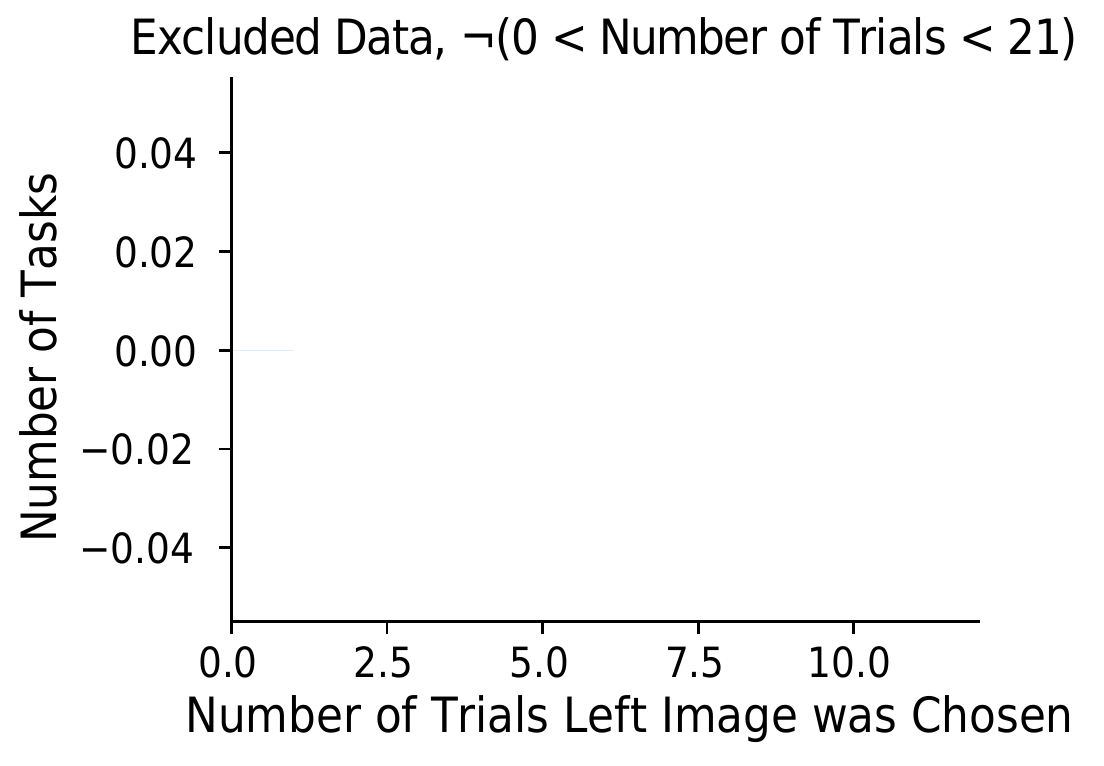}
    \caption{Row variability from excluded data.}
\end{subfigure}
\begin{subfigure}{0.32\textwidth}
    \includegraphics[width=\textwidth]{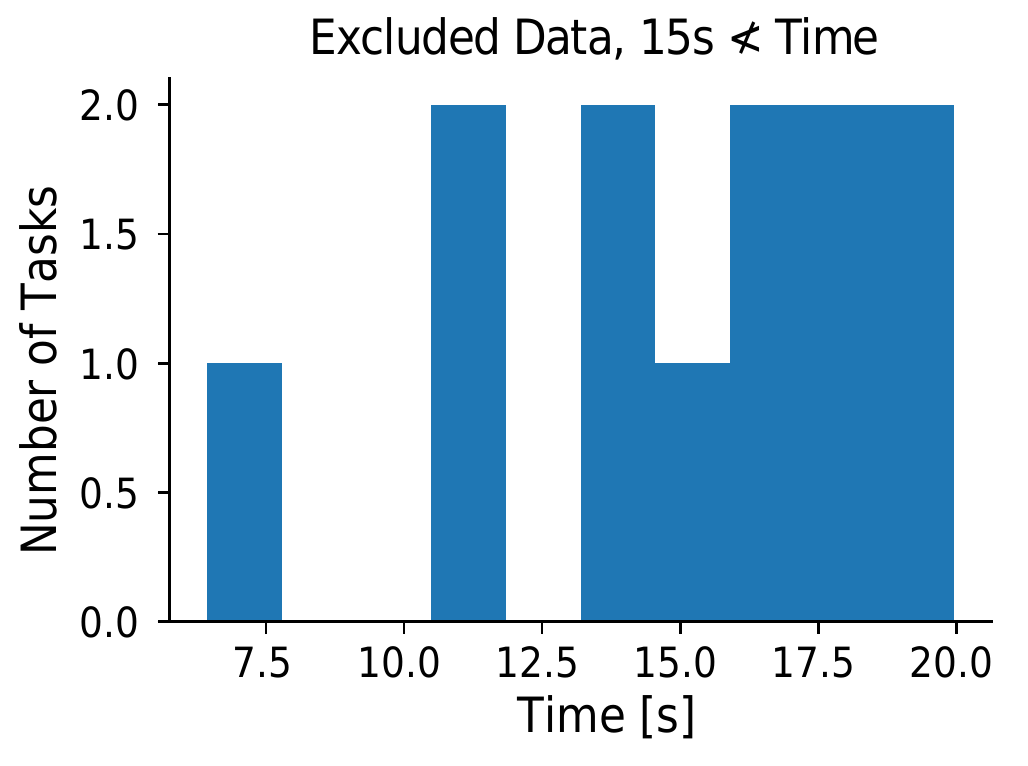}
    \caption{Instruction time from excluded data.}
\end{subfigure}

\begin{subfigure}{0.32\textwidth}
    \includegraphics[width=\textwidth]{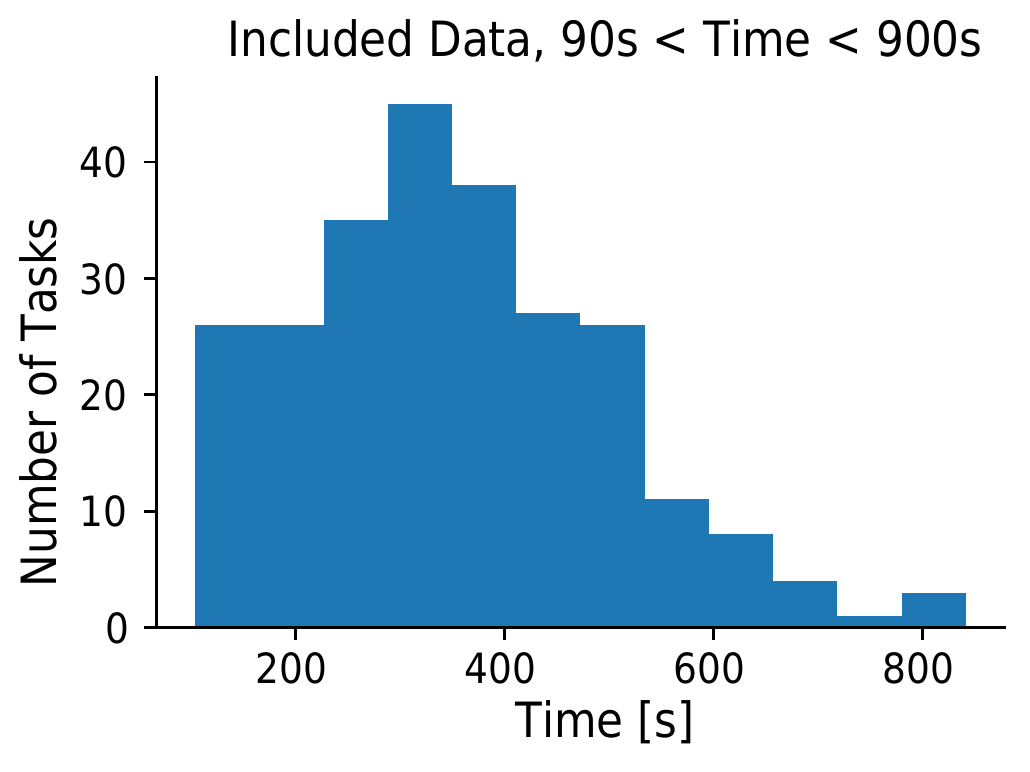}
    \caption{Total response time from included data.}
\end{subfigure}
\begin{subfigure}{0.33\textwidth}
    \includegraphics[width=\textwidth]{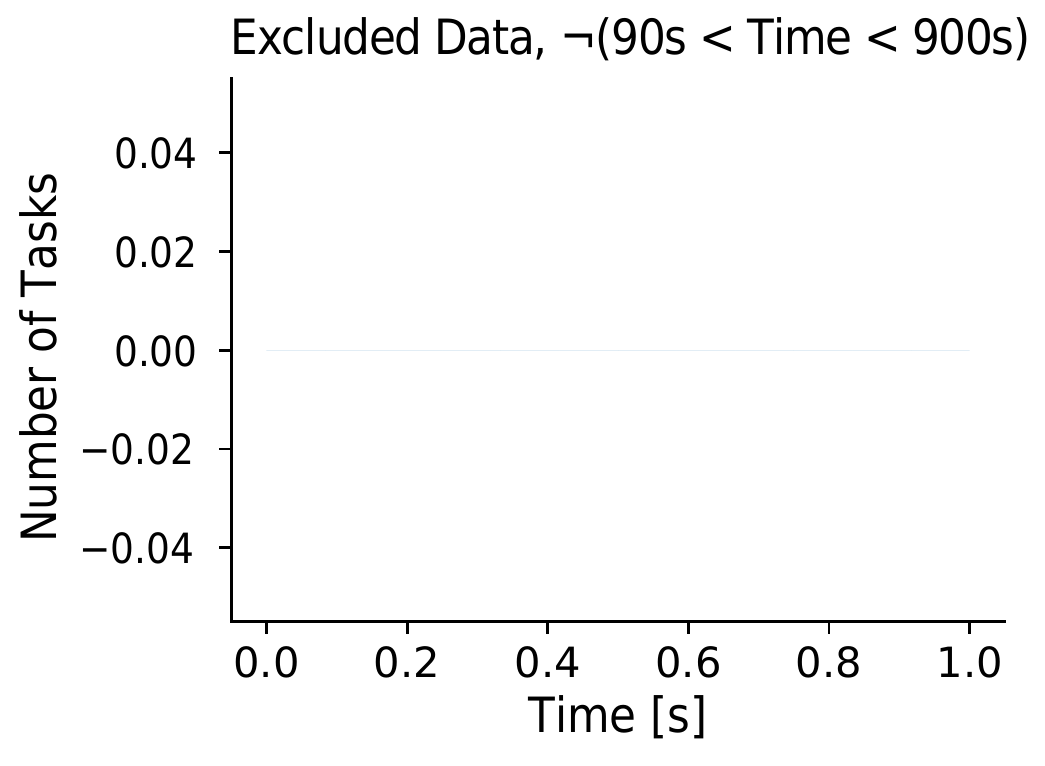}
    \caption{Total response time from excluded data.}
\end{subfigure}
\begin{subfigure}{0.64\textwidth}
    \begin{center}
    \includegraphics[width=\textwidth]{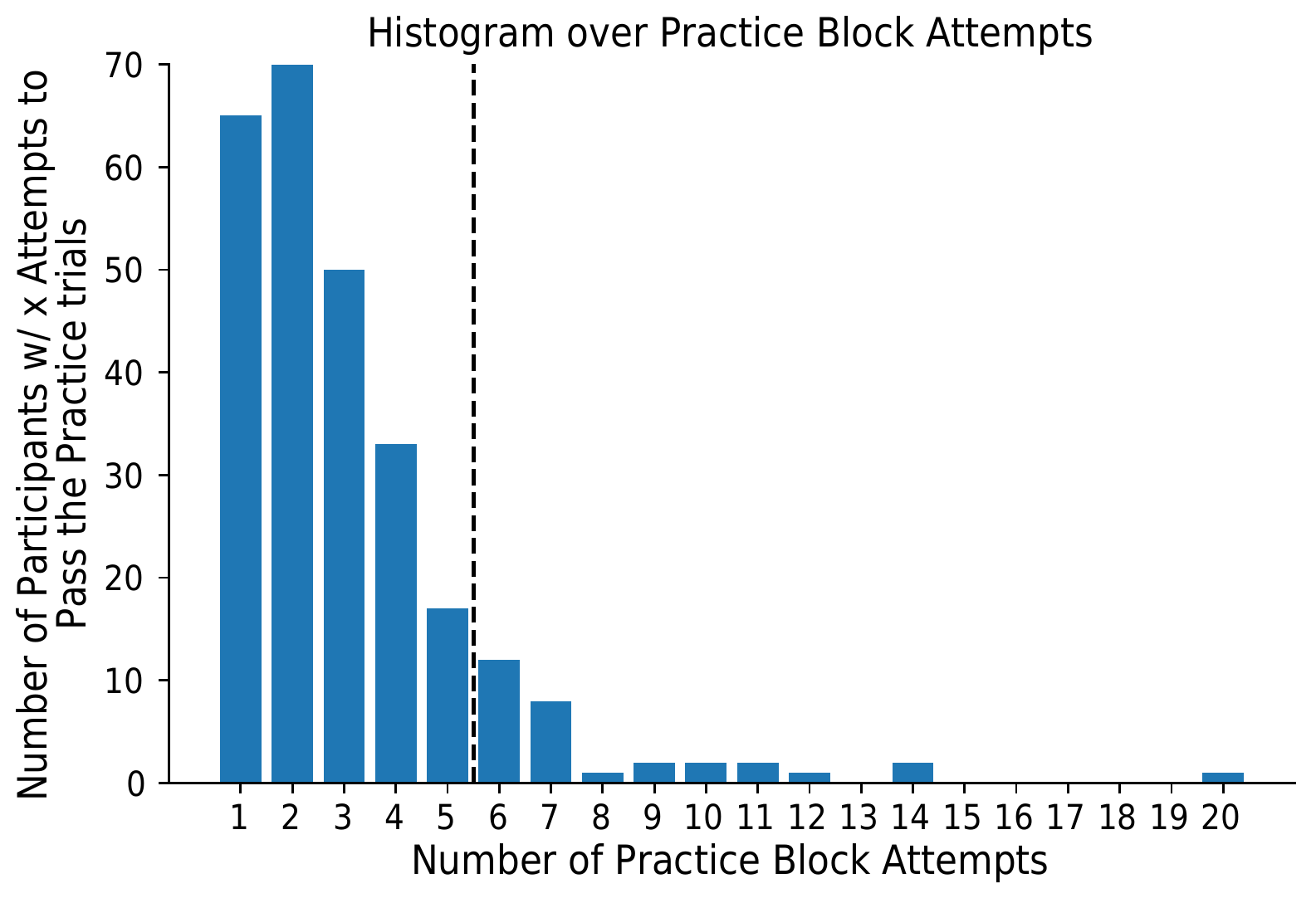}    
    \end{center}
    \caption{Practice Block Attempts: We include data from people who needed $5$ or fewer attempts.}
\end{subfigure}

\end{center}
\caption{Distributions of the individual values controlled by the exclusion criteria in the counterfactual-inspired experiment on MTurk. For the first four criteria, a - c and g (d - f and h) show the data for the included (excluded) data. The final criterion in i shows a joint distribution.}
\label{fig:intervention_experiment_mturk_analysis_details}
\end{figure}

\subsection{Replication of the Main Result of \citet{borowskiandzimmermann2020exemplary}}\label{app:replication}
To check whether collecting data on a crowdsourcing platform yields sensible data in our case, we first test whether we can replicate the main finding of our previous human psychophysical experiment on feature visualizations \citep{borowskiandzimmermann2020exemplary}. In the latter, we found in a well-controlled lab environment that natural reference images are more informative than synthetic ones when choosing which of two different images are more highly activating for a given unit. Below, we report how we alter the experimental set-up to turn the lab experiment into an online experiment on MTurk and what results we find.

\subsubsection{Experimental Set-up}
While keeping as many aspects as possible consistent with our original study \citep{borowskiandzimmermann2020exemplary}, we make a few changes:
(1) We run an online crowdsourced experiment on MTurk, instead of in a lab.
(2) Instead of testing the $45$ units used in the original Experiment~I, we only test one single branch of each Inception module, namely the $3\times3$ kernel size. This is a reasonable decision given that the main finding of the superiority of natural over synthetic images was present in all branches and that there was no significant difference per condition between different branches.
(3) We exchange the within-participant design for a between-participant design, i.e. one MTurk participant does one condition only, namely either the natural or synthetic reference condition. This version is more suitable for short online experiments.
(4) Instead of testing $10$ participants in the lab, we test $130$ MTurk participants per condition, i.e. $260$ in total. This number of participants is estimated with an a priori power analysis using the SIMR package \citep{green2016simr} to allow us to detect an effect half as large as the one observed in \citet{borowskiandzimmermann2020exemplary} $80\%$ of the time. Assumptions about variance, average performance, and effect size are chosen to be conservative relative to the original study because we expect MTurk participants' responses to be noisier.

\textbf{One HIT} on MTurk consists of $1$ extensively explained instruction trial, $2$ practice trials, and then $9$ main trials that are randomly interleaved with a total of $3$ catch trials. 
Each trial type is sampled from a disjoint pool of units: All participants see the same unit for the instruction trial; the catch trials are sampled from the same pool as in the original experiment, %
and the practice trials are the units that were used as interpretability judgment trials in \citep{borowskiandzimmermann2020exemplary}, namely mixed3a, kernel size $1 \times 1$, unit 43; mixed4b, \textsc{Pool}, unit 504; mixed5b, $1\times 1$, unit 17.
A total of $13$ participants see the same main trials that one lab participant saw. The order of the main and catch trials per participants is randomly arranged.

\paragraph{Exclusion Criteria}
If a participant's response does not meet one or more of the following criteria, which were determined before data collection, we discard it and post the same HIT again:
\begin{itemize}[leftmargin=*]
    \item Performance threshold for catch trials: two out of three trials have to be correctly answered
    \item Answer variability: at least one trial must be chosen from the less frequently selected side (to discard participants who only responded with ``up'' or ``down'')
    \item Time to read the instructions: at least $15$\,s
    \item Time for the whole experiment: at least $90$\,s and at most $600$\,s 
\end{itemize}

\paragraph{MTurk compensation} Based on an estimated and pilot experiment duration as well as an hourly rate of US\$\,$15$, we calculate the pay to be US\$\,$1.25$. We pay all MTurk participants who fully complete the experiment regardless of whether they succeed or fail in the exclusion criteria. The experiment without pilot experiments costs US\$\,$447$. MTurk participants whose data we include need a mean time of $220.70 \pm 71.58$\,s for the whole experiment, which results in an hourly compensation of $\approx 20.39$\, US\$/hour.

\subsubsection{Results}
\begin{figure}[!htpb]
    \begin{center}
        \begin{subfigure}{0.24\textwidth}
        \includegraphics[width=\textwidth]{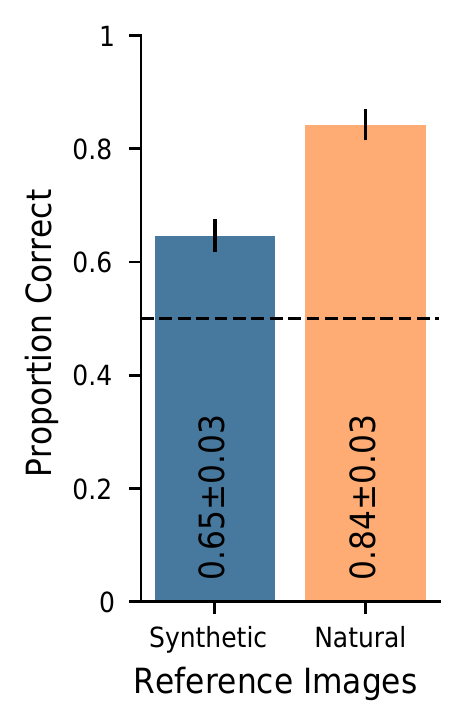}
        \caption{Performance.}
        \end{subfigure}
        
        \begin{subfigure}{0.32\textwidth}
        \includegraphics[width=\textwidth]{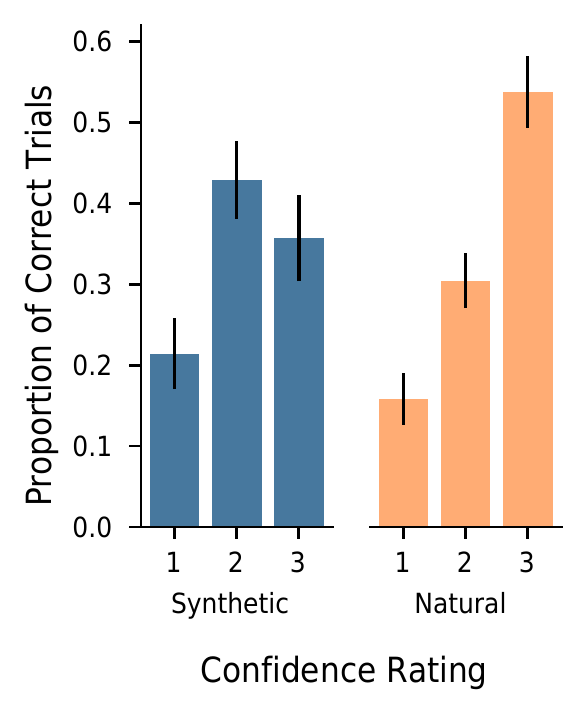}
        \caption{Confidence ratings on correctly answered trials.}
        \end{subfigure}
        \begin{subfigure}{0.32\textwidth}
        \includegraphics[width=\textwidth]{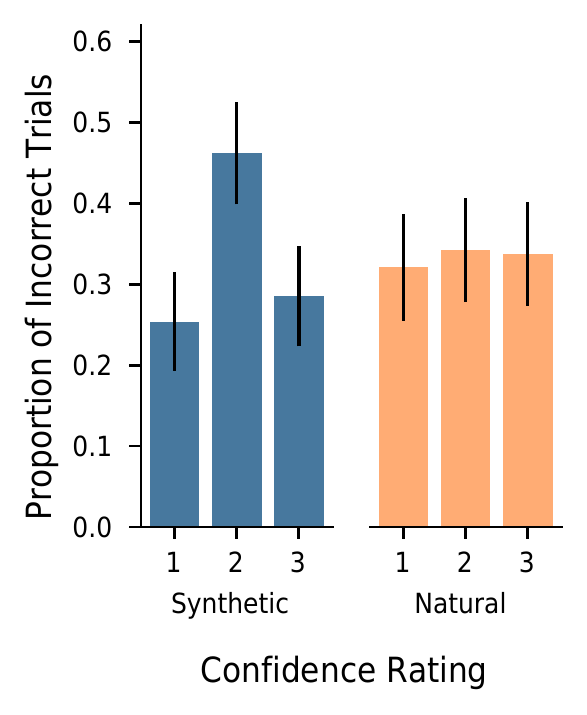}
        \caption{Confidence ratings on incorrectly answered trials.}
        \end{subfigure}
        \begin{subfigure}{0.32\textwidth}
        \includegraphics[width=\textwidth]{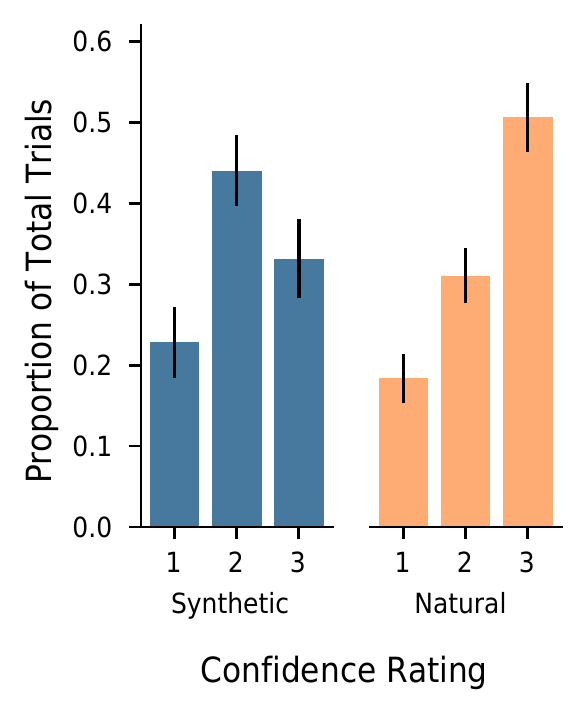}
        \caption{Confidence ratings on all trials. \quad\quad~ }
        \end{subfigure}
        
        \begin{subfigure}{0.26\textwidth}
        \includegraphics[width=\textwidth]{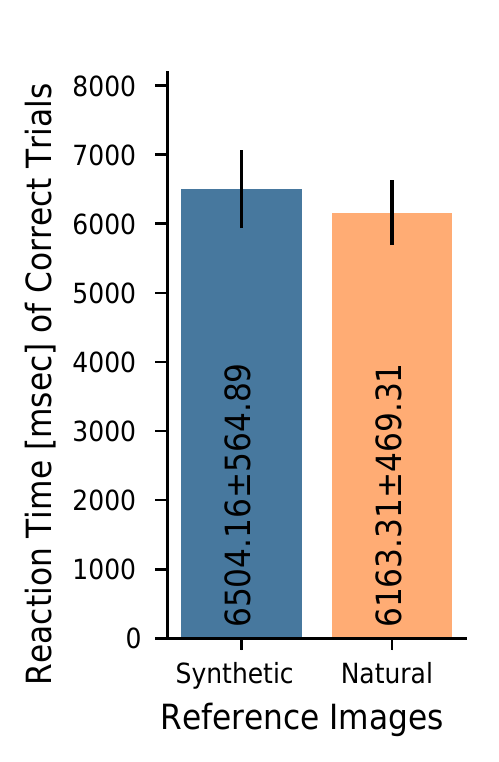}
        \caption{Reaction time on correctly answered trials.}
        \end{subfigure}
        \begin{subfigure}{0.26\textwidth}
        \includegraphics[width=\textwidth]{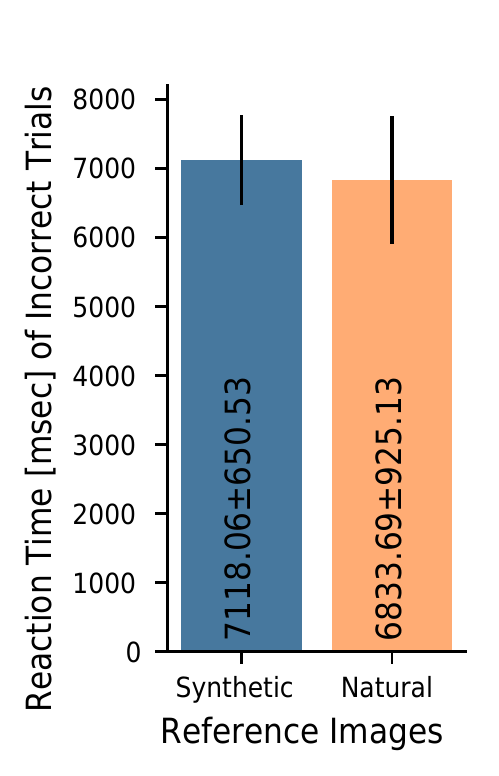}
        \caption{Reaction time on incorrectly answered trials.}
        \end{subfigure}
        \begin{subfigure}{0.26\textwidth}
        \includegraphics[width=\textwidth]{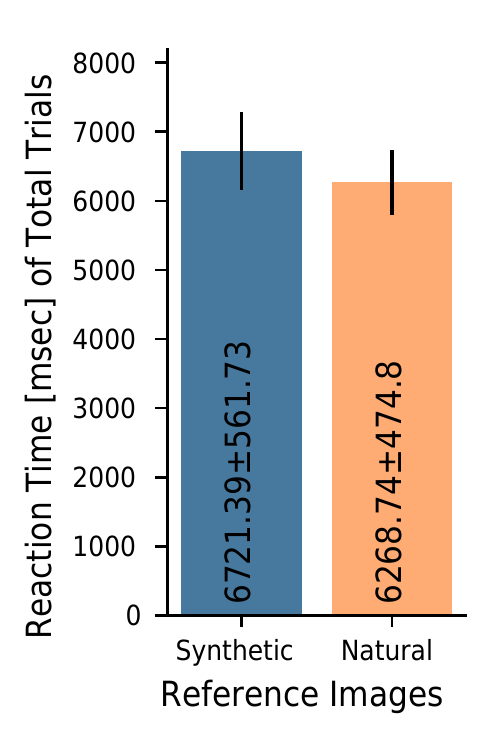}
        \caption{Reaction time on all trials. \quad\quad~ }
        \end{subfigure}
    \end{center}
    \caption{Results of the replication experiment of \citet{borowskiandzimmermann2020exemplary} on MTurk for kernel size $3 \times 3$: task performance (a), distribution of confidence ratings (b-d) and reaction times (e-g).
    }
    \label{fig:replication_experiment_I_perf_conf_rating_rt}
\end{figure}

MTurk participants achieve a higher performance when given natural than synthetic reference images: $84\pm3$\,\% vs. $65\pm3$\,\% (see Fig.~\ref{fig:replication_experiment_I_perf_conf_rating_rt}a). Qualitatively, this result is the same as in the original Experiment~I, see Figure~16 in \citet{borowskiandzimmermann2020exemplary}. More precisely, the data shows a 1.35 (2.1) times larger odds (accuracy) difference for the replication. Compared to the lab data, MTurk participants seem more confident on the synthetic condition (see Fig.~\ref{fig:replication_experiment_I_perf_conf_rating_rt}b-d), are faster in the synthetic condition (see Fig.~\ref{fig:replication_experiment_I_perf_conf_rating_rt}e-g), and are about as fast in the natural condition (see Fig.~\ref{fig:replication_experiment_I_perf_conf_rating_rt}e-g).

\begin{figure}[!htpb]
    \begin{center}
        \begin{subfigure}{0.38\textwidth}
            \includegraphics[width=\textwidth]{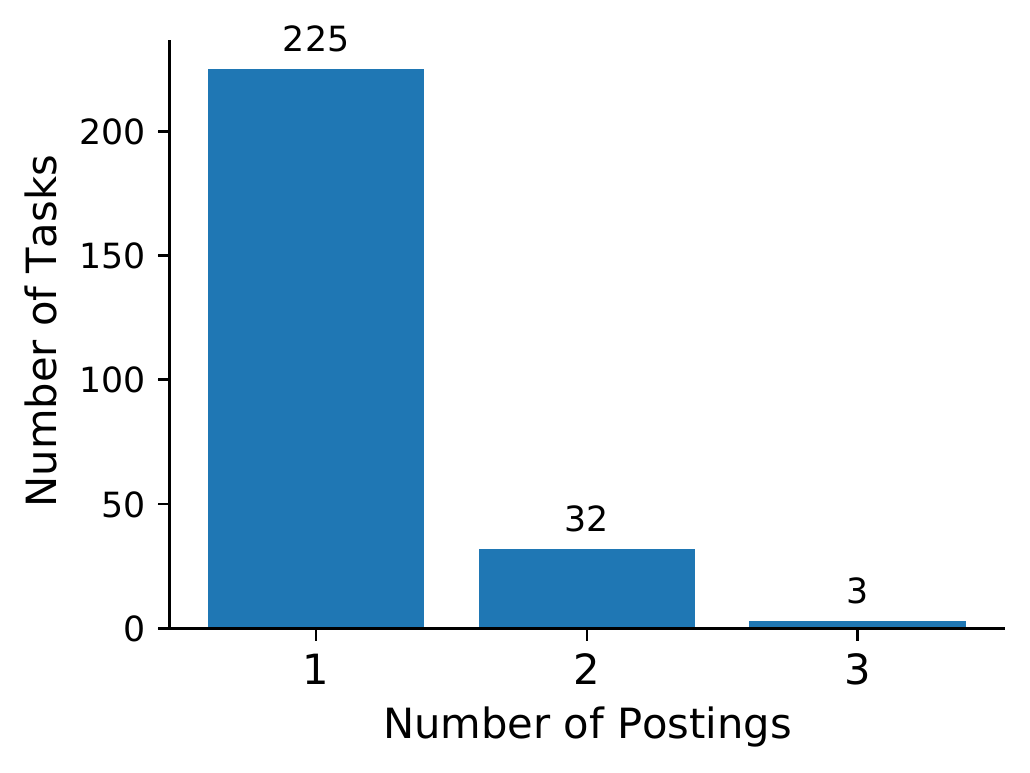}
            \caption{Number of times a HIT is posted.}
        \end{subfigure}
        \begin{subfigure}{0.38\textwidth}
            \includegraphics[width=\textwidth]{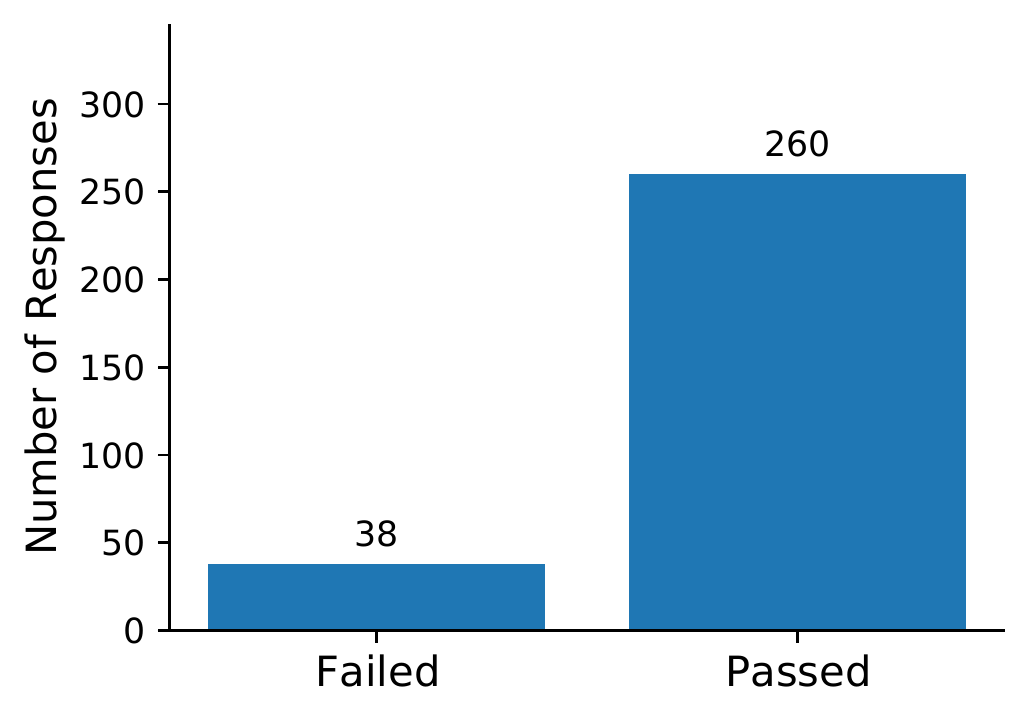}
            \caption{All exclusion criteria.}
        \end{subfigure}
        
        \begin{subfigure}{0.38\textwidth}
            \includegraphics[width=\textwidth]{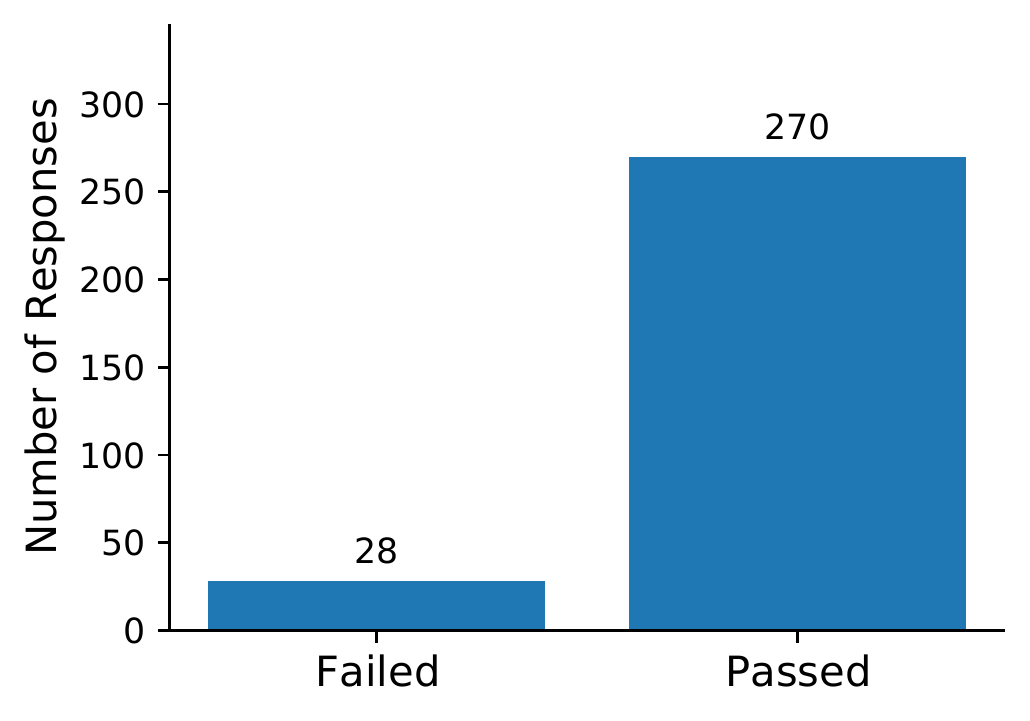}
            \caption{Exclusion criterion: catch trials.}
        \end{subfigure}
        \begin{subfigure}{0.38\textwidth}
            \includegraphics[width=\textwidth]{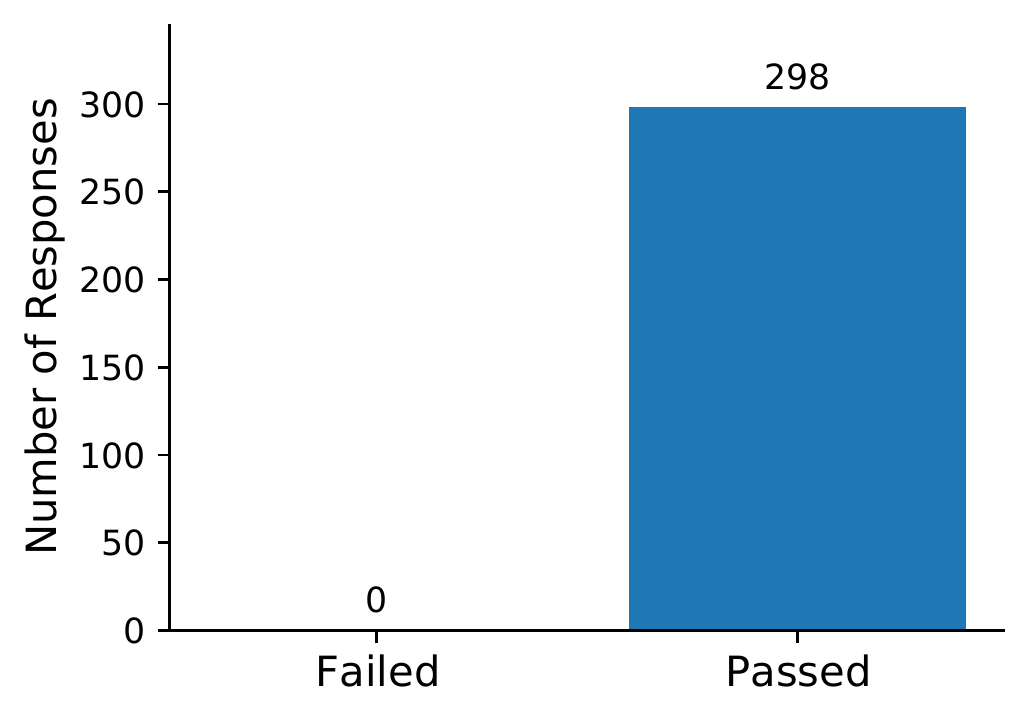}
            \caption{Exclusion criterion: row variability.}
        \end{subfigure}
        
        \begin{subfigure}{0.38\textwidth}
            \includegraphics[width=\textwidth]{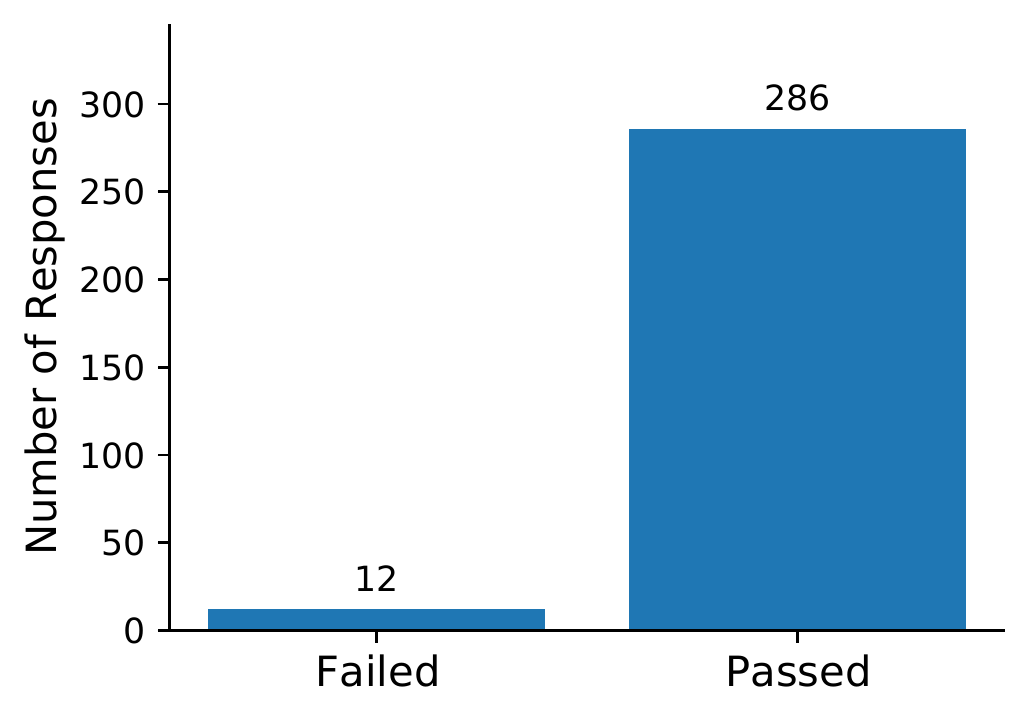}
            \caption{Exclusion criterion: instruction time.}
        \end{subfigure}
        \begin{subfigure}{0.38\textwidth}
            \includegraphics[width=\textwidth]{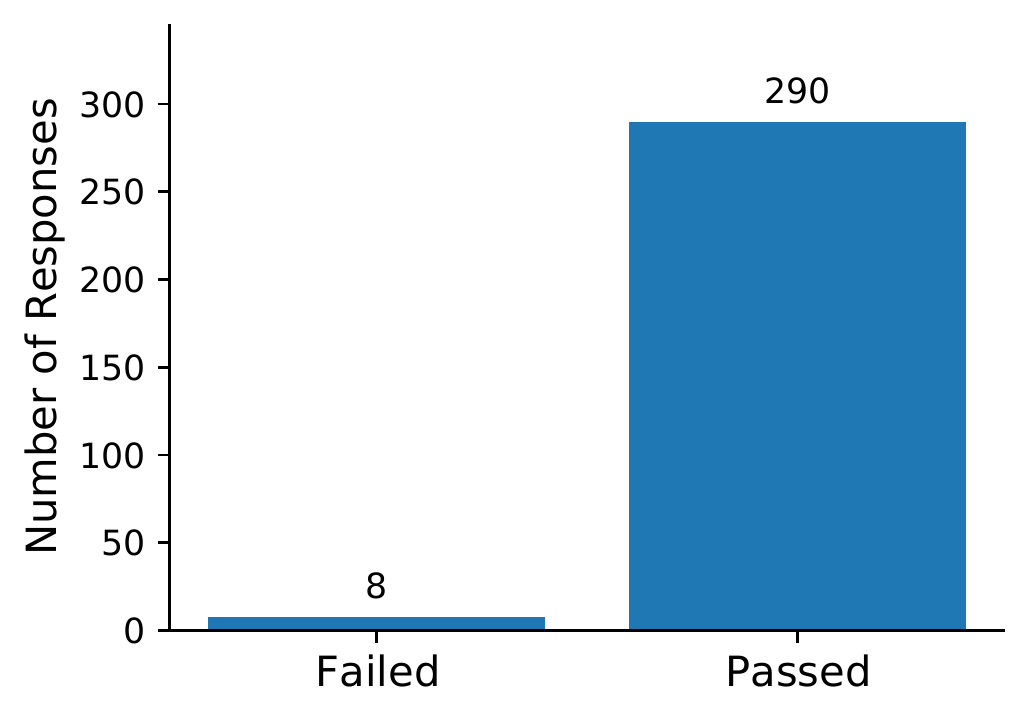}
            \caption{Exclusion criterion: total response time.}
        \end{subfigure}
    \end{center}
    \caption{(a) Number of times a HIT is posted. (b-f) Distributions of MTurk participants that passed/failed the exclusion criteria in the replication experiment on MTurk. Note that the sum of the counts of responses for the individual exclusion criteria in c-f is higher than the summary in b because a participant may have failed more than one exclusion criterion.}
    \label{fig:replication_experiment_mturk_analysis_postings_exclusion_criteria}
\end{figure}

\begin{figure}
    \begin{center}
        \begin{subfigure}{0.32\textwidth}
            \includegraphics[width=\textwidth]{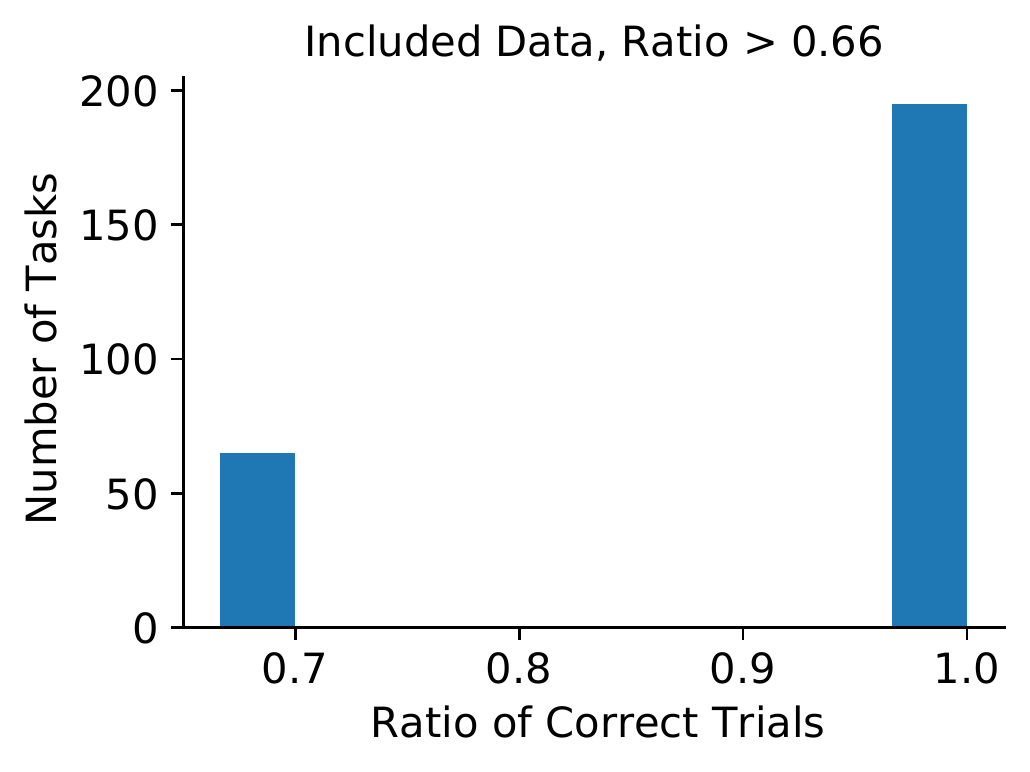}
            \caption{Catch trials from included data.\\\phantom{.}}
        \end{subfigure}
        \begin{subfigure}{0.32\textwidth}
            \includegraphics[width=\textwidth]{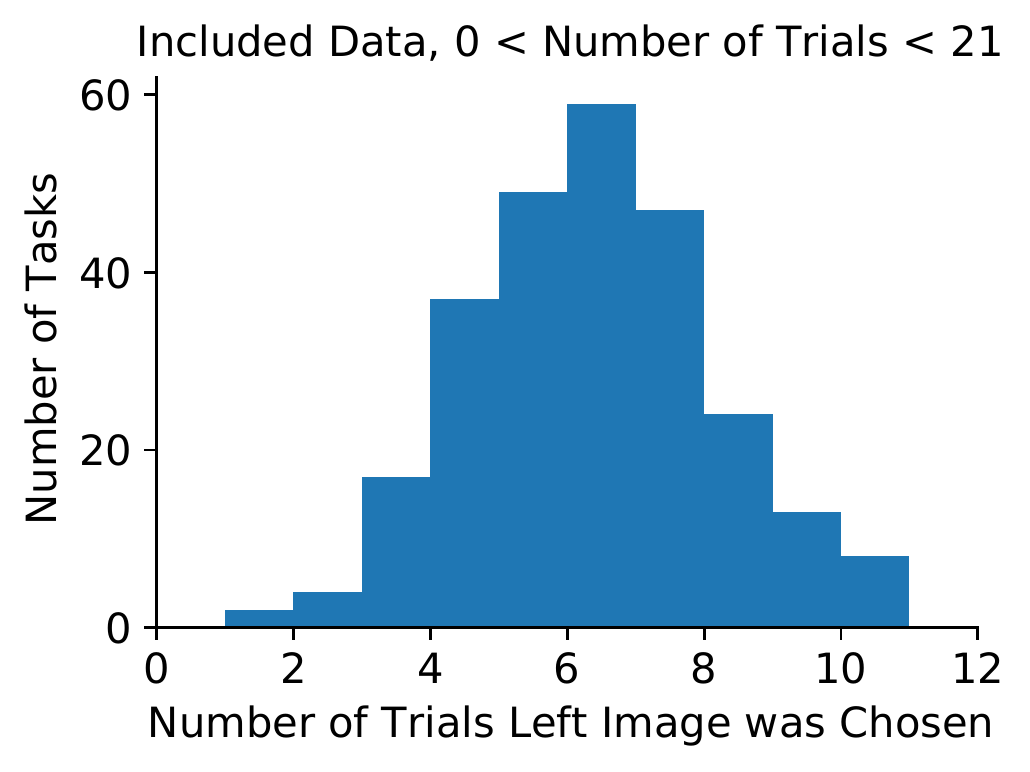}
            \caption{Row variability from included data.}
        \end{subfigure}
        \begin{subfigure}{0.32\textwidth}
            \includegraphics[width=\textwidth]{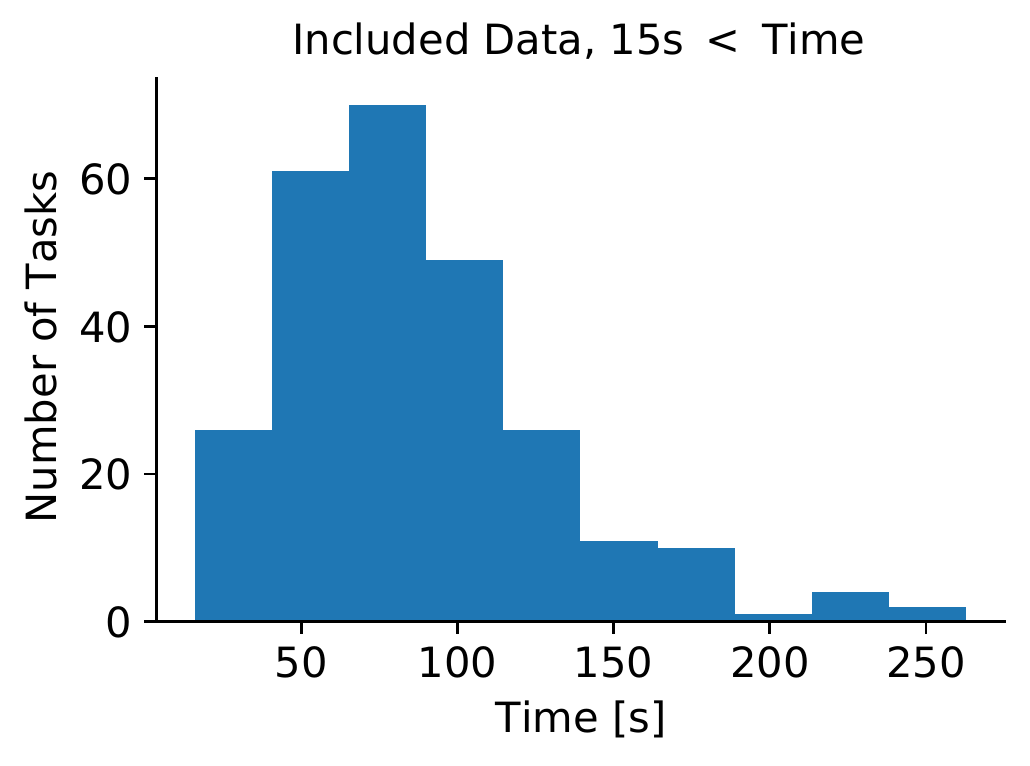}
            \caption{Instruction time from included data.}
        \end{subfigure}
        
        \begin{subfigure}{0.32\textwidth}
            \includegraphics[width=\textwidth]{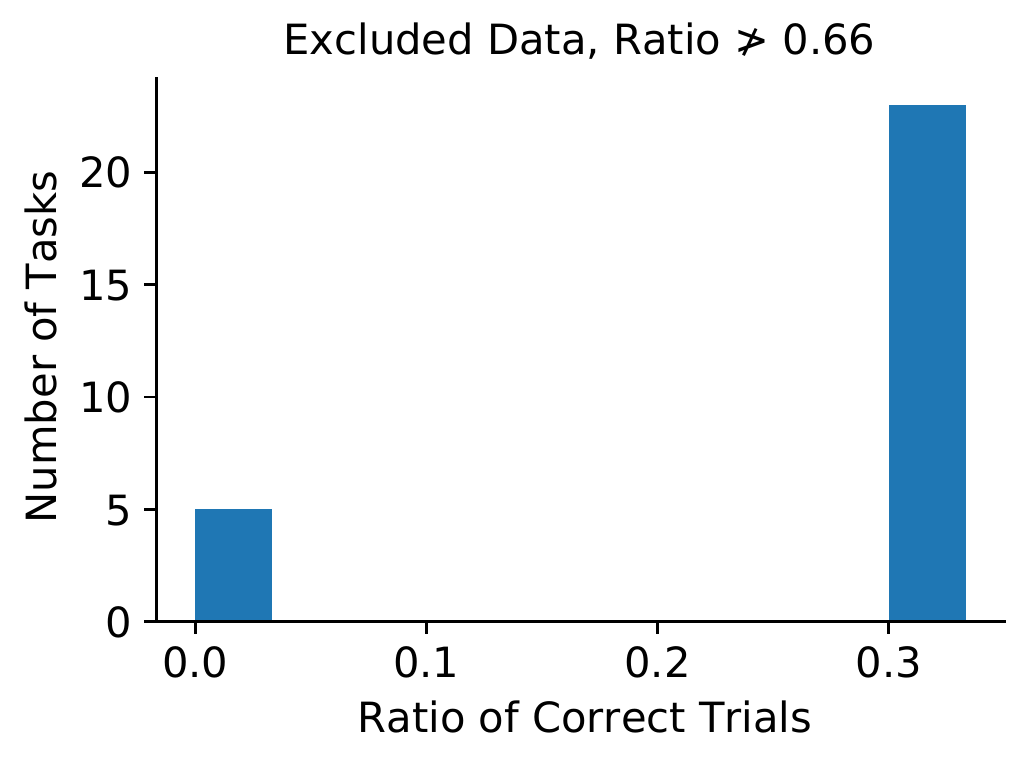}
            \caption{Catch trials from excluded data.\\\phantom{.}}
        \end{subfigure}
        \begin{subfigure}{0.32\textwidth}
            \includegraphics[width=\textwidth]{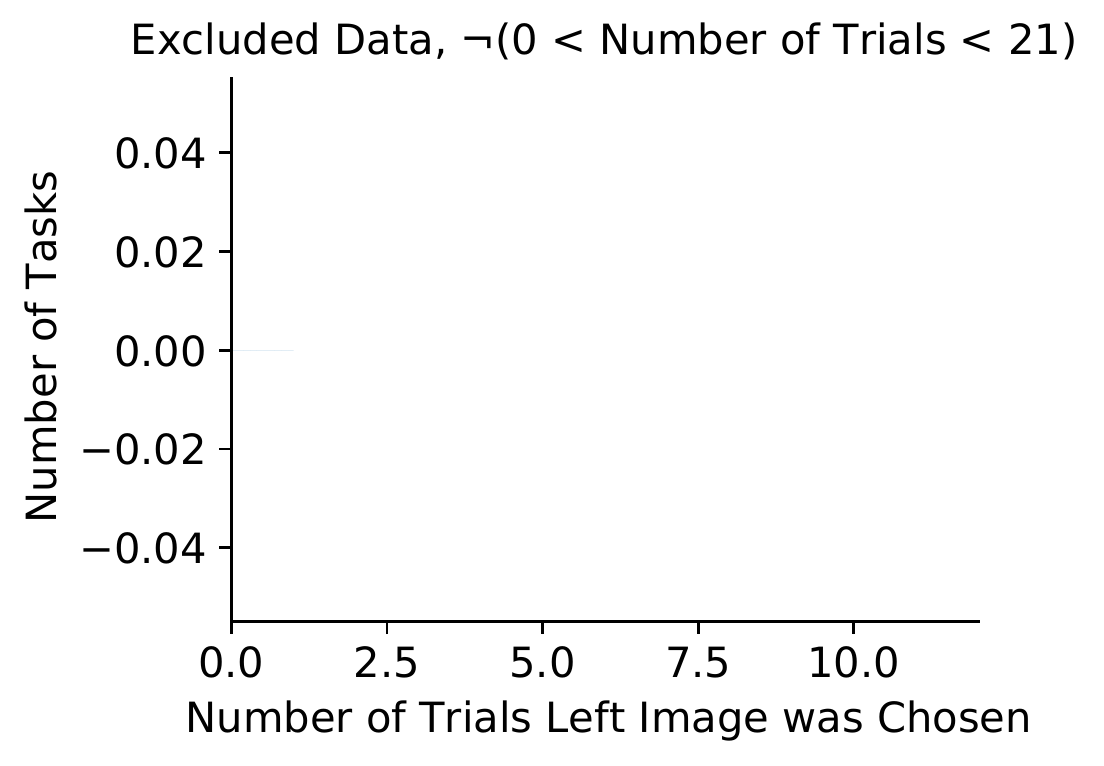}
            \caption{Row variability from excluded data.}
        \end{subfigure}
        \begin{subfigure}{0.32\textwidth}
            \includegraphics[width=\textwidth]{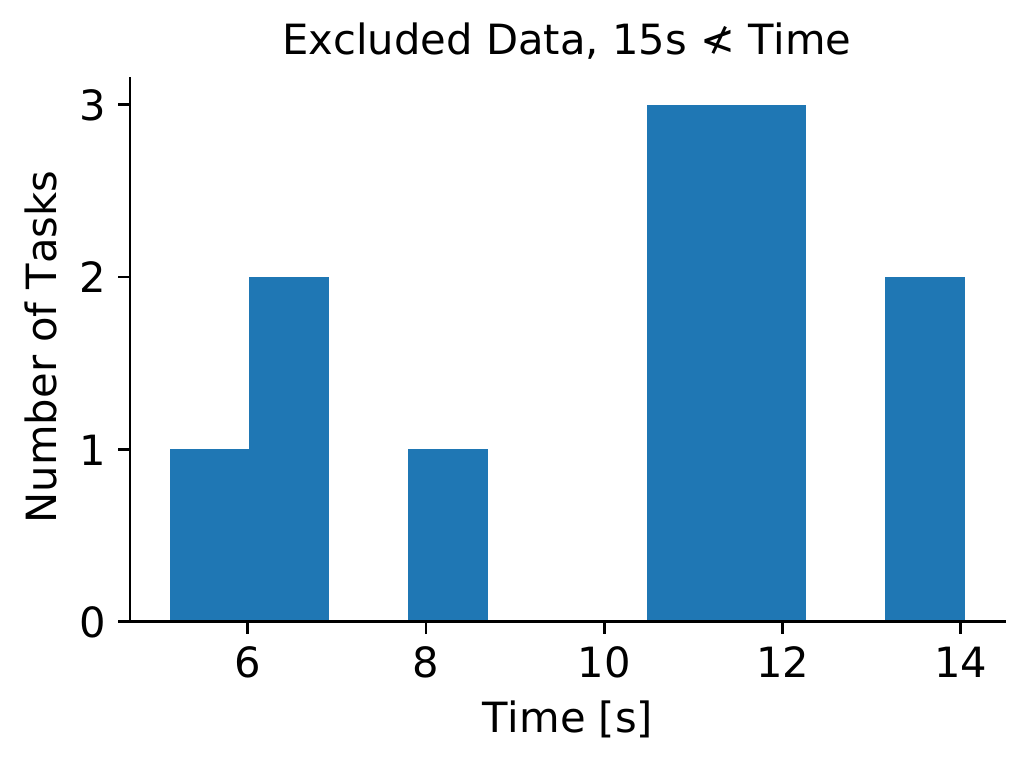}
            \caption{Instruction time from excluded data.}
        \end{subfigure}
        
        \begin{subfigure}{0.32\textwidth}
            \includegraphics[width=\textwidth]{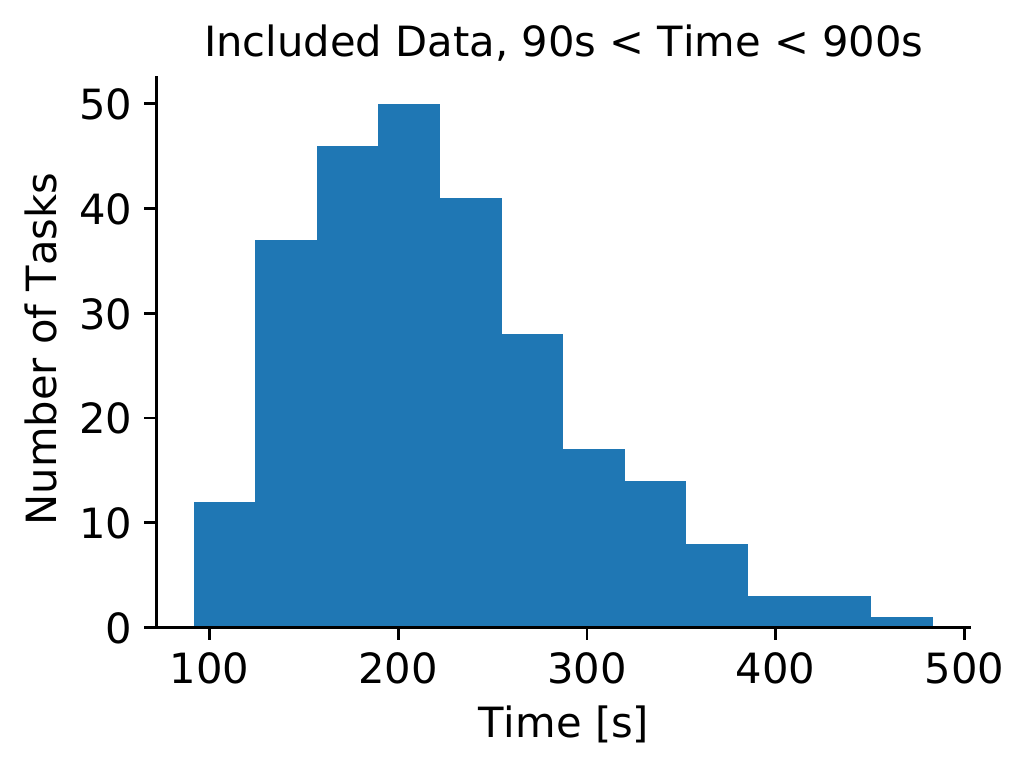}
            \caption{Total response time from included data.}
        \end{subfigure}
        \begin{subfigure}{0.33\textwidth}
            \includegraphics[width=\textwidth]{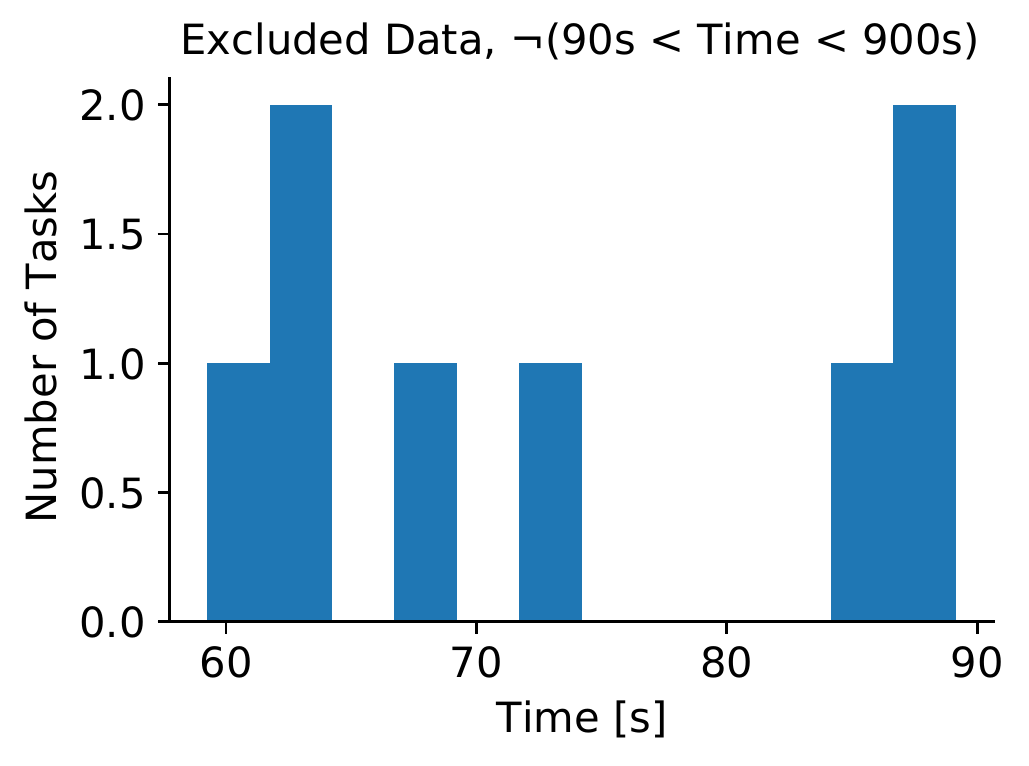}
            \caption{Total response time from excluded data.}
        \end{subfigure}
    \end{center}
    \caption{Distributions of the individual values controlled by the exclusion criteria in the replication experiment on MTurk. Figures a - c and g (d - f and h) show the data for the included (excluded) data.}
    \label{fig:replication_experiment_mturk_analysis_details}
\end{figure}

Fig.~\ref{fig:replication_experiment_mturk_analysis_postings_exclusion_criteria} shows that most participants passed the exclusion criteria. For more details on the number of postings per HIT and for more details on the MTurk participants' performance on the exclusion criteria, see \ref{fig:replication_experiment_mturk_analysis_details}.

\end{document}